\documentclass[10pt,twocolumn,letterpaper]{article}

\usepackage[pagenumbers]{cvpr}

\usepackage{makecell}
\usepackage{multicol}
\usepackage{multirow}
\usepackage{pifont}
\newcommand{\cmark}{\ding{51}}
\newcommand{\xmark}{\ding{55}}

\definecolor{robo_blue}{RGB}{66, 133, 244}
\definecolor{robo_red}{RGB}{192, 0, 0}
\definecolor{robo_yellow}{RGB}{251, 189, 5}
\definecolor{robo_green}{RGB}{0, 176, 80}
\definecolor{robo_gray}{RGB}{100, 100, 100}

\definecolor{cvprblue}{rgb}{0.21,0.49,0.74}
\usepackage[pagebackref,breaklinks,colorlinks,allcolors=cvprblue]{hyperref}

\usepackage{listings}
\usepackage{tcolorbox}
\usepackage{framed}
\usepackage{fontawesome}
\usepackage{titletoc}

\usepackage{tikz}
\newcommand*\circled[1]{\tikz[baseline=(char.base)]{
            \node[shape=circle,fill,inner sep=0.8pt] (char) {\textcolor{white}{#1}};}}

\tcbuselibrary{listings}
\tcbuselibrary{breakable}
\usepackage{marvosym}

\newtcolorbox{codebox}[1][]{
  colback=gray!10,
  colframe=black,
  boxrule=0.4pt,
  arc=2pt,
  top=2pt,
  bottom=2pt,
  left=3pt,
  right=3pt,
  boxsep=3pt,
  fontupper=\small,
  listing only,
  breakable,
  listing options={
    basicstyle=\ttfamily\footnotesize,
    language=Python,
    keywordstyle=\color{blue},
    stringstyle=\color{red},
    commentstyle=\color{gray},
    morekeywords={\quad}
  },
  #1
}

\newtcolorbox{widecodebox}[1][]{
  colback=gray!10,
  colframe=black,
  boxrule=0.4pt,
  arc=2pt,
  top=2pt,
  bottom=2pt,
  left=3pt,
  right=3pt,
  boxsep=3pt,
  fontupper=\small,
  listing only,
  listing options={
    basicstyle=\ttfamily\footnotesize,
    language=Python,
    keywordstyle=\color{blue},
    stringstyle=\color{red},
    commentstyle=\color{gray},
    morekeywords={\quad}
  },
  #1
}

\newcommand\blfootnote[1]{%
\begingroup
\renewcommand\thefootnote{}{}\footnote{#1}%
\addtocounter{footnote}{-1}%
\endgroup
}

\title{
    \vspace{-0.3cm}
    \includegraphics[width=0.046\linewidth]{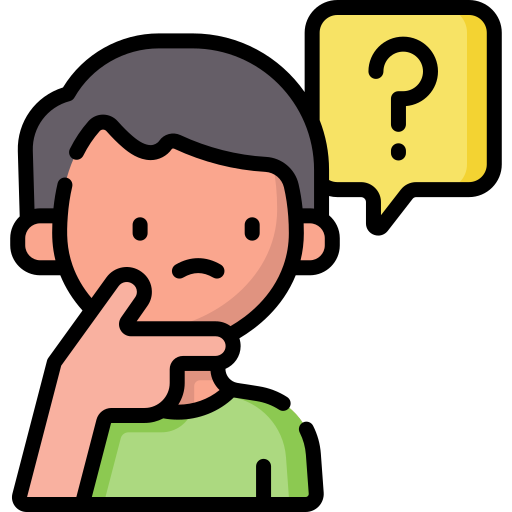}Are VLMs Ready for Autonomous Driving?\\An Empirical Study from the Reliability, Data, and Metric Perspectives
}

\author{
    Shaoyuan Xie$^{\dagger}$\quad
    Lingdong Kong$^{\ddagger,\lozenge,*}$\quad
    Yuhao Dong$^{\ddagger,\S}$\quad
    Chonghao Sima$^{\ddagger,\triangledown}$\\
    Wenwei Zhang$^{\ddagger}$\quad
    Qi Alfred Chen$^{\dagger}$\quad
    Ziwei Liu$^{\S}$\quad
    Liang Pan$^{\ddagger,\textrm{\Letter}}$\vspace{0.2cm}\\
    $^{\dagger}$University of California, Irvine\quad
    $^{\ddagger}$Shanghai AI Laboratory\quad
    $^{\lozenge}$National University of Singapore\\
    $^{\S}$S-Lab, Nanyang Technological University\quad
    $^{\triangledown}$The University of Hong Kong\vspace{0.2cm}\\
    \faGithubAlt~\textbf{Code \& Demo:} \href{https://drive-bench.github.io}{drive-bench.github.io}\\
    \faCar~\textbf{Dataset \& Benchmark:} \href{https://huggingface.co/datasets/drive-bench/arena}{huggingface.co/datasets/drive-bench/arena}
    \\
}
\begin{document}

\twocolumn[{
\renewcommand\twocolumn[1][]{#1}

\maketitle

\begin{center}
    \centering
    \vspace{-0.4cm}
    \includegraphics[width=\linewidth]{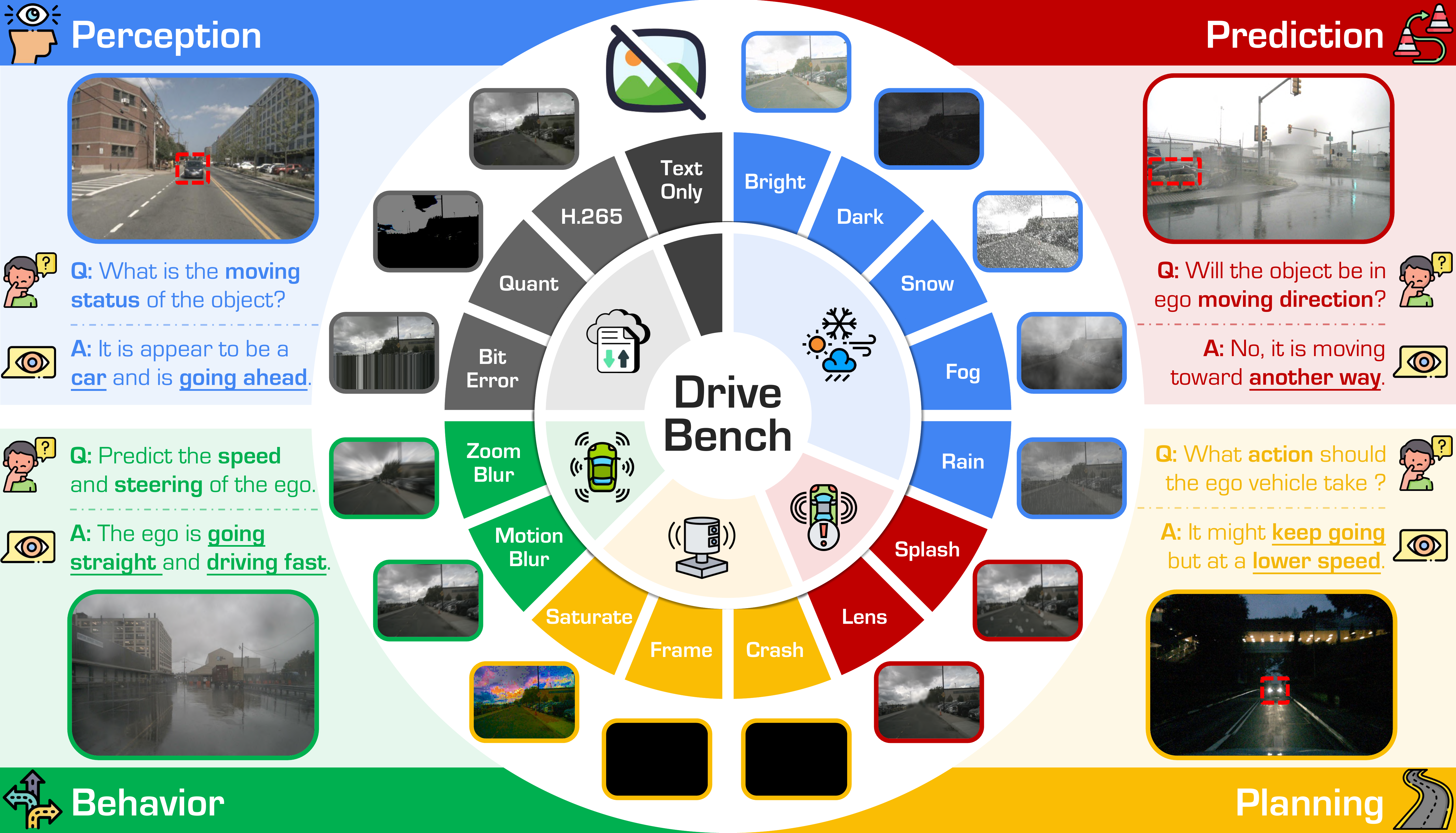}
    \vspace{-0.5cm}
    \captionof{figure}{\textbf{Overview of} \textsf{\textcolor{robo_blue}{Drive}\textcolor{robo_red}{Bench}}. Our benchmark evaluates the reliability and visual grounding of Vision-Language Models (VLMs) in autonomous driving across four mainstream driving tasks -- perception, prediction, planning, and explanation -- under a diverse spectrum of $\mathbf{17}$ settings (clean, corrupted, and text-only inputs). It includes $\mathbf{19,200}$ frames and $\mathbf{20,498}$ QA pairs spanning three question types: multiple-choice, open-ended, and visual grounding. By addressing diverse tasks and conditions, we aim to reveal VLM limitations and promote reliable, interpretable autonomous driving.}
    \label{fig:teaser-bench}
\end{center}
}]

\blfootnote{$(*)$ Project lead.~ ${(\textrm{\Letter})}$ Corresponding author.}
 
\vspace{-0.35cm}
\begin{abstract}
Recent advancements in Vision-Language Models (VLMs) have sparked interest in their use for autonomous driving, particularly in generating interpretable driving decisions through natural language. However, the assumption that VLMs inherently provide visually grounded, reliable, and interpretable explanations for driving remains largely unexamined. To address this gap, we introduce \textsf{\textcolor{robo_blue}{Drive}\textcolor{robo_red}{Bench}}, a benchmark dataset designed to evaluate VLM reliability across $17$ settings (clean, corrupted, and text-only inputs), encompassing $19,200$ frames, $20,498$ question-answer pairs, three question types, four mainstream driving tasks, and a total of $12$ popular VLMs. Our findings reveal that VLMs often generate plausible responses derived from general knowledge or textual cues rather than true visual grounding, especially under degraded or missing visual inputs. This behavior, concealed by dataset imbalances and insufficient evaluation metrics, poses significant risks in safety-critical scenarios like autonomous driving. We further observe that VLMs struggle with multi-modal reasoning and display heightened sensitivity to input corruptions, leading to inconsistencies in performance. To address these challenges, we propose refined evaluation metrics that prioritize robust visual grounding and multi-modal understanding. Additionally, we highlight the potential of leveraging VLMs’ awareness of corruptions to enhance their reliability, offering a roadmap for developing more trustworthy and interpretable decision-making systems in real-world autonomous driving contexts. The benchmark toolkit is publicly accessible.
\end{abstract}    
\section{Introduction}
\label{sec:intro}
With recent advancements in Vision-Language Models (VLMs) \cite{liu2023llava, liu2024llava1.5, liu2024llavanext, qwen, wang2024qwen2, abdin2024phi, chen2023internvl, chen2024far}, there has been increasing research interest in applying VLMs to autonomous driving applications \cite{sima2023drivelm, shao2024lmdrive, tian2024drivevlm, xu2024drivegpt4, ma2023dolphins, fu2024drive, wen2023dilu, wang2024omnidrive}. These efforts span both the design of end-to-end frameworks \cite{sima2023drivelm, tian2024drivevlm, shao2024lmdrive, xu2024drivegpt4} and the integration of VLMs to facilitate interpretable interactions and decisions through natural language \cite{jiang2024senna, ma2023dolphins, pan2024vlp}. Such interpretability is believed to enhance transparency, trustworthiness, and user confidence in autonomous systems \cite{yang2023survey}.

However, previous studies highlight significant limitations in evaluating end-to-end autonomous driving models in open-loop settings \cite{li2024ego}. Instead of focusing on trajectory prediction with potentially unreliable open-loop end-to-end VLMs \cite{sima2023drivelm, mao2023gpt, xu2024drivegpt4, jiang2024senna}, we address another fundamental -- yet underexplored -- question that has been widely assumed \cite{yang2023survey, mao2023gpt, tian2024drivevlm, shao2024lmdrive}:

\textit{``Are existing VLMs capable of providing reliable explanations grounded on visual cues for driving?’’}

To investigate, we examine whether driving decisions generated by VLMs are genuinely grounded in sensory information from the physical environment or reflect general knowledge and fabricated responses from textual cues.

\begin{figure}[t]
    \centering
    \includegraphics[width=\linewidth]{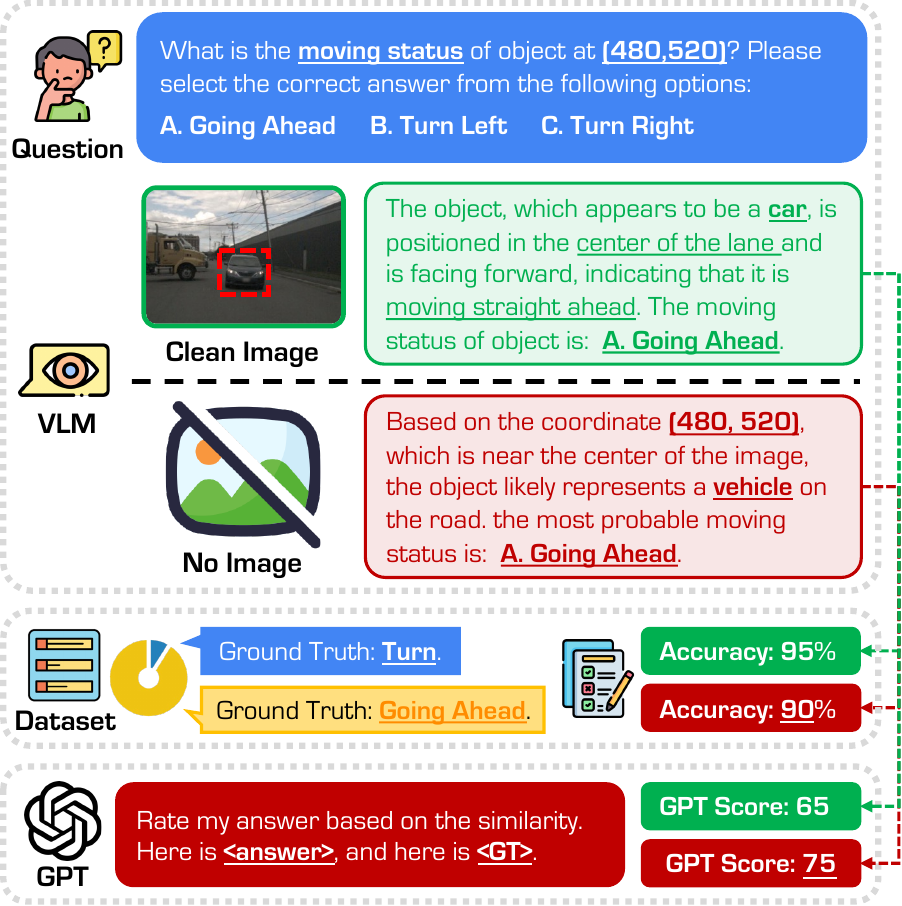}
    \vspace{-0.6cm}
    \caption{\textbf{Do VLMs provide reliable explanations based on visual cues in driving?} We study this from perspectives on reliability, data, and metrics. We find VLMs can fabricate quality answers to driving questions when visual information is absent. The fabricated answers can bypass current metrics, even GPT scores, due to imbalance, lack of context dataset, and problematical evaluation protocols. Our observations challenge the passive assumption that VLMs are more reliable than task-specific models in driving decisions 
    \cite{hu2023planning} because of visual-grounded interpretable responses.}
    \label{fig:teaser}
\end{figure}

\textbf{Model Reliability}. 
To address the fundamental question, we examine VLM reliability through an out-of-distribution (OoD) robustness lens. For this purpose, we introduce \textsf{\textcolor{robo_blue}{Drive}\textcolor{robo_red}{Bench}}, a benchmark encompassing four mainstream driving tasks and $\mathbf{15}$ types of data corruptions, including $\mathbf{19,200}$ images and $\mathbf{20,498}$ QA pairs tailored to real-world autonomous driving scenarios. To assess robustness under extreme conditions, we push visual degradation to its limits by using text-only prompts. Surprisingly, VLMs demonstrate comparable performance to their outputs under ``clean'' visual inputs, even when no visual cues are available (as illustrated in \cref{fig:teaser}). This observation starkly contrasts with human drivers, who would struggle under such degraded conditions. A closer analysis reveals that this apparent ``resilience'' is often a byproduct of imbalanced datasets and suboptimal evaluation protocols rather than inherent model robustness.

\textbf{Datasets.} 
We perform an in-depth analysis of existing \textit{``Driving with Language''} benchmarks \cite{sima2023drivelm, kim2018textual, qian2024nuscenes, xu2020explainable, chen2024automated} and identify critical shortcomings, particularly concerning dataset imbalance. Many of these benchmarks, built on popular driving datasets such as nuScenes \cite{qian2024nuscenes}, BDD \cite{yu2020bdd100k}, and Waymo Open \cite{sun2020waymo}, inherit limitations from their original designs \cite{li2024ego}. For instance, imbalanced data distributions skew evaluations, enabling overly simplistic answers such as \textit{Going Ahead’’} to achieve over $90\%$ accuracy for motion-related queries. Furthermore, the reliance on single-frame questions -- often reliant on temporal context -- creates challenges even for human annotators. Consequently, these benchmarks exhibit inherent biases and persistent negative samples, which diminish the interpretability and reliability of evaluation outcomes.

\textbf{Metrics.} 
We also reevaluate existing metric designs critically. Language-driven interactions in driving applications are often assessed using traditional pattern-matching metrics such as ROUGE \cite{lin2004rouge}, BLEU \cite{papineni2002bleu}, and CIDEr \cite{vedantam2015cider}, which were originally developed for summarization and translation tasks. However, as noted in \cite{tran2019does, anderson2016spice, evtikhiev2023out, akter2022revisiting}, these metrics face significant limitations in evaluating nuanced language-based driving decisions. Even modern evaluators like GPT-based scoring \cite{liu2024aligning, chen2024driving, goli2024frontiers, sima2023drivelm} provide limited insights without task-specific rubrics. These constraints underscore the need for metrics that effectively capture reasoning, contextual understanding, and safety-critical aspects. We advocate for the development of advanced evaluation metrics that incorporate task-specific rubrics, structured question formats, and contextual driving information to more accurately assess VLMs in the real world.

Through a series of comprehensive experiments, we derive several key insights from our analysis, spanning \textbf{$\mathbf{17}$ settings} (\ie, clean, text-only, and various corrupted inputs), \textbf{$\mathbf{12}$ VLMs} (including both open-sourced and commercial models), \textbf{$\mathbf{5}$ tasks} (perception, prediction, planning, behavior, and corruption identification), and \textbf{$\mathbf{3}$ evaluation metrics} (accuracy scores, traditional language metrics \cite{papineni2002bleu, lin2004rouge}, and GPT scores). These findings shed light on the current challenges in integrating VLMs into driving scenarios:
\\
\textcolor{robo_blue}{$\circled{1}$ \textbf{Fabricated responses under degradation:}} VLMs often produce plausible yet fabricated responses under degraded visual conditions, including scenarios where no visual cues exist. This raises concerns about their reliability and trustworthiness, as such behaviors are difficult to detect using existing datasets and evaluation protocols.
\\
\textcolor{robo_red}{$\circled{2}$ \textbf{Awareness of visual corruptions:}} While VLMs exhibit some awareness of visual corruptions, they only explicitly acknowledge these issues when directly prompted. This highlights the models’ limited capacity to autonomously assess the reliability of visual inputs and provide scenario-specific, safety-focused responses.
\\
\textcolor{robo_yellow}{$\circled{3}$ \textbf{Impact of dataset biases:}} Highly biased datasets and suboptimal evaluation protocols can create misleading perceptions of VLM performance. In many cases, VLMs rely on general knowledge rather than actual visual cues to generate responses, which can unexpectedly achieve high scores with existing metrics.
\\
\textcolor{robo_green}{$\circled{4}$ \textbf{Need for tailored metrics:}} Current evaluation metrics, including traditional language-based metrics \cite{papineni2002bleu, lin2004rouge} and GPT scores \cite{chen2024driving, sima2023drivelm}, fail to capture the nuanced requirements of autonomous driving tasks. There is an urgent need for the development of specialized metrics that account for reasoning, contextual understanding, and safety-critical aspects to evaluate VLMs more effectively.

Our findings through \textsf{\textcolor{robo_blue}{Drive}\textcolor{robo_red}{Bench}} not only highlight the need for improved datasets and evaluation protocols but also pave the way for developing safer, more interpretable VLMs for real-world autonomous systems.

\section{Related Work}
\label{sec:related_work}

\noindent\textbf{Driving with Language.} VLMs \cite{liu2023llava, liu2024llava1.5, liu2024llavanext, qwen, wang2024qwen2, abdin2024phi} have demonstrated remarkable human-level reasoning and understanding across diverse domains \cite{brohan2023rt, chen2024spatialvlm, tian2024drivevlm, stone2023open, hong2024cogagent, yang2025octopus, dong2024insight, liu2024chain, liu2024coarse, chen2024drivinggpt, xu2024vlm, cui2024drive}. This capability has raised the prospect of utilizing VLMs to manage complex and unpredictable scenarios in autonomous driving \cite{yang2023survey}. Additionally, the language-based interaction that VLMs offer can help mitigate the black-box nature of deep neural networks by providing explanatory feedback that accompanies their decisions. Driven by these advantages, a growing body of research has begun investigating the deployment of VLMs in autonomous driving \cite{fu2024drive, wen2023dilu, sima2023drivelm, tian2024drivevlm, ma2023dolphins, xu2024drivegpt4}. Early works \cite{fu2024drive, wen2023dilu} leveraged LLMs for decision-making in simplified driving simulators (\eg, HighwayEnv \cite{highway-env}) by offering context-driven descriptions. More recent advancements in VLM architectures \cite{sima2023drivelm, tian2024drivevlm, ma2023dolphins, xu2024drivegpt4} enable these models to interact directly with environments through multimodal (visual and language) inputs. However, despite these advancements, the robustness and reliability of VLMs in complex, real-world autonomous driving tasks remain largely untested, especially given that reliable performance across diverse driving situations is a fundamental requirement for their application in autonomous driving.

\noindent\textbf{Datasets \& Metrics.} To support the integration of VLMs in autonomous driving, several multimodal datasets have been proposed \cite{kim2018textual, xu2020explainable, qian2024nuscenes, deruyttere2019talk2car, wu2023language, sima2023drivelm}. These datasets typically augment established driving benchmarks, such as BDD \cite{yu2020bdd100k} and nuScenes \cite{qian2024nuscenes}, with language-based annotations that enable language-driven perception and decision-making. Certain datasets, such as DriveLM \cite{sima2023drivelm}, incorporate advanced structures like graphs \cite{qian2024nuscenes, sima2023drivelm} to further support the reasoning capabilities of VLMs, thus offering a richer context for nuanced decision-making. DriveLM \cite{sima2023drivelm} is notable for its extensive, human-annotated language data based on the nuScenes dataset, covering the full range of autonomous driving tasks, including perception, prediction, planning, and control. Nonetheless, despite these advances, current datasets and metrics may still lack the comprehensive scope needed to capture the full spectrum of real-world driving complexities, particularly in evaluating open-ended questions generated by VLMs, where responses require a detailed understanding of diverse scenarios.

\noindent\textbf{VLM Reliability.} Deep neural networks have historically struggled with out-of-distribution (OoD) data, a limitation of particular concern in autonomous driving, where failing to handle rare or unexpected scenarios could result in severe safety risks \cite{xie2024benchmarking, kong2023robo3d, kong2024robodepth}. Large-scale models such as CLIP \cite{radford2021learning}, trained on extensive internet-sourced datasets, have shown enhanced robustness to such challenging corner cases \cite{nguyen2022quality, fang2022data}, which suggests that VLMs, trained on vast, diverse datasets -- may have the inherent common-sense reasoning capabilities to address these challenges better than traditional, task-specific models \cite{yang2023survey, li2024r}. However, this hypothesis remains under-investigated in two critical areas: 1) the reliability of VLMs to maintain accurate reasoning when exposed to visual corruptions, and 2) their capacity to detect and interpret potential visual anomalies that could impact safe vehicle maneuvering. In this work, we provide a systematic evaluation of the reliability of current VLMs under conditions of visual corruption, identifying potential limitations that impact their applicability in real-world driving contexts.

\section{\text{\textcolor{robo_blue}{Drive}\textcolor{robo_red}{Bench}}: Driving with VLMs}

In this section, we detail the construction of our benchmark designed to assess the reliability of VLMs within the domain of autonomous driving.

\begin{figure}[t]
    \centering
    \begin{subfigure}[b]{0.484\linewidth}
        \centering
        \includegraphics[width=\linewidth]{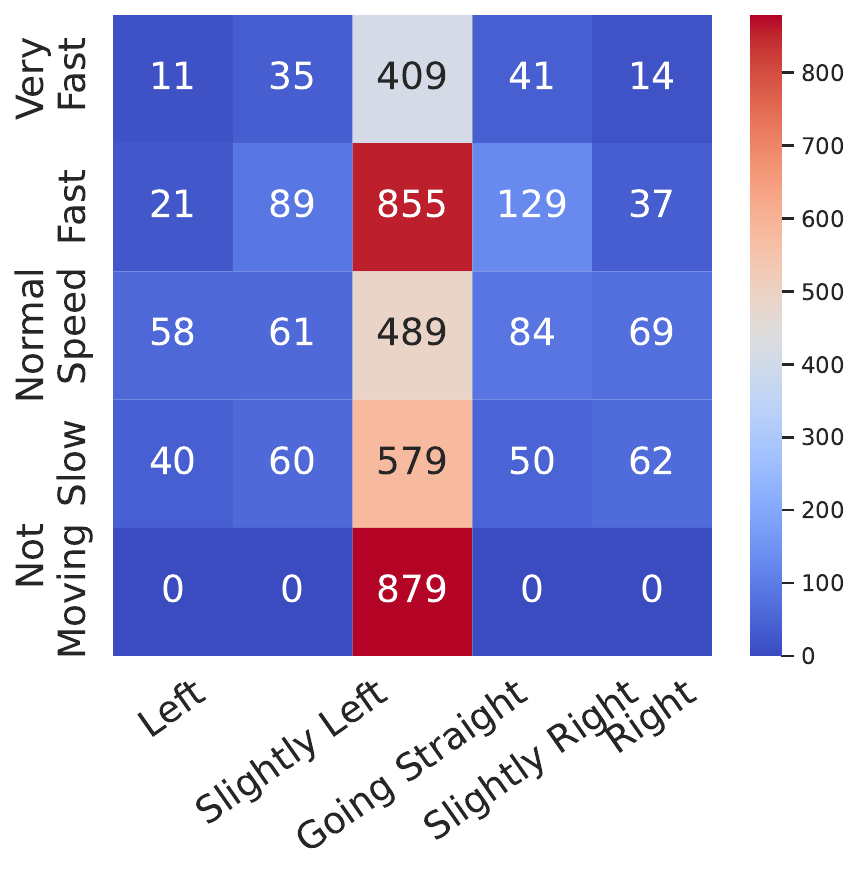}
        \caption{Training Set}
        \label{fig:drivelm-distribution-train}
    \end{subfigure}
    ~
    \begin{subfigure}[b]{0.484\linewidth}
        \centering
        \includegraphics[width=\linewidth]{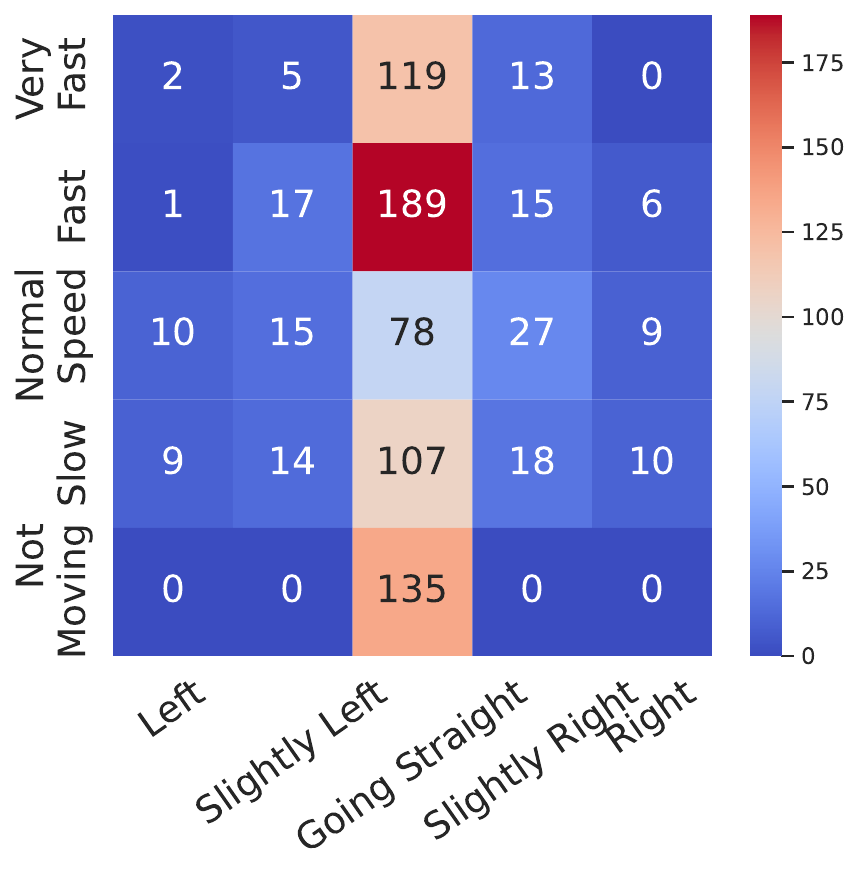}
        \caption{Validation Set}
        \label{fig:drivelm-distribution-val}
    \end{subfigure}
    \vspace{-0.6cm}
    \caption{\textbf{The behavior distributions of steering and speed} in DriveLM-nuScenes \cite{sima2023drivelm}. The majority actions of vehicle behaviors are \textit{``Going Ahead''}, which has also been noted in \cite{li2024ego}.}
    \label{fig:drivelm-distribution}
\end{figure}

\begin{figure}[t]
    \centering
    \begin{subfigure}[b]{0.484\linewidth}
        \centering
        \includegraphics[width=\linewidth]{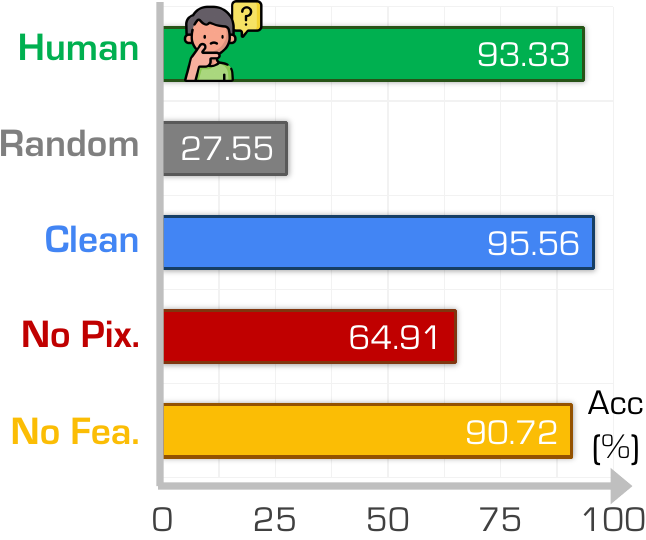}
        \caption{Perception Task}
    \end{subfigure}
    ~
    \begin{subfigure}[b]{0.484\linewidth}
        \centering
        \includegraphics[width=\linewidth]{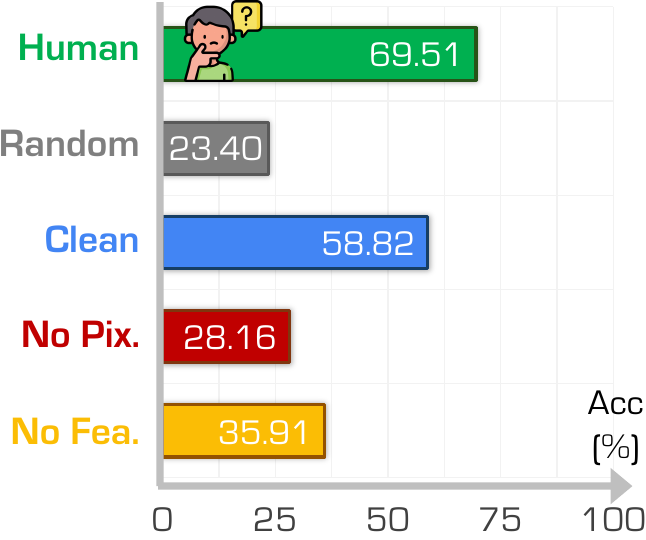}
        \caption{Behavior Task}
    \end{subfigure}
    \vspace{-0.7cm}
    \caption{\textbf{The accuracy scores of \includegraphics[width=0.04\linewidth]{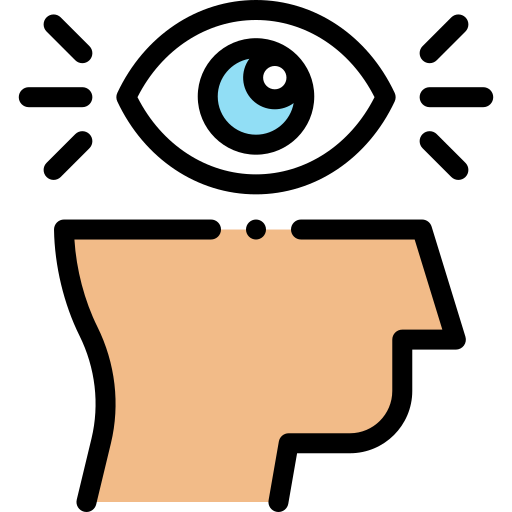}~perception and \includegraphics[width=0.04\linewidth]{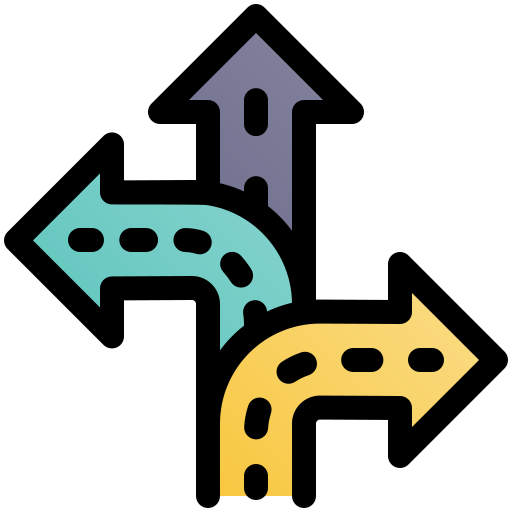}~behavior tasks} under different visual inputs. The results are from DriveLM-Agent \cite{sima2023drivelm}. \textbf{No Pix.} and \textbf{No Fea.} denote zero image pixel and zero feature, respectively.}
    \label{fig:llama-adapter}
\end{figure}

\begin{figure}[t]
    \centering
    \includegraphics[width=\linewidth]{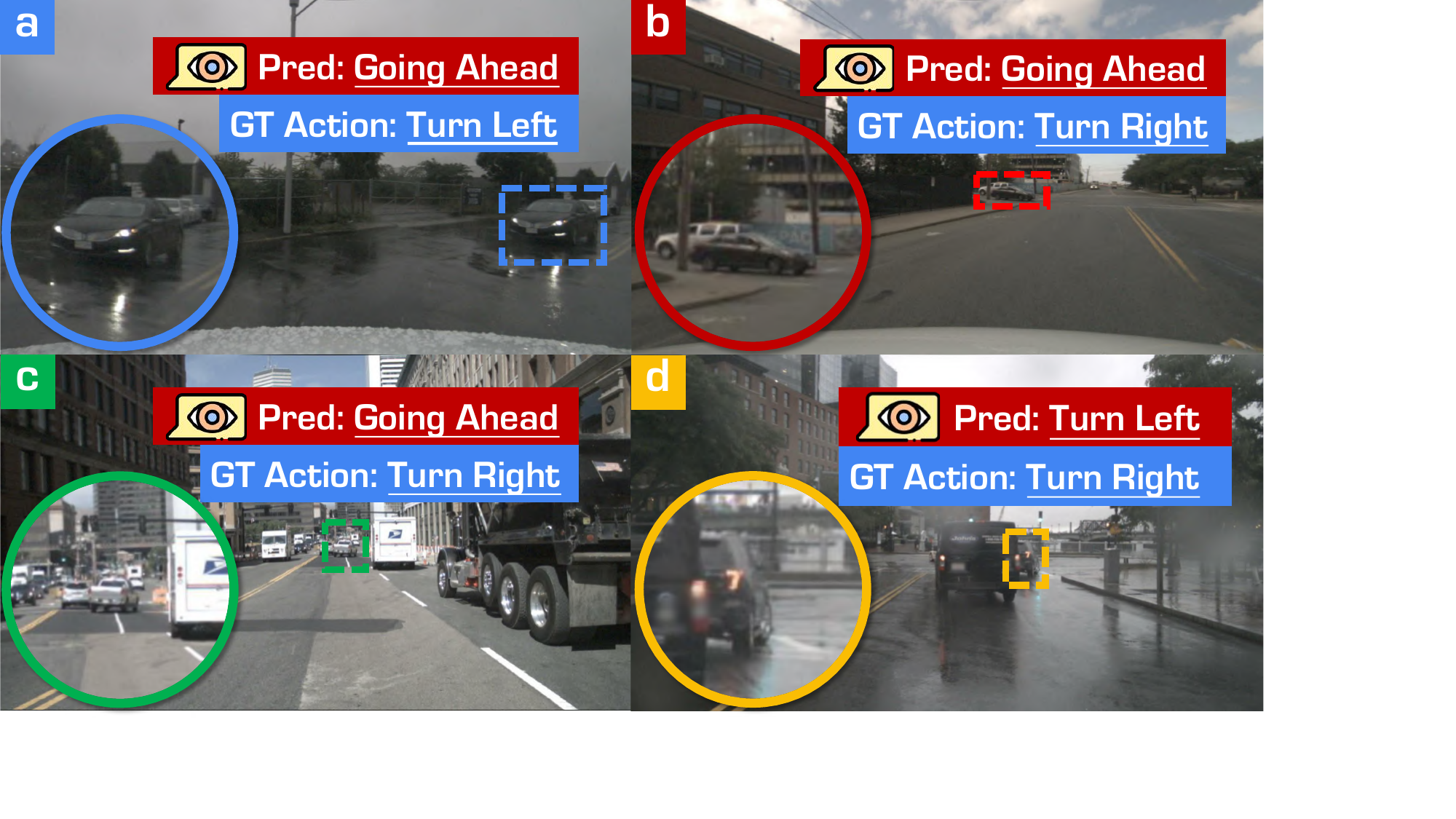}
    \vspace{-0.6cm}
    \caption{\textbf{Challenging cases in existing dataset}. The results are from GPT4-o \cite{achiam2023gpt}. \textbf{(a)}: The black sedan is turning left, indicated by the turn lights. \textbf{(b)}: The black sedan is turning right. The model predicts both \textit{Going Ahead}. The examples show challenging cases for \textit{Turn} choice, where the visual cues are too subtle or rely on temporal context for correct predictions. \textbf{(c)} and \textbf{(d)} are both \textit{Turning Right}, but the model fails to locate the objects based on center pixel positions due to the existence of overlapping or occlusion. Zoomed-in for more details.}
    \label{fig:gpt-fail-case}
\end{figure}

\subsection{Datasets}
In this section, we start building our \textsf{\textcolor{robo_blue}{Drive}\textcolor{robo_red}{Bench}} with representative driving with language datasets~\cite{sima2023drivelm}. We choose DriveLM~\cite{sima2023drivelm} as it is one of the most representative datasets for driving with languages, considering the impact that the dataset serves as one of the benchmarks in Foundation Models for Autonomous Systems Workshop~\cite{contributors2023drivelmrepo, opendrivelab2024workshop}. The dataset spans five tasks, including perception, prediction, planning, behavior, and control. For each task, different sets of questions are applied, such as multiple-choice questions (MCQs), and visual question answering (VQA). The comparison between our dataset and related benchmarks can be seen in \cref{tab:stats}.

\noindent{\textbf{Distribution Bias.}} Through detailed examination, we identify a significant distribution bias in the dataset, which is naturally inherited from the nuScenes dataset \cite{caesar2020nuscenes, li2024ego}. Specifically, in behavior-based multiple-choice questions (MCQs) that inquire about the future movement of the ego vehicles, approximately $78.6\%$ of responses are labeled as \textit{Going Straight} as shown in~\cref{fig:drivelm-distribution}. Consequently, randomly selecting \textit{Going Straight} as an answer can yield accuracy levels exceeding $70\%$, which is concerning since the fine-tuning processes~\cite{sima2023drivelm} further encourage the model to memorize majority choices. To address this imbalance, in \textsf{\textcolor{robo_blue}{Drive}\textcolor{robo_red}{Bench}}, we carefully re-sampled the data to create a more balanced distribution among different options.

\begin{table*}[t]
    \centering
    \caption{\textbf{Comparisons among evaluation benchmarks} for driving. ``\textbf{Per.}'', ``\textbf{Pre.}'', ``\textbf{Beh.}'', ``\textbf{Pla.}'', ``\textbf{Rob.}'' refer to the Perception, Prediction, Behavior, Planning, and Robustness tasks, respectively. GPT$_{\text{ctx}}$ represents GPT scores augmented with context information.}
    \vspace{-0.2cm}
    \label{tab:stats}
    \resizebox{\linewidth}{!}{
    \begin{tabular}{r|ccccc|cc|c|c}
    \toprule
    \multirow{2}{*}{\textbf{Benchmark}} & \includegraphics[width=0.025\linewidth]{figures/icons/perception.png} & \includegraphics[width=0.025\linewidth]{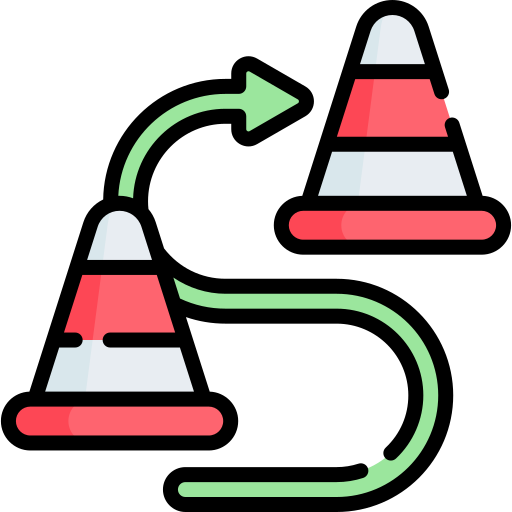} & \includegraphics[width=0.025\linewidth]{figures/icons/behavior.png} & \includegraphics[width=0.025\linewidth]{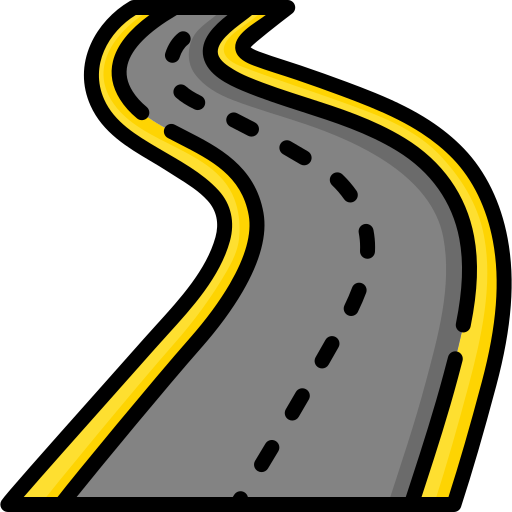} & \includegraphics[width=0.025\linewidth]{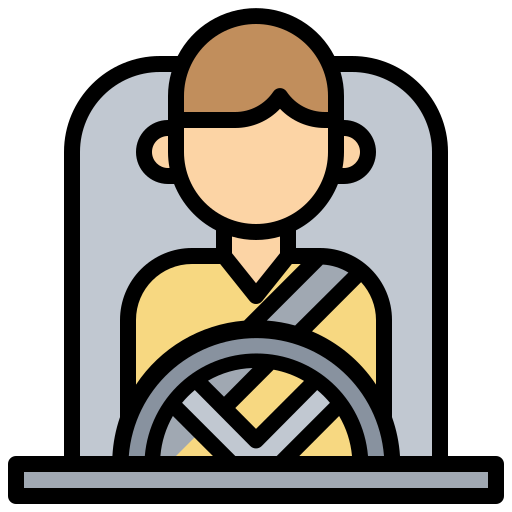} & \textbf{\# Frames} & \textbf{\# QA Pairs} & \multirow{2}{*}{\textbf{Logic}} & \multirow{2}{*}{\textbf{Evaluation Metrics}}
    \\
    & \textbf{Per.} & \textbf{Pre.} & \textbf{Beh.} & \textbf{Pla.} & \textbf{Rob.} & \textcolor{gray}{(Test)} & \textcolor{gray}{(Test)} &
    \\\midrule\midrule
    BDD-X \cite{kim2018textual} & \textcolor{robo_green}{\cmark} & \textcolor{robo_red}{\xmark} & \textcolor{robo_red}{\xmark} & \textcolor{robo_red}{\xmark} & \textcolor{robo_red}{\xmark} & - & - & None & Language
    \\
    BDD-OIA \cite{xu2020explainable} & \textcolor{robo_green}{\cmark} & \textcolor{robo_red}{\xmark} & \textcolor{robo_green}{\cmark} & \textcolor{robo_red}{\xmark} & \textcolor{robo_red}{\xmark} & - & - & None & F1 Score
    \\
    nuScenes-QA \cite{qian2024nuscenes} & \textcolor{robo_green}{\cmark} & \textcolor{robo_red}{\xmark} & \textcolor{robo_red}{\xmark} & \textcolor{robo_red}{\xmark} & \textcolor{robo_red}{\xmark} & $36,114$ & $83,337$ & None & Acc
    \\
    Talk2Car \cite{deruyttere2019talk2car} & \textcolor{robo_green}{\cmark} & \textcolor{robo_red}{\xmark} & \textcolor{robo_red}{\xmark} & \textcolor{robo_green}{\cmark} & \textcolor{robo_red}{\xmark} & $\sim 1.8$k & $2,447$ & None & Acc
    \\
    nuPrompt \cite{wu2023language} & \textcolor{robo_green}{\cmark} & \textcolor{robo_red}{\xmark} & \textcolor{robo_red}{\xmark} & \textcolor{robo_red}{\xmark} & \textcolor{robo_red}{\xmark} & $\sim 36$k & $\sim 6$k & None & AMOTA
    \\
    DRAMA \cite{malla2023drama} & \textcolor{robo_green}{\cmark} & \textcolor{robo_red}{\xmark} & \textcolor{robo_red}{\xmark} & \textcolor{robo_green}{\cmark} & \textcolor{robo_red}{\xmark} & - & $\sim 14$k & Chain & Language
    \\
    Rank2Tel \cite{sachdeva2024rank2tell} & \textcolor{robo_green}{\cmark} & \textcolor{robo_red}{\xmark} & \textcolor{robo_red}{\xmark} & \textcolor{robo_green}{\cmark} & \textcolor{robo_red}{\xmark} & - & - & Chain & Acc, Language
    \\
    DirveMLLM~\cite{guo2024drivemllm} & \textcolor{robo_green}{\cmark} & \textcolor{robo_red}{\xmark} & \textcolor{robo_red}{\xmark} & \textcolor{robo_red}{\xmark} & \textcolor{robo_red}{\xmark} & $880$ & - & None & Acc
    \\
    DriveVLM~\cite{tian2024drivevlm} & \textcolor{robo_green}{\cmark} & \textcolor{robo_red}{\xmark} & \textcolor{robo_green}{\cmark} & \textcolor{robo_green}{\cmark} & \textcolor{robo_red}{\xmark} & - & - & None & GPT$_{\text{ctx}}$
    \\
    DriveLM \cite{sima2023drivelm} & \textcolor{robo_green}{\cmark} & \textcolor{robo_green}{\cmark} & \textcolor{robo_green}{\cmark} & \textcolor{robo_green}{\cmark} & \textcolor{robo_red}{\xmark} & $4,794$ & $15,480$ & Graph & Language, GPT
    \\\midrule
    \textsf{\textcolor{robo_blue}{Drive}\textcolor{robo_red}{Bench}} & \textcolor{robo_green}{\cmark} & \textcolor{robo_green}{\cmark} & \textcolor{robo_green}{\cmark} & \textcolor{robo_green}{\cmark} & \textcolor{robo_green}{\cmark} & $\mathbf{19,200}$ & $\mathbf{20,498}$ & \textbf{Graph} & \textbf{Acc}, \textbf{Language}, \textbf{GPT}, \textbf{GPT}$_{\text{ctx}}$
    \\\bottomrule
\end{tabular}}
\end{table*}

\begin{figure*}[t]
    \centering
    \includegraphics[width=\linewidth]{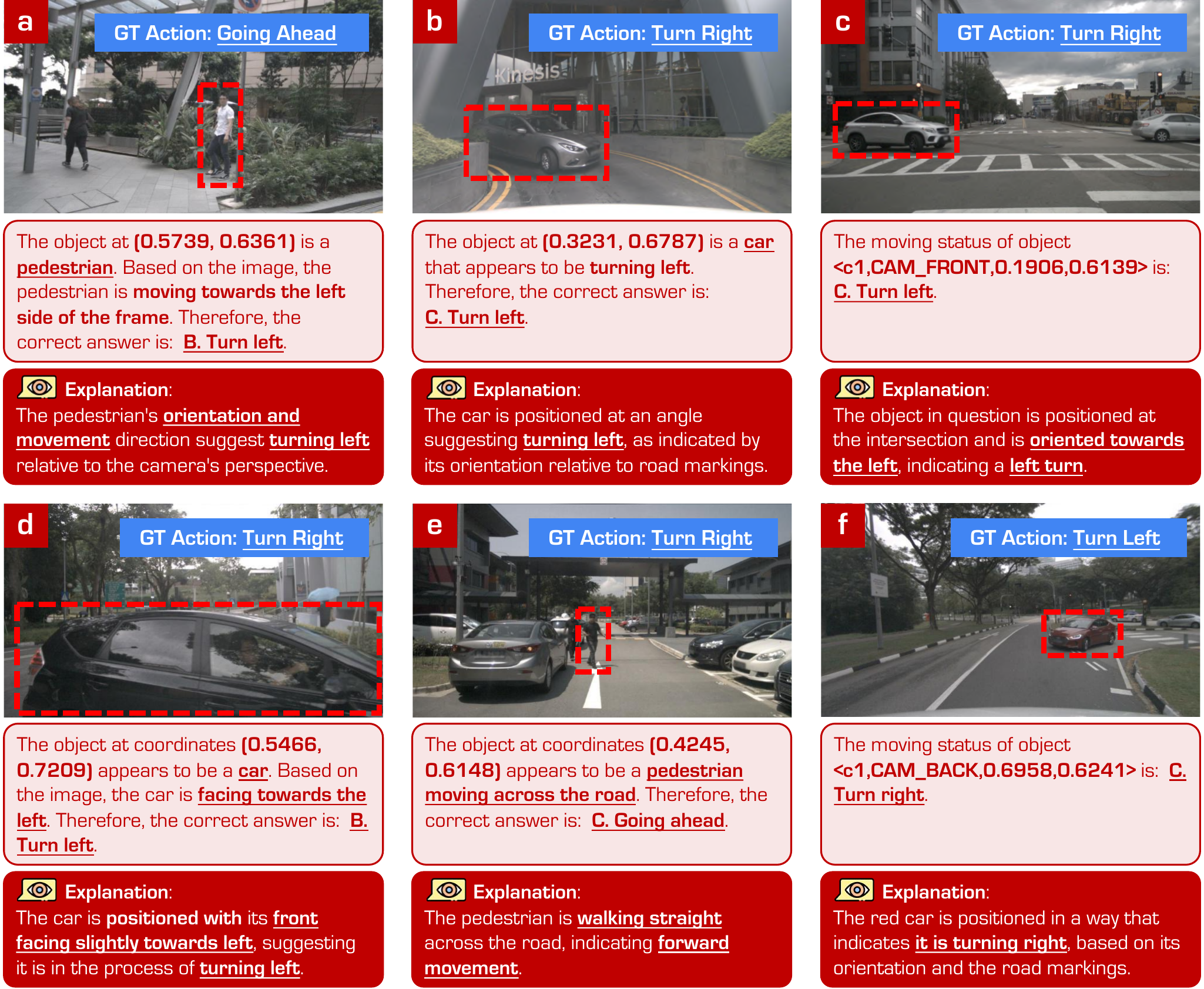}
    \vspace{-0.6cm}
    \caption{\textbf{GPT-4o failure cases}. \textbf{(a)}: GPT-4o reasons the moving status of the pedestrian by the moving position related to the frame, instead of the coordinate of the moving object itself, thus leading to wrong perception results. \textbf{(b)}: The model struggles to distinguish the correct direction based on the coordinate of the target object. \textbf{(c)}: GPT-4o reasons the moving status of the SUV by the relative location of the object to the current frame causes the wrong perception results. \textbf{(d)}: GPT-4o fails to perceive the orientation of the car. \textbf{(e)}: The dataset contains examples that need multiple frames to reason successfully, GPT-4o fails to address these examples with a single image input. \textbf{(f)}: GPT-4o reasons the moving status of the SUV by the relative location of the object to the current frame causes the wrong perception results.}
    \label{fig:gpt-failure-case}
\end{figure*}

\noindent{\textbf{Challenging Cases.}} Furthermore, we evaluate the advanced GPT-4o \cite{achiam2023gpt} and analyze the failure cases, as illustrated in \cref{fig:gpt-fail-case}. We find that annotations such as \textit{Turn Left} or \textit{Turn Right} are factually accurate but often require temporal context or subtle indicators (\eg, turn signal lights) to be correctly interpreted. Additionally, vehicles overlap with each other in some cases, making the pixel position-based question too nuanced for current VLMs to distinguish (\textit{e.g.}, \textit{What's the object at (540, 600) and object at (530, 610), respectively}). As a result, existing VLMs demonstrate considerable difficulty in accurately interpreting cues that rely on temporal information or subtle visual indicators, leading to a disproportionate number of negative outcomes in evaluations. To prevent such samples from skewing our findings, we eliminate instances highly dependent on temporal context or present significant interpretive challenges for current VLMs. Our data selection strategy prioritizes instances that GPT-4o \cite{achiam2023gpt} can correctly interpret, thereby indicating the availability of sufficient visual cues for single-frame-based predictions. We also analyze other failure cases of GPT-4o~\cite{achiam2023gpt}, as shown in \cref{fig:gpt-failure-case}. In these examples, humans can answer with correct answers while GPT-4o~\cite{achiam2023gpt} heavily relies on the object relation position to the frames for decision-making. Thus, we keep those examples in our dataset.

\begin{figure}
    \centering
    \includegraphics[width=\linewidth]{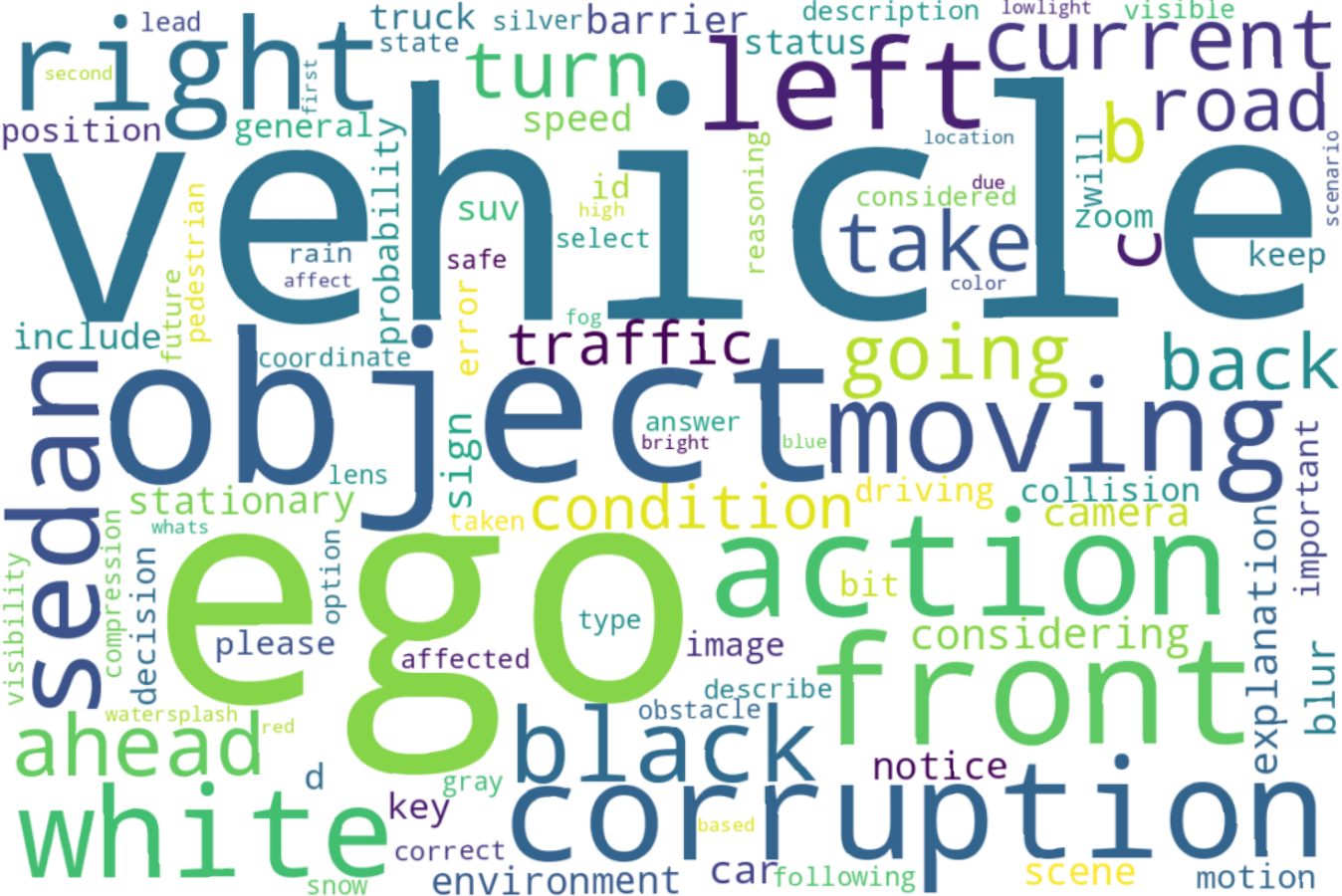}
    \vspace{-0.6cm}
    \caption{The \textbf{word cloud} collected from the QA pairs in the proposed benchmark, highlighting the main focus on different autonomous driving tasks in \textsf{\textcolor{robo_blue}{Drive}\textcolor{robo_red}{Bench}}. The larger the font size, the higher the frequency of occurrence.}
    \label{fig:word-cloud}
\end{figure}

\subsection{Driving Tasks}
Our \textsf{\textcolor{robo_blue}{Drive}\textcolor{robo_red}{Bench}} covers four mainstream driving tasks, including \includegraphics[width=0.04\linewidth]{figures/icons/perception.png}~\textbf{perception}, \includegraphics[width=0.04\linewidth]{figures/icons/prediction.png}~\textbf{prediction}, \includegraphics[width=0.04\linewidth]{figures/icons/planning.png}~\textbf{planning}, and \includegraphics[width=0.04\linewidth]{figures/icons/behavior.png}~\textbf{behavior}, examples are  shown in \cref{fig:teaser-bench}. 

The \textbf{perception} task focuses on identifying the moving status of targeted objects and analyzing the surrounding environments. The \textbf{prediction} task focuses on predicting the future movement of these objects. The \textbf{planning} task focuses on suggesting actions to navigate the vehicle safely and efficiently in complex driving scenarios. Finally, the \textbf{behavior} task is designed to predict the future steering and speed of the ego vehicles. By addressing these tasks, the dataset ensures comprehensive coverage of the critical aspects required for evaluating the capabilities of driving VLMs in handling complex and dynamic scenarios.

\begin{figure}[t]
    \centering
    \includegraphics[width=\linewidth]{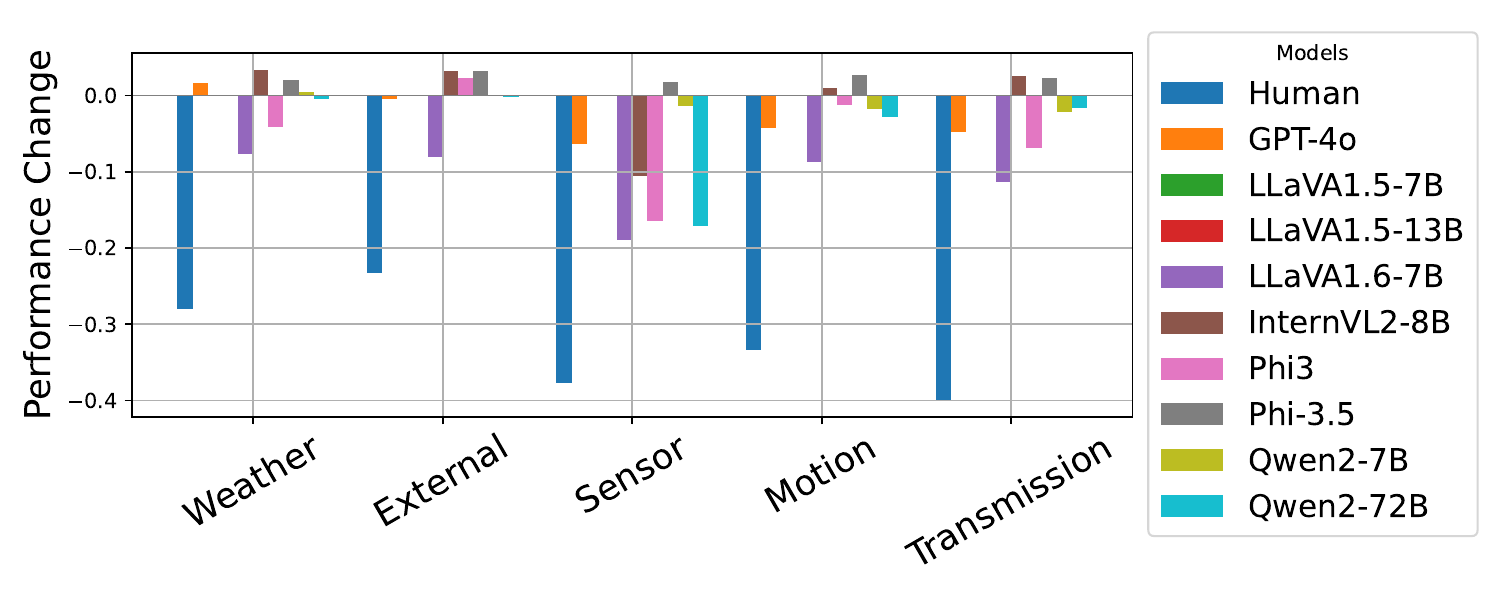}
    \vspace{-0.7cm}
    \caption{\textbf{Illustration of performance degradation}. After applying each corruption, we evaluate the MCQs accuracy changes compared with clean inputs. We observe that human performance largely decreases while most VLMs remain unchanged.}
    \label{fig:acc-degradation}
\end{figure}

\begin{table*}[t]
    \centering
    \caption{\textbf{Evaluations of VLMs across different driving tasks} (perception, prediction, planning, and behavior). ``\textcolor{robo_green}{Clean}'' represents clean image inputs. ``\textcolor{robo_red}{Corr.}'' represents corruption image inputs, averaged across fifteen corruptions. ``\textcolor{robo_blue}{T.O.}'' represents text-only evaluation. For humans, we only evaluate MCQ questions in perception and behavior tasks. The evaluations are based on GPT scores, where we tailored detailed rubrics for each task and question type. We \colorbox{robo_red!10}{highlight} scores higher than clean performance under corruption.}
    \vspace{-0.2cm}
    \label{tab:benchmark}
    \resizebox{\linewidth}{!}{
    \begin{tabular}{r|r|c|ccc|ccc|ccc|ccc}
    \toprule
    \multirow{2}{*}{\textbf{Method}} & \multirow{2}{*}{\textbf{Size}} & \multirow{2}{*}{\textbf{Type}} & \multicolumn{3}{c|}{\includegraphics[width=0.024\linewidth]{figures/icons/perception.png}~\textbf{Perception}} & \multicolumn{3}{c|}{\includegraphics[width=0.024\linewidth]{figures/icons/prediction.png}~\textbf{Prediction}} & \multicolumn{3}{c|}{\includegraphics[width=0.024\linewidth]{figures/icons/planning.png}~\textbf{Planning}} & \multicolumn{3}{c}{\includegraphics[width=0.024\linewidth]{figures/icons/behavior.png}~\textbf{Behavior}}
    \\
    & & & \textcolor{robo_green}{{Clean}} & \textcolor{robo_red}{{Corr.}} & \textcolor{robo_blue}{{T.O.}} & \textcolor{robo_green}{{Clean}} & \textcolor{robo_red}{{Corr.}} & \textcolor{robo_blue}{{T.O.}} & \textcolor{robo_green}{{Clean}} & \textcolor{robo_red}{{Corr.}} & \textcolor{robo_blue}{{T.O.}} & \textcolor{robo_green}{{Clean}} & \textcolor{robo_red}{{Corr.}} & \textcolor{robo_blue}{{T.O.}}
    \\\midrule\midrule
    \cellcolor{robo_green!10}\includegraphics[width=0.024\linewidth]{figures/icons/human.png}~\textcolor{robo_green}{Human} & \cellcolor{robo_green!10}\textcolor{robo_green}{-} & \cellcolor{robo_green!10}\textcolor{robo_green}{-} & \cellcolor{robo_green!10}\textcolor{robo_green}{$47.67$} & \textcolor{robo_red}{$38.32$} \cellcolor{robo_green!10} & \cellcolor{robo_green!10}\textcolor{robo_green}{-} & \cellcolor{robo_green!10}\textcolor{robo_green}{-} & \cellcolor{robo_green!10}\textcolor{robo_green}{-} & \cellcolor{robo_green!10}\textcolor{robo_green}{-} & \cellcolor{robo_green!10}\textcolor{robo_green}{-} & \cellcolor{robo_green!10}\textcolor{robo_green}{-} & \cellcolor{robo_green!10}\textcolor{robo_green}{-} & \cellcolor{robo_green!10} \textcolor{robo_green}{$69.51$} & \cellcolor{robo_green!10} \textcolor{robo_red}{$54.09$} & \cellcolor{robo_green!10}\textcolor{robo_green}{-}
    \\\midrule
    \textcolor{gray}{GPT-4o~\cite{achiam2023gpt}} & - & \textcolor{gray}{Commercial} & \textcolor{gray}{$35.37$} & \textcolor{gray}{$35.25$} & \cellcolor{robo_red!10}\textcolor{gray}{$36.48 $} & \textcolor{gray}{$51.30$} & \textcolor{gray}{$49.94$} & \textcolor{gray}{$49.05$} & \textcolor{gray}{$75.75$} & \textcolor{gray}{$75.36$} & \textcolor{gray}{$73.21$} & \textcolor{gray}{$45.40$} & \textcolor{gray}{$44.33$} & \cellcolor{robo_red!10}\textcolor{gray}{$50.03$} 
    \\\midrule
    LLaVA-1.5 \cite{liu2024llava1.5} & $7$ B & Open & $23.22$ & $22.95$ & $22.31$ & $22.02$ & $17.54$ & $14.64$ & $29.15$ & \cellcolor{robo_red!10}$31.51$ & \cellcolor{robo_red!10}$32.45$ & $13.60$ & \cellcolor{robo_red!10}$13.62$ \cellcolor{robo_red!10}& $14.91$
    \\
    LLaVA-1.5 \cite{liu2024llava1.5} & $13$ B & Open & $23.35$ & \cellcolor{robo_red!10}$23.37$ & $22.37$ & $36.98$ & \cellcolor{robo_red!10}$37.78$ & $23.98$ & $34.26$ & \cellcolor{robo_red!10}$34.99$ & \cellcolor{robo_red!10}$38.85$ & $32.99$ & $32.43$ & $32.79$
    \\
    LLaVA-NeXT \cite{liu2024llavanext} & $7$ B & Open & $24.15$ & $19.62$ & $13.86$ & $35.07$ & \cellcolor{robo_red!10}$35.89$ & $28.36$ & $45.27$ & $44.36$ & $27.58$ & $48.16$ & $39.44$ & $11.92$
    \\
    InternVL2 \cite{chen2023internvl} & $8$ B & Open & $\mathbf{32.36}$ & \cellcolor{robo_red!10}$32.68$ & \cellcolor{robo_red!10}$33.60$ & $45.52$ & $37.93$ & \cellcolor{robo_red!10}$48.89$ & $53.27$ & \cellcolor{robo_red!10}$55.25$ & $34.56$ &  $\mathbf{54.58}$ & $40.78$ & $20.14$ 
    \\
    Phi-3 \cite{abdin2024phi} & $4.2$ B & Open & $22.88$ & \cellcolor{robo_red!10}$23.93$ & \cellcolor{robo_red!10}$28.26$ & $40.11$ & $37.27$ & $22.61$ & $\underline{60.03}$ & \cellcolor{robo_red!10}$61.31$ & $46.88$ & $45.20$ & $44.57$ & $28.22$
    \\
    Phi-3.5 \cite{abdin2024phi} & $4.2$ B & Open & $27.52$ & $27.51$ & \cellcolor{robo_red!10}$28.26$ & $45.13$ & $38.21$ & $4.92$ & $31.91$ & $28.36$ & \cellcolor{robo_red!10}$46.30$ & $37.89$ & \cellcolor{robo_red!10}$49.13$ \cellcolor{robo_red!10}& $39.16$
    \\
    Oryx \cite{liu2024oryx} & $7$ B & Open & $17.02$ & \cellcolor{robo_red!10}$15.97$ & \cellcolor{robo_red!10}$18.47$ & $\underline{48.13}$ & $46.63$ & $12.77$ & $53.57$ & \cellcolor{robo_red!10}$55.76$ & $48.26$ & $33.92$ & $33.81$ & $23.94$
    \\
    Qwen2-VL \cite{wang2024qwen2} & $7$ B & Open & $28.99$ & $27.85$ & \cellcolor{robo_red!10}$35.16$ & $37.89$ & \cellcolor{robo_red!10}$39.55$ & $37.77$ & $57.04$ & $54.78$ & $41.66$ & $49.07$ & \cellcolor{robo_red!10}$47.68$ & \cellcolor{robo_red!10}$54.48$
    \\
    Qwen2-VL \cite{wang2024qwen2} & $72$ B & Open & $\underline{30.13}$ & $26.92$ & $17.70$ & $\mathbf{49.35} $ & $43.49$ & $5.57$ & $\mathbf{61.30}$ & \cellcolor{robo_red!10}$63.07$ & $53.35$ & $\underline{51.26}$ & $49.78$ & $39.46$
    \\
    \midrule
    DriveLM \cite{sima2023drivelm} & $7$ B & Specialist & $\mathbf{16.85}$ & $16.00$ & $8.75$ & $\mathbf{44.33}$ & $39.71$ & $4.70$ & $\mathbf{68.71}$ & $67.60$ & $65.24$ & $\mathbf{42.78}$ & $40.37$ & $27.83$ 
    \\
    Dolphins \cite{ma2023dolphins} & $7$ B & Specialist & $9.59$ & \cellcolor{robo_red!10}$10.84$ & \cellcolor{robo_red!10}$11.01$ & $32.66$ & $29.88$ & \cellcolor{robo_red!10}$39.98$ & $52.91$ & \cellcolor{robo_red!10}$53.77$ & \cellcolor{robo_red!10}$60.98$ & $8.81$ & $8.25$ & $11.92$
    \\
    \bottomrule
\end{tabular}}
\end{table*}

\subsection{Corruption Data}
Autonomous driving is an application where numerous types of corruption can exist \cite{kong2024robodepth, kong2023robo3d, xie2024benchmarking, kong2024robodrive, hao2024mapbench, li2024is}. Previous works find that task-specific models are inherently vulnerable to OoD corruptions, which prohibit precise perception of surroundings. On the other hand, large models \cite{nguyen2022quality, fang2022data} are shown to be resilient toward OoD corruption, given their vast amount of training data \cite{radford2021learning}. Therefore, given the safety-critical applications and promising robustness of large models, one natural question is, how reliable are existing VLMs towards visual corruption in driving? 

Our \textsf{\textcolor{robo_blue}{Drive}\textcolor{robo_red}{Bench}} crafts a total of $\mathbf{15}$ different corruption types (\cf \cref{fig:teaser-bench}), spanning across \includegraphics[width=0.04\linewidth]{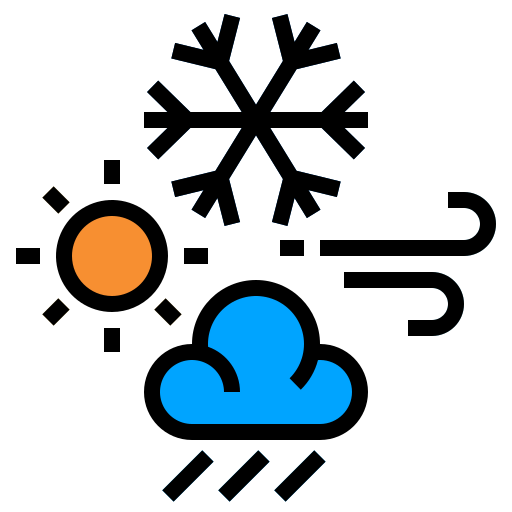}~\textbf{weather conditions} ($^1$\texttt{Brightness}, $^2$\texttt{Dark}, $^3$\texttt{Fog}, $^4$\texttt{Snow}, and $^5$\texttt{Rain}), \includegraphics[width=0.04\linewidth]{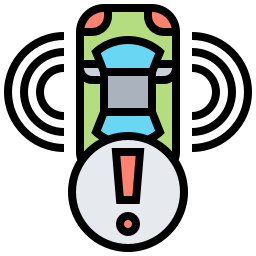}~\textbf{external disturbances} ($^6$\texttt{Water Splash} and $^7$\texttt{Lens Obstacle}), \includegraphics[width=0.04\linewidth]{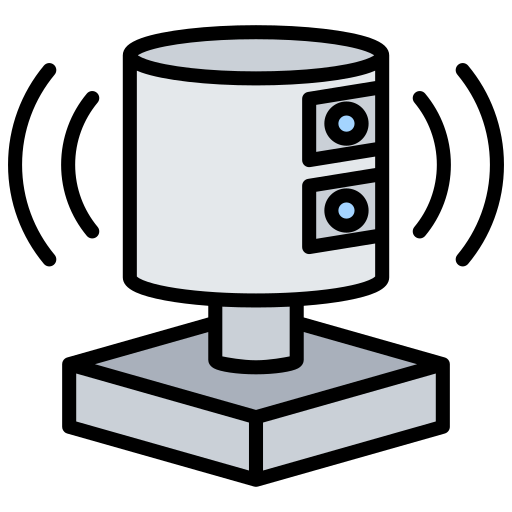}~\textbf{sensor failures} ($^8$\texttt{Camera Crash}, $^9$\texttt{Frame Lost}, and $^{10}$\texttt{Saturate}), \includegraphics[width=0.04\linewidth]{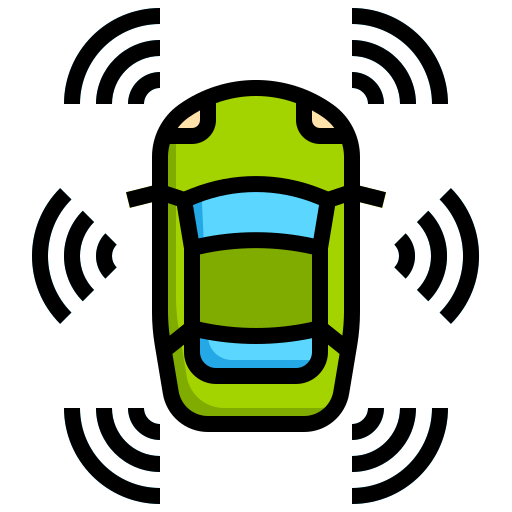}~\textbf{motion blurs} ($^{11}$\texttt{Motion Blur} and $^{12}$\texttt{Zoom Blur}), and \includegraphics[width=0.04\linewidth]{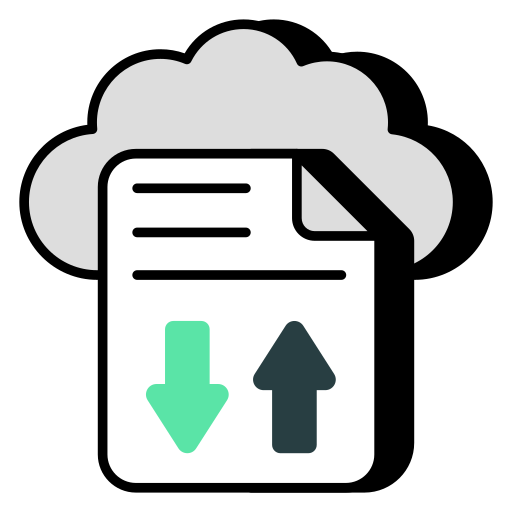}~\textbf{data transmission errors} ($^{13}$\texttt{Bit Error}, $^{14}$\texttt{Color Quant}, and $^{15}$\texttt{H.265 Compression}). The corruption encompasses a range of potential OoD scenarios the vehicles might encounter. From a reliability perspective, These corruptions are the key to our evaluation and insights into VLMs' visual-grounded driving capabilities. For more detailed definitions and examples, please refer to \cref{sec:bench_construction}.

It should be noted that while existing research has explored VLM hallucinations and trustworthiness~\cite{ji2023survey, wang2023evaluation, li2023evaluating, tu2023many}, it has not yet been rigorously examined within the context of driving applications. Autonomous driving requires approaching the reliability of VLMs from different aspects, as language-based driving decisions are naturally linked to physical and context-specific real-world scenarios.

\subsection{Vision-Language Models (VLMs)}
\label{sec:bench-model}
Recent advancements in VLMs applied to autonomous driving mainly include: 1) leveraging VLMs to generate and interpret driving decisions in high-level language~\cite{ma2023dolphins, tian2024drivevlm, jiang2024senna}, and 2) employing VLMs for end-to-end autonomous driving in low-level commands in open-loop~\cite{sima2023drivelm, xu2024drivegpt4, mao2023gpt, tian2024drivevlm, wang2024omnidrive, jiang2024senna} or closed-loop~\cite{shao2024lmdrive} settings. A fundamental motivation and underlying assumption across this domain is that VLMs can generate interpretable and explainable responses, thereby reducing the opaque ``black-box" nature inherent in traditional task-specific models~\cite{li2022bevformer, hu2023planning}. 

To encompass the full scope of existing advanced VLMs, the current version of \textsf{\textcolor{robo_blue}{Drive}\textcolor{robo_red}{Bench}} evaluates a diverse set of $\mathbf{12}$ popular VLMs, including both commercial and open-source models, as well as models fine-tuned specifically for autonomous driving applications~\cite{ma2023dolphins, sima2023drivelm}. This selection reflects the latest developments in state-of-the-art VLMs for driving. To ensure consistency, we apply a standardized system prompt across all models (further details are provided in the Appendix~\ref{sec:vlm-prompt}). The prompt explicitly instructs the VLMs to generate auxiliary explanations, enabling GPT-based evaluation of MCQs, which only have a selection alone by default, as detailed in the next section.

\subsection{Evaluation Metrics}
In \textsf{\textcolor{robo_blue}{Drive}\textcolor{robo_red}{Bench}}, we consider several metrics following~\cite{sima2023drivelm}, including accuracy, BLEU~\cite{papineni2002bleu}, ROUGE-L~\cite{lin2004rouge}, and GPT score~\cite{sima2023drivelm, chen2024driving}. For MCQs, we utilize both accuracy, as the most direct measure, and GPT scores to capture nuances in the explanatory quality beyond simple answer selection. We employ GPT-3.5-turbo for GPT Score evaluation. To better capture subtleties between responses, we prompt the model with detailed rubrics that account for answer correctness, coherence, and the alignment of explanations with the final answer. Rubrics are adapted for each specific task and question type to better reflect human-preferred responses. Detailed information on the evaluation prompts and rubrics can be found in \cref{sec:gpt_evaluations}.
\section{Experiments}
\label{sec:experiments}

We conduct extensive benchmark experiments and analyses in \textsf{\textcolor{robo_blue}{Drive}\textcolor{robo_red}{Bench}}, with detailed discussions leading to our observations and conclusions by step.

\subsection{Experimental Setups}

We set the temperature to $0.2$ and \texttt{top-p} to $0.2$, with a maximum output token limit of $512$. For DriveLM-Agent \cite{sima2023drivelm}, we adhere to the configurations outlined in \cite{contributors2023drivelmrepo}. Specifically, we utilize LLaMA-Adapter-V2 \cite{gao2023llama} as the base model, fine-tuned on the DriveLM-nuScenes dataset. The fine-tuning process is conducted on eight A800 GPUs with a batch size of $4$, over $4$ epochs. For other open-source models, we download the official model weight from HuggingFace and inference using the \texttt{vLLM}~\cite{kwon2023efficient} framework. More details about the used model weight can be found in Appendix~\ref{sec:public-resource}. For GPT-4o, we query the official APIs from OpenAI with the same configuration mentioned above.

\subsection{Observations \& Discussions}
\label{sec:discussion}

\subsubsection{Corruption Resilience}
\label{sec:exp-corrupt}

\begin{figure}[t]
    \centering
    \includegraphics[width=\linewidth]{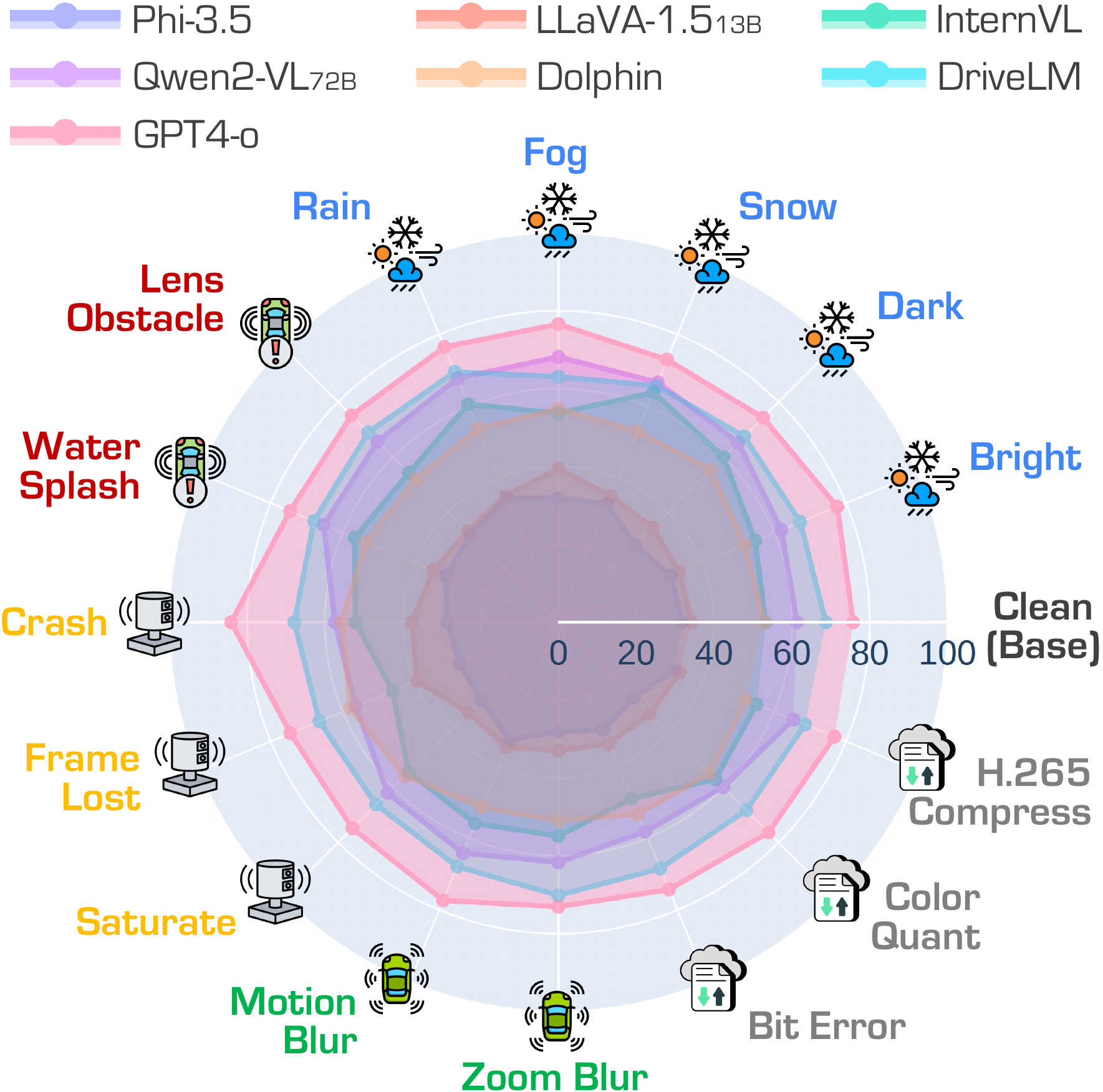}
    \vspace{-0.6cm}
    \caption{\textbf{Radar chart comparisons among different models}. The performance for each input corruption type is averaged across all the $1,261$ questions spanning four different driving tasks. The evaluation metric used here is the GPT score.
    }
    \label{fig:all-radar}
\end{figure}

The primary results, evaluated using GPT, are summarized in \cref{tab:benchmark} and \cref{fig:all-radar}. These findings reveal that, even in the presence of image corruption, the model performance remains largely unaffected, demonstrating notable resilience to such perturbations. Specifically, as illustrated in \cref{fig:all-radar}, the performance trends align closely with those observed in other benchmarks. For instance, GPT-4o and Qwen2-VL$_{\text{72B}}$ consistently achieve state-of-the-art results. Furthermore, when evaluating the resilience against corruption, most vision-language models (VLMs) maintain comparable performance to that observed with clean image inputs, even in open-ended visual question answering (VQA) tasks. To understand the source of this resilience, we investigate whether it stems from the inherent robustness of these VLMs, due to their pre-training on extensive web-scale datasets, or if other factors contribute to this phenomenon.

\begin{table*}[t]
    \centering
    \caption{\textbf{Comparisons of accuracy scores between ``clean'' and fully ``black'' (no image) inputs}. We observe a large portion of models have no clear performance degradation even when the visual information is absent, suggesting the VLMs response might mainly be based on majority biases (\eg, \textit{Going Ahead}, in most driving scenarios), instead of leveraging visual cues from sensors.}
    \vspace{-0.2cm}
    \label{tab:noimage}
    \resizebox{\linewidth}{!}{
    \begin{tabular}{l|c|c|c|c|c|c|c|c|c}
        \toprule
        \textbf{Task} & \textbf{Image} & \textcolor{robo_green}{Human} &  \textcolor{gray}{GPT-4o}~\cite{achiam2023gpt} & LLaVA-NeXT~\cite{liu2024llavanext} & LLaVA-1.5$_\text{13B}$~\cite{liu2024llava1.5} & Phi-3~\cite{abdin2024phi} & Phi-3.5~\cite{abdin2024phi} & Qwen2-VL$_\text{7B}$~\cite{wang2024qwen2} & Qwen2-VL$_\text{72B}$~\cite{wang2024qwen2} 
        \\ 
        \midrule\midrule
        \multirow{2}{*}{\textbf{Perception}} & \multicolumn{1}{c|}{Clean} & \cellcolor{robo_green!10}\textcolor{robo_green}{$93.3$} & 
        \textcolor{gray}{$59.0$} & $55.0$ & $50.0$ & $54.5$ & $56.5$ & $59.0$ & $60.0$ 
        \\
        & \multicolumn{1}{c|}{No Image} & \cellcolor{robo_green!10}\textcolor{robo_green}{-} & 
        \cellcolor{gray!20}\textcolor{gray}{$59.5$} \textcolor{red}{\scriptsize$\uparrow 0.5$} & 
        \cellcolor{gray!20} $34.5$ \textcolor{black}{\scriptsize$\downarrow 20.5$} & 
        \cellcolor{gray!20} $50.0$ \textcolor{black}{\scriptsize$\downarrow 0.0$}  & 
        \cellcolor{gray!20} $17.5$ \textcolor{black}{\scriptsize$\downarrow 37.0$} & 
        \cellcolor{gray!20} $58.5$ \textcolor{red}{\scriptsize$\uparrow 2.0$} & 
        \cellcolor{gray!20} $56.5$ \textcolor{black}{\scriptsize$\downarrow 2.5$} & 
        \cellcolor{gray!20} $23.5$ \textcolor{black}{\scriptsize$\downarrow 36.5$} 
        \\
        \midrule
        \multirow{2}{*}{\textbf{Behavior}} & \multicolumn{1}{c|}{Clean} & \cellcolor{robo_green!10}\textcolor{robo_green}{$69.5$} & 
        \textcolor{gray}{$25.5$} & $33.5$ & $32.5$ & $26.5$ & $36.5$ & $30.0$ & $23.0$ 
        \\
        & \multicolumn{1}{c|}{No Image} & \cellcolor{robo_green!10}\textcolor{robo_green}{-} & 
        \cellcolor{gray!20}\textcolor{gray}{$24.0$} \textcolor{black}{\scriptsize$\downarrow 1.5$} & 
        \cellcolor{gray!20}$24.0$ \textcolor{black}{\scriptsize$\downarrow 9.5$} & 
        \cellcolor{gray!20}$33.0$ \textcolor{red}{\scriptsize$\uparrow 0.5$} & 
        \cellcolor{gray!20}$30.0$ \textcolor{red}{\scriptsize$\uparrow 3.5$} & 
        \cellcolor{gray!20}$40.0$ \textcolor{red}{\scriptsize$\uparrow 3.5$} & 
        \cellcolor{gray!20}$23.0$ \textcolor{black}{\scriptsize$\downarrow 7.0$} & 
        \cellcolor{gray!20}$36.5$ \textcolor{red}{\scriptsize$\uparrow 13.5$} 
        \\
        \bottomrule
    \end{tabular}}
\end{table*}
\begin{table*}[t]
\centering
\caption{\textbf{Comparisons of accuracy changes before and after prompting VLMs with explicit corruption context}. We notice a clear trend of performance degradation after mentioning the corruption type in the question. The results suggest VLMs are aware of the current corruption and acknowledge they can not respond to the severely degraded visual information when explicitly prompted.}
\vspace{-0.2cm}
\label{tab:corruption-context}
\resizebox{\linewidth}{!}{
\begin{tabular}{r|ccccc|cc|ccc|cc|ccc}
    \toprule
    \textbf{Method} & \textbf{Bright} & \textbf{Dark} & \textbf{Snow} & \textbf{Fog} & \textbf{Rain} & \textbf{Lens} & \textbf{Water} & \textbf{Cam} & \textbf{Frame} & \textbf{Saturate} & \textbf{Motion} & \textbf{Zoom} & \textbf{Bit} & \textbf{Quant} & \textbf{H.265} 
    \\
    \midrule\midrule
    \textcolor{gray}{GPT-4o} &
    \cellcolor{robo_red!50}{\cellcolor{robo_red!19}\textcolor{gray}{$-8.69$}} & \cellcolor{robo_red!10}{\cellcolor{robo_red!24}\textcolor{gray}{$-12.98$}} & \cellcolor{robo_red!10}{\cellcolor{robo_red!19}\textcolor{gray}{$-8.25$}} & \cellcolor{robo_red!10}{\cellcolor{robo_red!20}\textcolor{gray}{$-9.00$}} & \cellcolor{robo_red!10}{\cellcolor{robo_red!16}\textcolor{gray}{$-6.00$}} & \cellcolor{robo_red!10}{\cellcolor{robo_red!14}\textcolor{gray}{$-3.81$}} & \cellcolor{robo_red!10}{\cellcolor{robo_red!16}\textcolor{gray}{$-5.82$}} & \cellcolor{robo_red!10}{\cellcolor{robo_red!24}\textcolor{gray}{$-12.94$}} & \cellcolor{robo_red!10}{\cellcolor{robo_red!22}\textcolor{gray}{$-10.99$}} & \cellcolor{robo_red!10}{\cellcolor{robo_red!19}\textcolor{gray}{$-8.52$}} & \cellcolor{robo_red!10}{\cellcolor{robo_red!17}\textcolor{gray}{$-6.98$}} & \textcolor{gray}{$0.57$} & \cellcolor{robo_red!10}{\cellcolor{robo_red!19}\textcolor{gray}{$-8.22$}} & \cellcolor{robo_red!10}{\cellcolor{robo_red!15}\textcolor{gray}{$-4.79$}} & \cellcolor{robo_red!10}{\cellcolor{robo_red!25}\textcolor{gray}{$-14.30$}} 
    \\\midrule
    LLaVA-1.5$_\text{7B}$ &
    $0.26$ & $1.04$ & $0.25$ & $0.00$ & $0.00$ & $1.40$ & $2.60$ & \cellcolor{robo_red!10}{\cellcolor{robo_red!13}$-2.79$} & \cellcolor{robo_red!10}{\cellcolor{robo_red!20}$-8.97$} & $0.51$ & \cellcolor{robo_red!10}{\cellcolor{robo_red!10}$-0.52$} & $2.57$ & $2.22$ & \cellcolor{robo_red!10}{\cellcolor{robo_red!11}$-1.32$} & \cellcolor{robo_red!10}{\cellcolor{robo_red!13}$-2.66$} 
    \\
    LLaVA-1.5$_\text{13B}$ &
    $0.26$ & $1.04$ & $0.25$ & $0.00$ & $0.00$ & $1.96$ & $2.60$ & \cellcolor{robo_red!10}{\cellcolor{robo_red!11}$-1.27$} & \cellcolor{robo_red!10}{\cellcolor{robo_red!10}$-0.26$} & $0.51$ & $1.04$ & $2.57$ & $2.22$ & \cellcolor{robo_red!10}{\cellcolor{robo_red!10}$-0.26$} & \cellcolor{robo_red!10}{\cellcolor{robo_red!12}$-2.07$} 
    \\
    LLaVA-NeXT &
    \cellcolor{robo_red!10}{\cellcolor{robo_red!16}$-5.83$} & \cellcolor{robo_red!10}{\textbf{\cellcolor{robo_red!32}$-20.63$}} & \cellcolor{robo_red!10}{\textbf{\cellcolor{robo_red!45}$-31.95$}} & \cellcolor{robo_red!10}{\textbf{\cellcolor{robo_red!25}$-14.00$}} & \cellcolor{robo_red!10}{\textbf{\cellcolor{robo_red!30}$-18.50$}} & \cellcolor{robo_red!10}{\textbf{\cellcolor{robo_red!44}$-31.39$}} & \cellcolor{robo_red!10}{\textbf{\cellcolor{robo_red!50}$-36.97$}} & \cellcolor{robo_red!10}{\cellcolor{robo_red!16}$-6.13$} & \cellcolor{robo_red!10}{\cellcolor{robo_red!30}$-18.29$} & \cellcolor{robo_red!10}{\textbf{\cellcolor{robo_red!29}$-17.67$}} & \cellcolor{robo_red!10}{\textbf{\cellcolor{robo_red!37}$-24.85$}} & \cellcolor{robo_red!10}{\textbf{\cellcolor{robo_red!46}$-33.29$}} & \cellcolor{robo_red!10}{\textbf{\cellcolor{robo_red!31}$-19.50$}} & $$5.89$$ & \cellcolor{robo_red!10}{\textbf{\cellcolor{robo_red!33}$-21.19$}} 
    \\
    InternVL$_\text{8B}$ &
    \cellcolor{robo_red!10}{\cellcolor{robo_red!18}$-7.24$} & \cellcolor{robo_red!10}{\cellcolor{robo_red!19}$-8.92$} & \cellcolor{robo_red!10}{\cellcolor{robo_red!21}$-10.74$} & \cellcolor{robo_red!10}{\cellcolor{robo_red!20}$-9.50$} & \cellcolor{robo_red!10}{\cellcolor{robo_red!18}$-7.50$} & \cellcolor{robo_red!10}{\cellcolor{robo_red!18}$-7.54$} & \cellcolor{robo_red!10}{\cellcolor{robo_red!17}$-6.24$} & \cellcolor{robo_red!10}{\cellcolor{robo_red!29}$-17.51$} & \cellcolor{robo_red!10}{\cellcolor{robo_red!10}$-0.23$} & \cellcolor{robo_red!10}{\cellcolor{robo_red!12}$-2.46$} & \cellcolor{robo_red!10}{\cellcolor{robo_red!12}$-2.35$} & \cellcolor{robo_red!10}{\cellcolor{robo_red!17}$-7.00$} & \cellcolor{robo_red!10}{\cellcolor{robo_red!17}$-6.67$} & \cellcolor{robo_red!10}{\cellcolor{robo_red!18}$-7.71$} & \cellcolor{robo_red!10}{\cellcolor{robo_red!15}$-4.65$} 
    \\
    Phi-3.5 &
    \cellcolor{robo_red!10}{\textbf{\cellcolor{robo_red!20}$-9.78$}} & \cellcolor{robo_red!10}{\cellcolor{robo_red!18}$-7.48$} & \cellcolor{robo_red!10}{\cellcolor{robo_red!18}$-7.75$} & \cellcolor{robo_red!10}{\cellcolor{robo_red!20}$-9.00$} & \cellcolor{robo_red!10}{\cellcolor{robo_red!19}$-8.50$} & \cellcolor{robo_red!10}{\cellcolor{robo_red!19}$-8.60$} & \cellcolor{robo_red!10}{\cellcolor{robo_red!18}$-7.48$} & \cellcolor{robo_red!10}{\cellcolor{robo_red!28}$-16.37$} & \cellcolor{robo_red!10}{\cellcolor{robo_red!20}$-9.31$} & \cellcolor{robo_red!10}{\cellcolor{robo_red!20}$-9.50$} & \cellcolor{robo_red!10}{\cellcolor{robo_red!19}$-8.48$} & \cellcolor{robo_red!10}{\cellcolor{robo_red!19}$-8.07$} & \cellcolor{robo_red!10}{\cellcolor{robo_red!17}$-6.94$} & \cellcolor{robo_red!10}{\cellcolor{robo_red!22}$-11.29$} & \cellcolor{robo_red!10}{\cellcolor{robo_red!22}$-11.16$} 
    \\
    Phi-3 &
    \cellcolor{robo_red!10}{\cellcolor{robo_red!14}$$-4.22$$} & $8.67$ & $0.75$ & \cellcolor{robo_red!10}{\cellcolor{robo_red!15}$-5.00$} & \cellcolor{robo_red!10}{\cellcolor{robo_red!21}$-10.00$} & \cellcolor{robo_red!10}{\cellcolor{robo_red!22}$-11.31$} & \cellcolor{robo_red!10}{\cellcolor{robo_red!46}$-33.22$} & $3.03$ & $8.29$ & \cellcolor{robo_red!10}{\cellcolor{robo_red!19}$-8.51$} & \cellcolor{robo_red!10}{\cellcolor{robo_red!16}$-5.42$} & $3.57$ & $17.89$ & \cellcolor{robo_red!10}{\textbf{\cellcolor{robo_red!30}$-18.81$}} & \cellcolor{robo_red!10}{\cellcolor{robo_red!24}$$-13.12$$} 
    \\
    Qwen2-VL$_\text{7B}$ &
    \cellcolor{robo_red!10}{\cellcolor{robo_red!20}$-9.74$} & \cellcolor{robo_red!10}{\cellcolor{robo_red!18}$-7.96$} & \cellcolor{robo_red!10}{\cellcolor{robo_red!20}$-9.75$} & \cellcolor{robo_red!10}{\cellcolor{robo_red!20}$-9.50$} & \cellcolor{robo_red!10}{\cellcolor{robo_red!20}$-9.00$} & \cellcolor{robo_red!10}{\cellcolor{robo_red!16}$-5.93$} & \cellcolor{robo_red!10}{\cellcolor{robo_red!17}$-6.98$} & \cellcolor{robo_red!10}{\textbf{\cellcolor{robo_red!33}$-20.94$}} & \cellcolor{robo_red!10}{\textbf{\cellcolor{robo_red!42}$-29.85$}} & \cellcolor{robo_red!10}{\cellcolor{robo_red!19}$-8.49$} & \cellcolor{robo_red!10}{\cellcolor{robo_red!19}$-8.46$} & \cellcolor{robo_red!10}{\cellcolor{robo_red!13}$-3.00$} & \cellcolor{robo_red!10}{\cellcolor{robo_red!15}$-5.06$} & \cellcolor{robo_red!10}{\cellcolor{robo_red!20}$-9.38$} & \cellcolor{robo_red!10}{\cellcolor{robo_red!22}$-11.07$} 
    \\
    Qwen2-VL$_\text{72B}$ &
    \cellcolor{robo_red!10}{\cellcolor{robo_red!17}$-6.70$} & \cellcolor{robo_red!10}{\cellcolor{robo_red!20}$-8.96$} & \cellcolor{robo_red!10}{\cellcolor{robo_red!19}$-8.25$} & \cellcolor{robo_red!10}{\cellcolor{robo_red!20}$-9.50$} & \cellcolor{robo_red!10}{\cellcolor{robo_red!22}$-11.00$} & \cellcolor{robo_red!10}{\cellcolor{robo_red!19}$-8.04$} & \cellcolor{robo_red!10}{\cellcolor{robo_red!17}$-6.90$} & $7.19$ & $11.01$ & \cellcolor{robo_red!10}{\cellcolor{robo_red!21}$-10.51$} & \cellcolor{robo_red!10}{\cellcolor{robo_red!18}$-7.44$} & \cellcolor{robo_red!10}{\cellcolor{robo_red!13}$-2.93$} & \cellcolor{robo_red!10}{\cellcolor{robo_red!17}$-6.61$} & \cellcolor{robo_red!10}{\cellcolor{robo_red!20}$-9.29$} & \cellcolor{robo_red!10}{\cellcolor{robo_red!24}$-13.07$} 
    \\
\bottomrule
\end{tabular}}
\end{table*}

\noindent\textbf{Human Evaluations.} To further validate that the applied corruptions indeed impact the driving scenario and to explore the performance gap between humans and VLMs, we conduct a human evaluation. Specifically, we sub-sample the dataset and design a user interface to facilitate human performance assessment (more details in Appendix~\ref{sec:human-eval}).

The results are shown in \cref{tab:benchmark}. Since human responses include only a choice without a time-intensive written explanation, GPT scores, which we design to reward detailed explanations, may introduce unfairness in comparison. Therefore, we also report accuracy degradation, as shown in \cref{fig:acc-degradation}. Interestingly, we observe a significant accuracy drop for human participants under corrupted conditions, whereas most VLMs exhibit subtle performance variations across different corruption types. We next explore if it is due to model robustness \cite{fang2022data} or other possible reasons.

\noindent\textbf{Text-Only Prompts.} Given the above results, we further investigate the effects of extreme corruption by providing VLMs with fully black images, reducing the input to text-only prompts with no visual information. The results, shown in \cref{tab:benchmark}, reveals an intriguing pattern: GPT scores for text-only prompts are closely aligned with those obtained with clean image inputs. This trend persists across different tasks and models, suggesting that the phenomenon is not solely due to model robustness.

Considering the GPT score also takes the quality of explanations into account in addition to answer correctness, we further analyze accuracy on MCQs to isolate the potential scoring advantages due to explanations (discussed in \cref{sec:bench-model}). Given that MCQs in perception tasks have three answer choices and MCQs in behavior tasks have four, the expected random-guessing accuracy would be approximately $33\%$ and $25\%$, respectively. We report results only for models that exceed this random baseline on clean inputs, as shown in \cref{tab:noimage}. Interestingly, a significant portion of the models show minimal or no accuracy degradation, even in the absence of visual cues. The results are even more concerning when considering open-ended questions (\ie, prediction and planning). For instance, responses from the state-of-the-art GPT-4o \cite{achiam2023gpt} under text-only conditions maintain approximately $95\%$ of the performance seen with clean image prompts. 

\begin{figure}
    \centering
    \includegraphics[width=\linewidth]{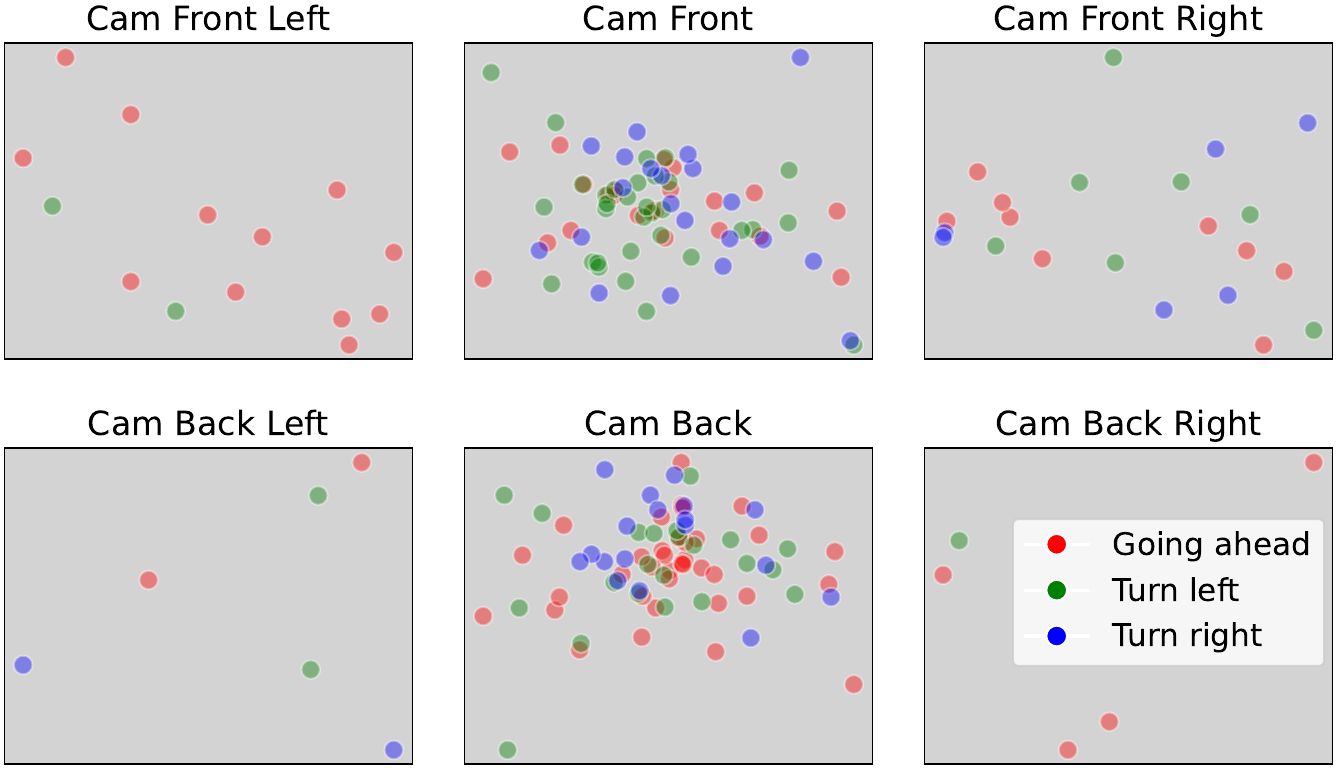}
    \vspace{-0.6cm}
    \caption{\textbf{Prediction spatial distribution} of Qwen2-VL$_\text{7B}$ \cite{wang2024qwen2} under text-only prompts. The model can potentially ``guess'' the MCQ answers without visual information by leveraging text cues, \eg, camera and coordinate positions in the questions.}
    \label{fig:cam-distribution}
\end{figure}

Upon further examination shown in \cref{fig:cam-distribution}, we observe that the high performance of VLMs under text-only conditions is likely influenced by the extensive general knowledge acquired during training. For instance, the models can guess the moving status of one surrounding object based on text cues referring to which camera it has been seen and the corresponding position in that image (examples can be found in \cref{fig:teaser}). These observations yield two key insights: 

\begin{itemize}
    \item VLMs are capable of producing plausible responses to driving-related questions based solely on natural language prompts. This capability is likely attributed to the extensive general knowledge and common sense reasoning capabilities developed during their pre-training.
    \item The current evaluation protocols for assessing VLMs in autonomous driving reveal significant shortcomings. Even advanced evaluation methods, such as those leveraging GPT-based scoring, fail to capture the nuances of the reliability of VLMs' responses.
\end{itemize}

To investigate the first insight further, we pose the question: \textit{``Are driving VLMs aware of the underlying corruptions in images when they fabricate their answers?''} This question serves as the basis for the expanded analysis presented below.

\begin{table*}[t]
    \centering
    \caption{\textbf{Evaluations on corruption awareness}. ``\textcolor{robo_blue}{MCQ}" represents questions that directly ask about the current corruption types. ``\textcolor{robo_red}{VQA}'' represents questions in perception, prediction, and planning but augmented with explicit corruption context information, averaged across three driving tasks. ``\textcolor{robo_green}{CAP}'' represents captioning questions that ask detailed descriptions at both object-level and corruption-level.}
    \vspace{-0.2cm}
    \label{tab:benchmark_robustness}
    \resizebox{\linewidth}{!}{
    \begin{tabular}{r|ccc|ccc|ccc|ccc|ccc}
    \toprule
    \multirow{2}{*}{\textbf{Method}} & \multicolumn{3}{c|}{\includegraphics[width=0.024\linewidth]{figures/icons/weather.png}~\textbf{Weather}} & \multicolumn{3}{c|}{\includegraphics[width=0.024\linewidth]{figures/icons/external.png}~\textbf{External}}& \multicolumn{3}{c|}{\includegraphics[width=0.024\linewidth]{figures/icons/sensor.png}~\textbf{Sensor}} & \multicolumn{3}{c|}{\includegraphics[width=0.024\linewidth]{figures/icons/motion.png}~\textbf{Motion}} & \multicolumn{3}{c}{\includegraphics[width=0.024\linewidth]{figures/icons/transmission.png}~\textbf{Transmission}}
    \\
    & \textcolor{robo_blue}{{MCQ}} & \textcolor{robo_red}{{VQA}} & \textcolor{robo_green}{{CAP}} & \textcolor{robo_blue}{{MCQ}} & \textcolor{robo_red}{{VQA}} & \textcolor{robo_green}{{CAP}} & \textcolor{robo_blue}{{MCQ}} & \textcolor{robo_red}{{VQA}} & \textcolor{robo_green}{{CAP}} & \textcolor{robo_blue}{{MCQ}} & \textcolor{robo_red}{{VQA}} & \textcolor{robo_green}{{CAP}} & \textcolor{robo_blue}{{MCQ}} & \textcolor{robo_red}{{VQA}} & \textcolor{robo_green}{{CAP}}
    \\\midrule\midrule
    \textcolor{gray}{GPT-4o~\cite{achiam2023gpt}} & \textcolor{gray}{$57.20$} & \textcolor{gray}{$57.28$} & \textcolor{gray}{$54.90$} & \textcolor{gray}{$29.25$} & \textcolor{gray}{$56.60$} & \textcolor{gray}{$61.98$} & \textcolor{gray}{$44.25$} & \textcolor{gray}{$54.95$} & \textcolor{gray}{$56.53$} & \textcolor{gray}{$34.25$} & \textcolor{gray}{$59.20$} & \textcolor{gray}{$56.25$} & \textcolor{gray}{$36.83$} & \textcolor{gray}{$53.95$} & \textcolor{gray}{$57.57$}
    \\\midrule
    \rowcolor{gray!10} LLaVA-1.5$_\text{7B}$~\cite{liu2024llava1.5} & $\underline{69.70}$ & $35.49$ & $35.91$ & $26.50$ & $29.17$ & $34.95$ & $18.83$ & $30.64$ & $33.15$ & $71.25$ & $33.43$ & $35.18$ & $10.17$ & $27.28$ & $34.38$
    \\
    LLaVA-1.5$_\text{13B}$~\cite{liu2024llava1.5} & $61.60$ & $39.76$ & $37.76$ & $15.50$ & $34.55$ & $37.83$ & $24.08$ & $35.48$ & $36.08$ & $\mathbf{79.75}$ & $36.46$ & $36.42$ & $15.50$ & $32.53$ & $34.33$
    \\
    \rowcolor{gray!10} LLaVA-NeXT \cite{liu2024llavanext} & $69.70$ & $36.96$ & $48.52$ & $\underline{48.50}$ & $30.32$ & $57.18$ & $21.83$ & $30.40$ & $44.37$ & $66.00$ & $34.20$ & $\mathbf{50.44}$ & $11.83$ & $29.43$ & $53.50$
    \\
    InternVL2~\cite{chen2023internvl} & $59.90$ & $48.72$ & $48.60$ & $\mathbf{50.75}$ & $\underline{47.74}$ & $\mathbf{57.82}$ & $29.92$ & $45.06$ & $51.14$ & $68.25$ & $49.51$ & $\underline{49.67}$ & $30.00$ & $\underline{43.42}$ & $\mathbf{54.24}$
    \\
    \rowcolor{gray!10} Phi-3 \cite{abdin2024phi} & $40.00$ & $40.59$ & $45.61$ & $25.00$ & $31.44$ & $45.99$ & $16.83$ & $35.58$ & $43.71$ & $31.25$ & $42.92$ & $48.43$ & $27.67$ & $33.04$ & $41.35$
    \\
    Phi-3.5 \cite{abdin2024phi} & $60.60$ & $41.82$ & $45.97$ & $21.25$ & $36.89$ & $30.95$ & $25.58$ & $34.66$ & $39.30$ & $33.00$ & $46.03$ & $49.33$ & $\underline{39.67}$ & $33.47$ & $39.67$
    \\
    \rowcolor{gray!10} Oryx \cite{liu2024oryx} & $53.20$ & $40.43$ & $\mathbf{48.95}$ & $45.00$ & $40.68$ & $56.06$ & $\underline{50.50}$ & $36.71$ & $48.55$ & $\underline{72.50}$ & $40.01$ & $48.33$ & $\underline{39.67}$ & $36.98$ & $49.87$ 
    \\
    Qwen2-VL$_\text{7B}$ \cite{wang2024qwen2} & $\mathbf{76.70}$ & $\underline{49.33}$ & $45.12$ & $37.50$ & $47.62$ & $51.24$ & $22.83$ & $\underline{39.45}$ & $\underline{47.23}$ & $57.00$ & $\underline{47.40}$ & $47.74$ & $35.83$ & $42.31$ & $48.60$
    \\
    \rowcolor{gray!10} Qwen2-VL$_\text{72B}$ \cite{wang2024qwen2} & $59.80$ & $\mathbf{51.05}$ & $\underline{48.55}$ & $45.50$ & $\mathbf{50.57}$ & $\underline{57.25}$ & $\mathbf{52.25}$ & $\mathbf{45.89}$ & $\mathbf{48.59}$ & $58.25$ & $\mathbf{50.85}$ & $47.88$ & $\mathbf{44.83}$ & $\mathbf{46.23}$ & $\underline{50.50}$
    \\
    \midrule
    DriveLM \cite{sima2023drivelm} & $21.20$ & $\mathbf{42.86}$ & $20.04$ & $\mathbf{21.25}$ & $\mathbf{37.49}$ & $21.92$ & $9.00$ & $\mathbf{36.68}$ & $15.56$ & $\mathbf{22.25}$ & $\mathbf{42.05}$ & $17.07$ & $17.50$ & $\mathbf{39.56}$ & $10.37$
    \\
    \rowcolor{gray!10}Dolphins \cite{ma2023dolphins} & $\mathbf{54.30}$ & $30.21$ & $\mathbf{31.08}$ & $3.00$ & $30.42$ & $\mathbf{29.38}$ & $\mathbf{9.42}$ & $26.83$ & $\mathbf{26.30}$ & $9.25$ & $29.82$ & $\mathbf{28.05}$ & $\mathbf{21.50}$ & $28.86$ & $\mathbf{27.65}$
    \\
    \bottomrule
\end{tabular}}
\end{table*}

\begin{figure}
    \centering
    \includegraphics[width=\linewidth]{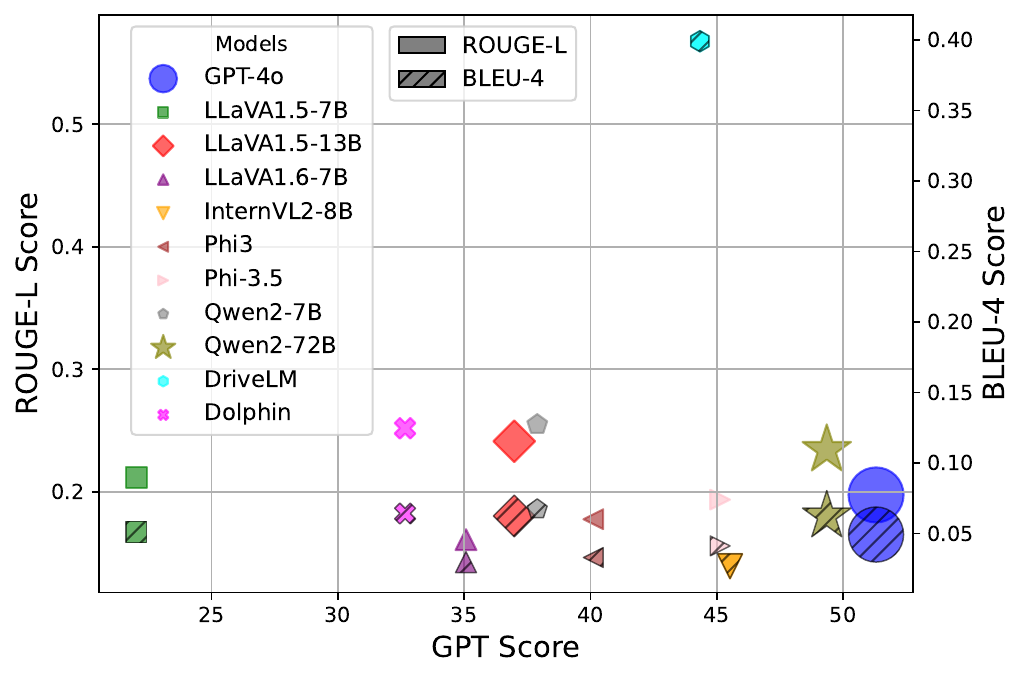}
    \vspace{-0.7cm}
    \caption{\textbf{Evaluation results when using different metrics}. The language metrics, such as ROUGE-L \cite{lin2004rouge} and BLEU-4 \cite{papineni2002bleu}, exhibit high consistency; while the GPT score demonstrates noticeable gaps. We also observe that fine-tuned process benefits DriveLM \cite{sima2023drivelm, gao2023llama} significantly in regulating its response format, thus leading to misleading high performance under language metrics.}
    \label{fig:metric}
\end{figure}

\begin{figure}[t]
    \centering
    \begin{subfigure}[b]{0.49\linewidth}
        \centering
        \includegraphics[width=\linewidth]{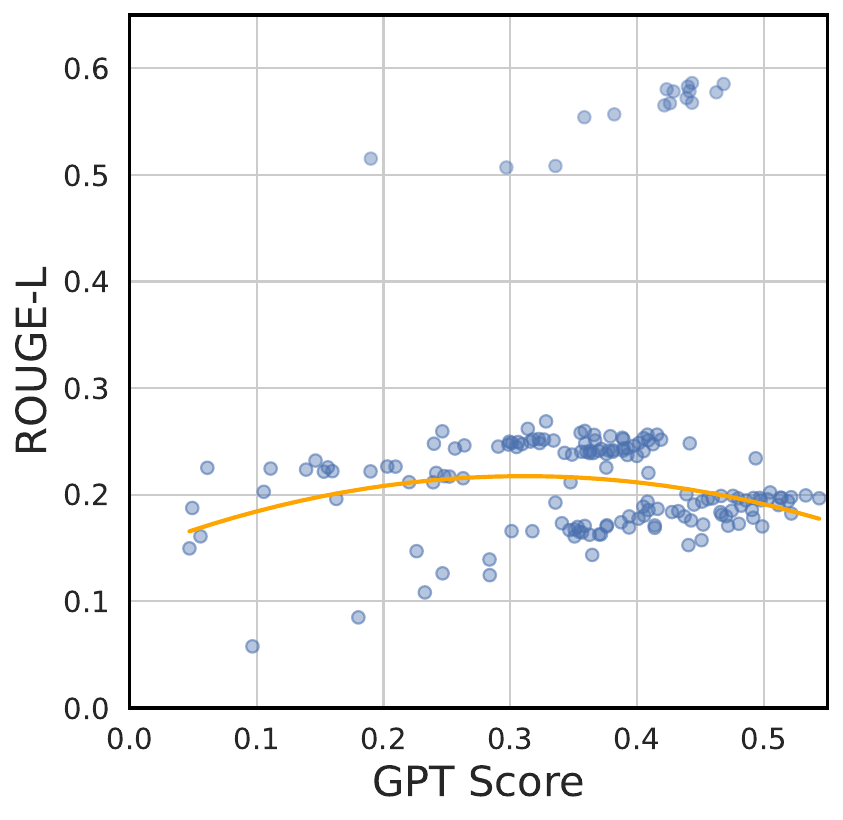}
        \caption{Open-Ended Questions}
        \label{fig:metric-scatter}
    \end{subfigure}
    \hfill
    \begin{subfigure}[b]{0.49\linewidth}
        \centering
        \includegraphics[width=\linewidth]{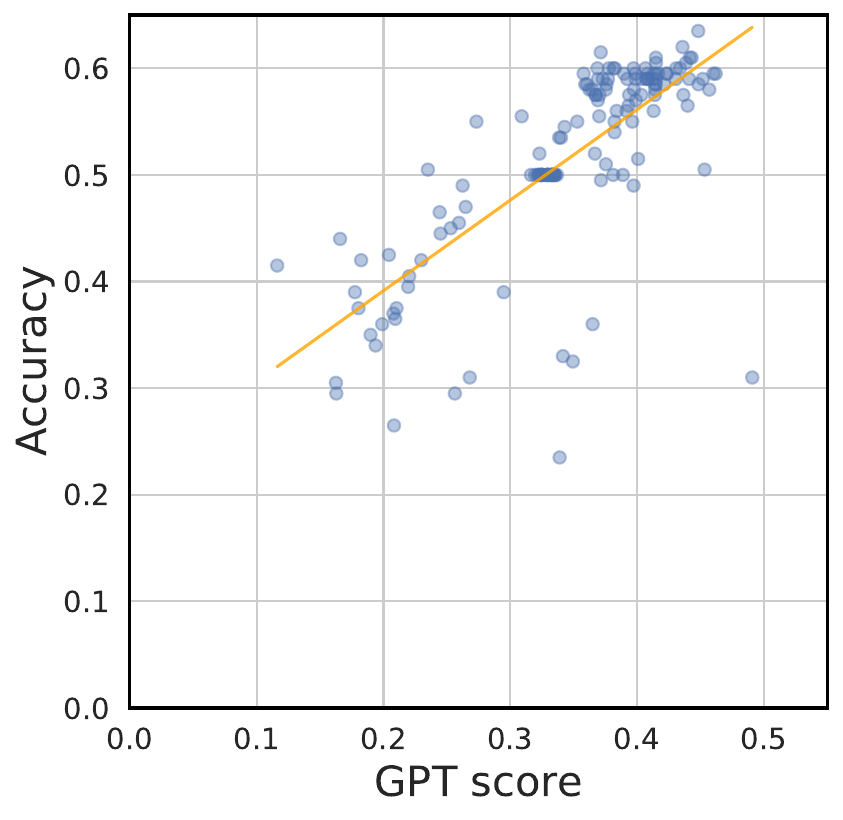}
        \caption{Multiple Choice Questions}
        \label{fig:metric-mcq}
    \end{subfigure}
    \vspace{-0.15cm}
    \caption{\textbf{Correlations when using different metrics}. We study how well accuracy or ROUGE-L \cite{lin2004rouge} matches the GPT scores for open-ended questions and multiple-choice questions (MCQs), respectively. We find that ROUGE-L \cite{lin2004rouge} fails to reflect semantic information (\eg, key object) that is critical in driving. Contrarily, accuracy aligns well with the GPT score for MCQ while the GPT score can further capture nuanced differences in explanation when the answer is correct.}
    \label{fig:merged-metrics-scatter}
\end{figure}

\begin{figure*}[t]
    \centering
    \begin{subfigure}[b]{0.32\linewidth}
        \centering
        \includegraphics[width=\linewidth]{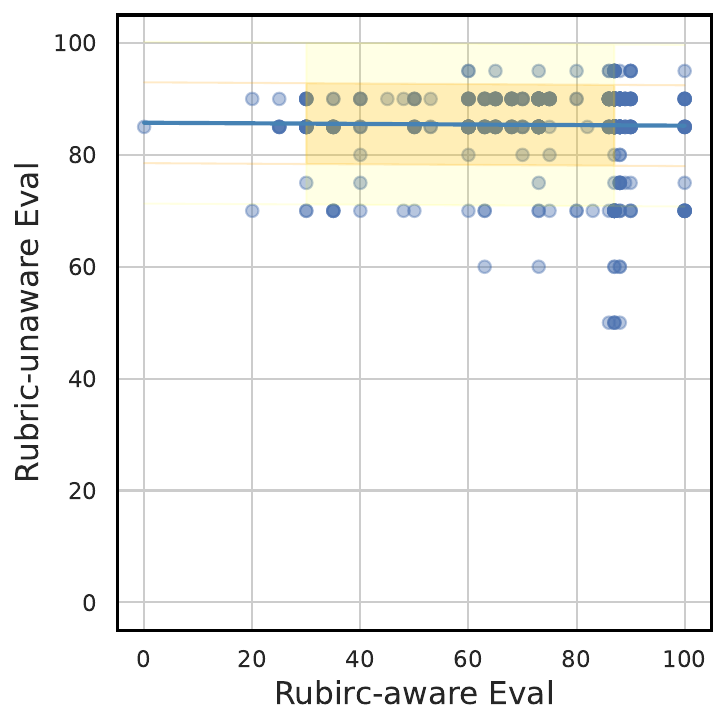}
        \caption{Rubric-Aware GPT Evaluation}
        \label{fig:metric-rubric}
    \end{subfigure}
    ~
    \begin{subfigure}[b]{0.32\linewidth}
        \centering
        \includegraphics[width=\linewidth]{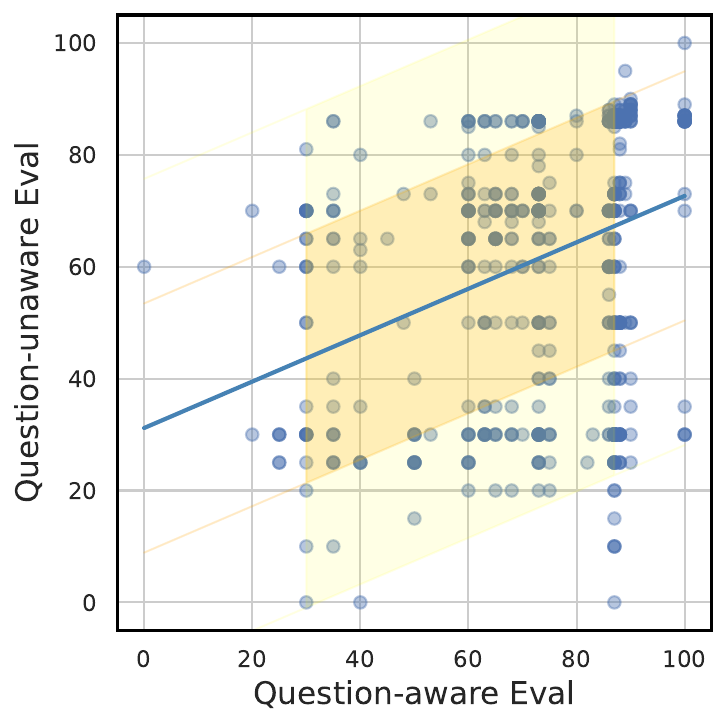}
        \caption{Question-Aware GPT Evaluation}
        \label{fig:metric-question}
    \end{subfigure}
    ~
    \begin{subfigure}[b]{0.32\linewidth}
        \centering
        \includegraphics[width=\linewidth]{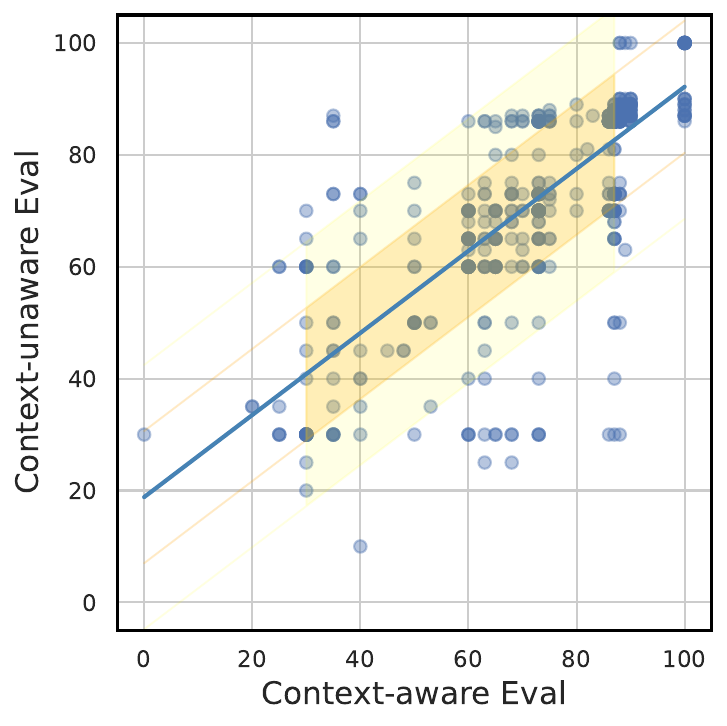}
        \caption{Context-Aware GPT Evaluation}
        \label{fig:metric-context}
    \end{subfigure}
    \vspace{-0.2cm}
    \caption{\textbf{Comparisons among different evaluation types (rubric, question-aware, and context-aware)}. The GPT scores vary depending on the rubric, question, and physical driving context. With more information added, the results become more distinguishable.}
    \label{fig:metric-gpt}
\end{figure*}

\begin{figure*}[t]
    \centering
    \includegraphics[width=\linewidth]{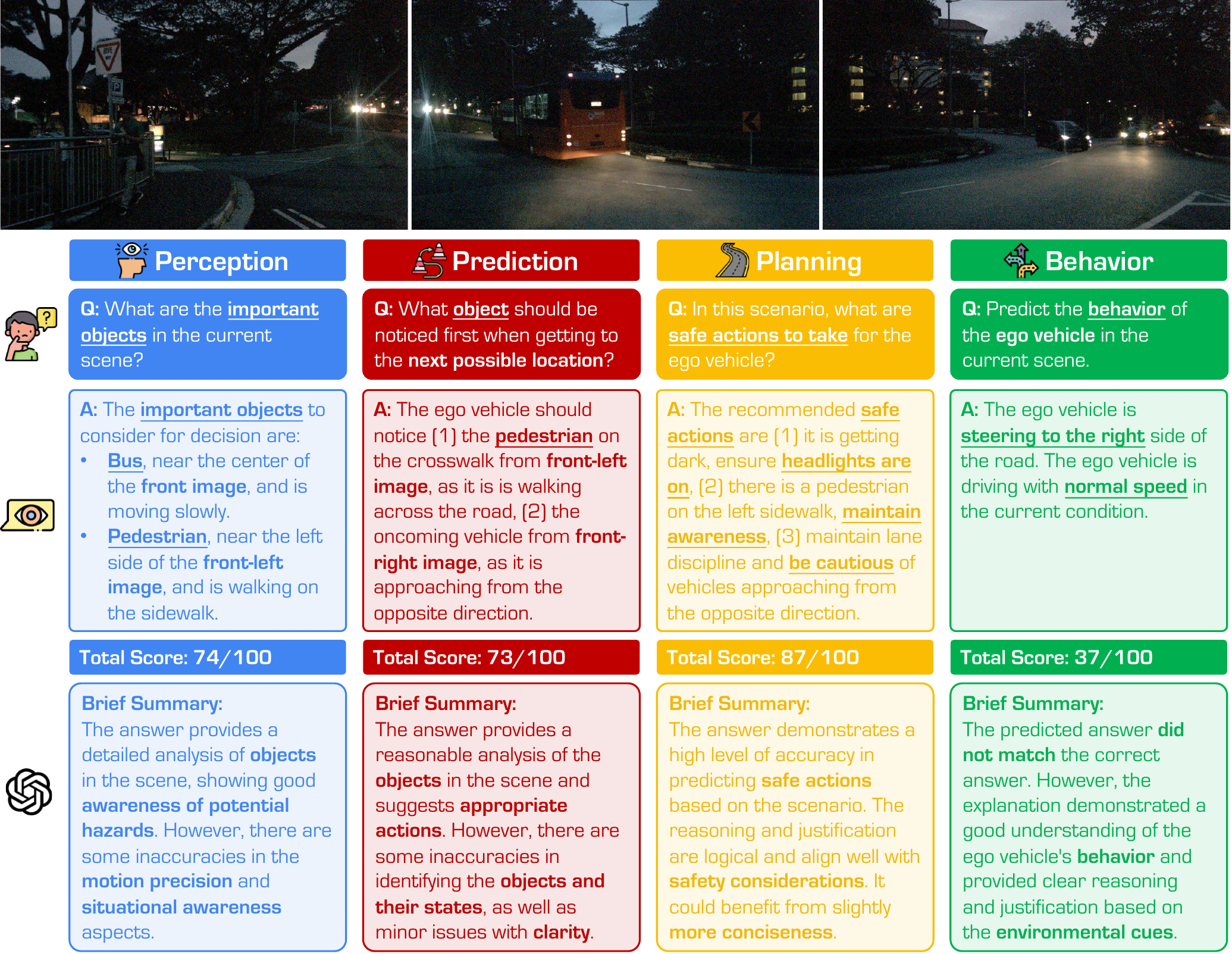}
    \vspace{-0.6cm}
    \caption{Examples of GPT-4o~\cite{achiam2023gpt} responses to four tasks and the corresponding evaluation results under the \textcolor{robo_red}{\textbf{dark condition}}. We observe that GPT-4o~\cite{achiam2023gpt} is aware of the low-light environment and can identify the \textit{bus} and \textit{pedestrian} from the image, showing resilience.}
    \label{fig:examples_benchmark_1}
\end{figure*}

\begin{figure*}[t]
    \centering
    \includegraphics[width=\linewidth]{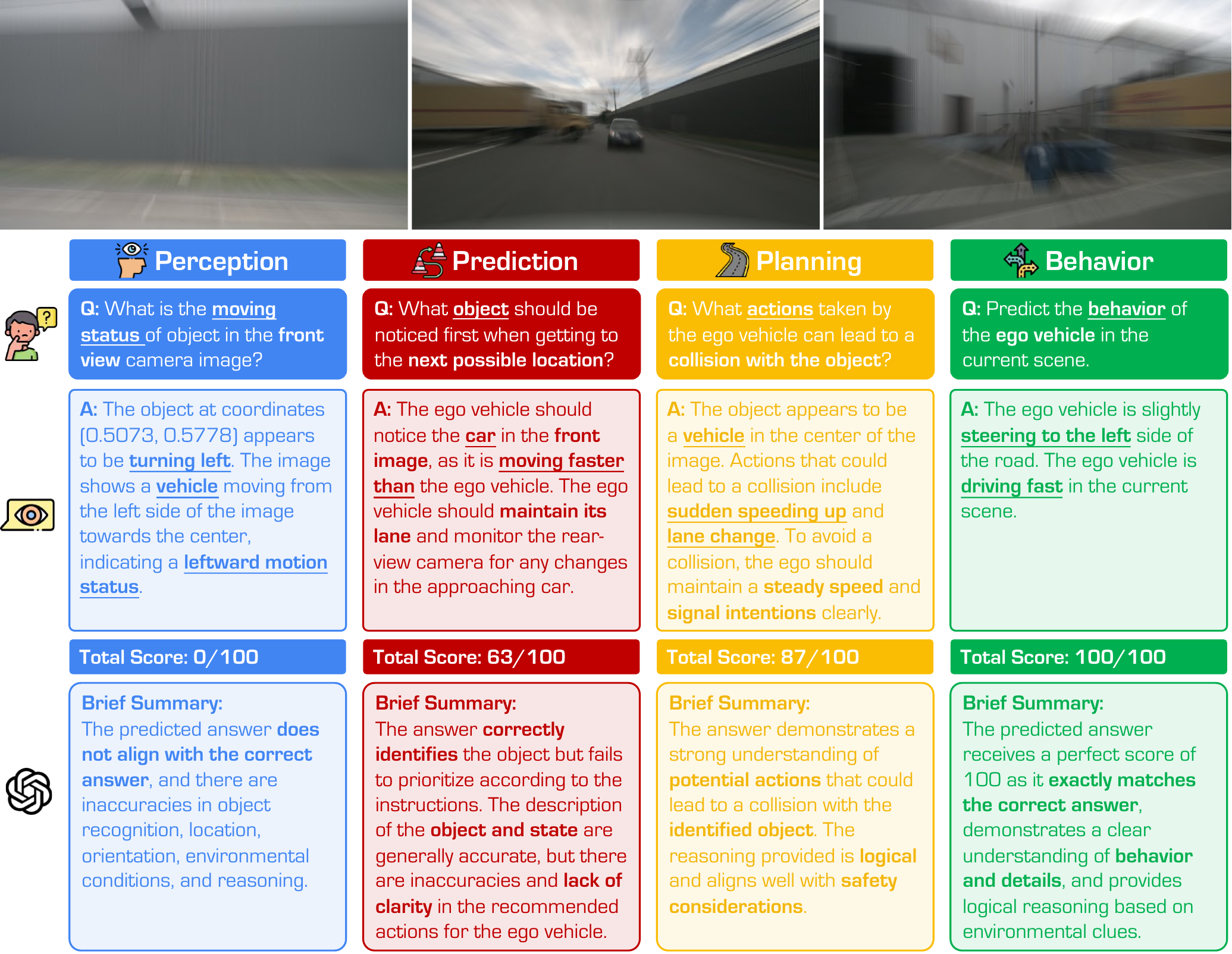}
    \vspace{-0.6cm}
    \caption{Examples of GPT-4o~\cite{achiam2023gpt} responses to four tasks and the corresponding evaluation results under the \textcolor{robo_red}{\textbf{motion blur condition}}. We observe that GPT-4o~\cite{achiam2023gpt} are influenced by this type of corruption and tend to predict \textit{``driving fast''} based on it. The example shows the potential of visual corruption to influence high-level driving decisions.}
    \label{fig:examples_benchmark_2}
\end{figure*}

\begin{figure*}[t]
    \centering
    \includegraphics[width=\linewidth]{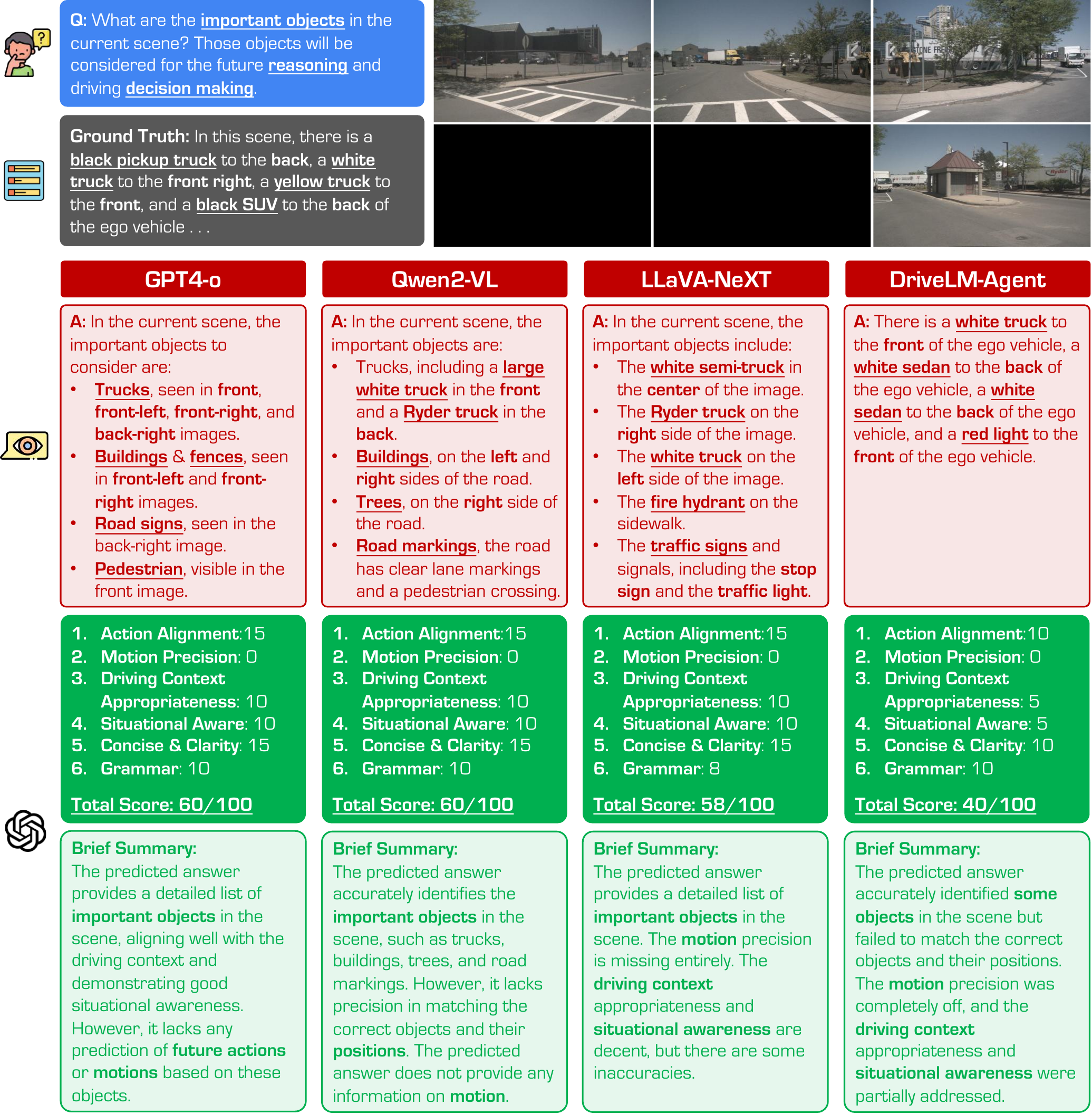}
    \vspace{-0.6cm}
    \caption{Examples of different VLM responses under the \textcolor{robo_red}{\textbf{frame lost condition}}. We observe that GPT-4o~\cite{achiam2023gpt} responses with visible objects while LLaVA-NeXT~\cite{liu2024llavanext} and DriveLM~\cite{sima2023drivelm} tend to hallucinate objects that cannot be seen from the provided images.}
    \label{fig:examples_benchmark_3}
\end{figure*}

\begin{figure*}[t]
    \centering
    \includegraphics[width=\linewidth]{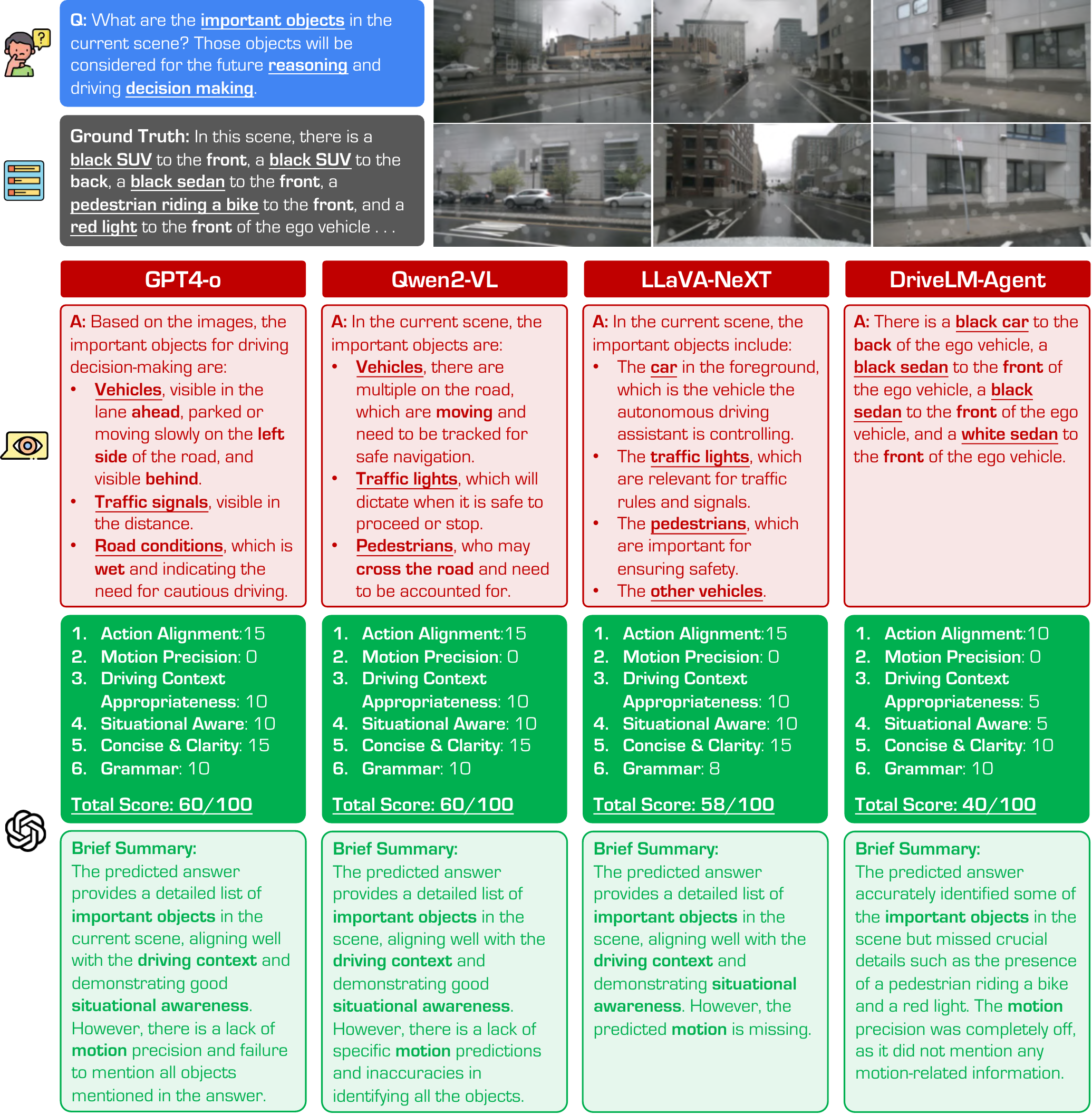}
    \vspace{-0.6cm}
    \caption{Examples of different VLM responses under the \textcolor{robo_red}{\textbf{water splash condition}}. We observe that, under severe visual corruptions, VLMs respond with ambiguous and general answers based on their learned knowledge, without referring to the visual information. Most responses include traffic signals and pedestrians, even though they are not visible in the provided images.}
    \label{fig:examples_benchmark_4}
\end{figure*}

\subsubsection{Corruption Awareness} 
We explore whether the fabricated ``reasonable" answer of VLMs under corruption might stem from a lack of \textit{awareness} regarding potential visual corruptions. To investigate this, we conduct two experiments: E-1) involves explicit corruption reference when prompting the model, \eg, \textit{``what are the important objects \textbf{in the snowy day}}'', and E-2) we directly ask the model to identify the current type of image corruption, \eg, \textit{``what is the current corruption''}.

In E-1, we analyze changes in MCQ accuracy when VLMs are explicitly prompted with references to visual corruption. As shown in \cref{tab:corruption-context}, the results demonstrate a notable trend of decreasing accuracy across various models and corruption types. Certain models exhibit substantial performance declines in the presence of corruption prompts; for example, LLaVA-NeXT$_\text{7B}$ \cite{liu2024llavanext} experiences an accuracy reduction of approximately $19.62\%$. A closer examination of model responses reveals increased uncertainty when the corruption context is included in the prompt. For instance, the model may respond with statements such as \textit{``based on the image provided, it is not possible to accurately determine the moving status of the object (480, 520) given the camera crash corruption''}. These findings suggest that some models exhibit a degree of corruption awareness when explicitly prompted, recognizing potential unreliability in their responses under conditions of severe visual degradation. The differences in responses between prompts with and without explicit corruption references lead to two significant insights: 
\begin{itemize}
    \item VLMs demonstrate a limited capacity to identify visual corruption when not explicitly prompted.
    \item In the absence of corruption-specific prompts, VLMs frequently rely on general knowledge rather than on the degraded visual information in the current driving scene, often generating responses that feign an understanding of the visual context.
\end{itemize}

These insights raise critical concerns, as reliable autonomous driving depends on accurate situational awareness and precise interpretation of environmental cues. The observed behavior suggests that current VLMs may lack the capacity for actively corruption-aware reasoning, defaulting instead to generalized responses and common sense knowledge (\eg, \textit{``going ahead"}, \textit{``maintain a safe distance"}, \textit{``be cautious"}, \etc). This tendency can be considered a form of hallucination \cite{ji2023survey}, which undermines the visual grounding reliability of VLMs within the safety-critical context of autonomous driving.

Conversely, models such as LLaVA-1.5 \cite{liu2024llava1.5} with 7B and 13B parameters exhibit minimal performance changes even when corruption-specific prompts are provided. This observation, when combined with the previous findings, suggests two possible explanations: 1) these models may lack the capability to detect image corruption, or 2) while aware of the corruption, their responses remain dominated by general knowledge rather than visual cues, even in clean situations, thus lead to unchanged performance.

To investigate the first hypothesis, we conduct E-2, in which we explicitly prompt the VLMs to identify the type of visual corruption or determine the number of corrupted cameras in scenarios involving camera crash and frame lost corruptions \cite{xie2024benchmarking}. The results, summarized under the MCQ columns in \cref{tab:benchmark_robustness}, indicate that LLaVA-1.5 \cite{liu2024llava1.5} achieves high accuracy in identifying corruption types, particularly in weather and motion corruption scenarios, suggesting that it indeed possesses corruption awareness.

To evaluate the second hypothesis, we analyze the confusion matrix of responses from LLaVA-1.5 \cite{liu2024llava1.5}. Remarkably, the model consistently outputs \textit{Going ahead} across all corruption scenarios, regardless of the actual visual context (shown in \cref{fig:llava1.5-13b-spatial} in Appendix). This uniformity in answering indicates the model response is based on general knowledge rather than counting on visual inputs. Combining with the findings in \cref{sec:exp-corrupt}, we conclude below:
\begin{itemize}
    \item Advanced VLMs tend to rely predominantly on text-based cues to generate responses under conditions of visual degradation, even though they are aware of it.
    \item Less advanced models demonstrate a stronger dependence on general knowledge acquired during training, resulting in responses dominated by learned priors rather than situational visual information.
\end{itemize}


\subsubsection{Fine-Tuned VLM Models}
Thus far, our analysis has primarily focused on general-purpose VLMs applied in driving. While this focus is essential, given that these models serve as foundational architectures for task-specific fine-tuning, it remains insufficient for evaluating driving-specific capabilities, as these general models are not tailored or optimized for autonomous driving tasks. In this section, we shift our attention to VLMs fine-tuned specifically on driving datasets, reflecting the growing body of research dedicated to this area \cite{sima2023drivelm, ma2023dolphins, tian2024drivevlm}. 

Specifically, we select DriveLM \cite{sima2023drivelm} and Dolphin \cite{ma2023dolphins} as representative models for our analysis, as both exhibit promising results and are explicitly fine-tuned to enhance visual-grounded driving decision-making abilities. DriveLM is fine-tuned on the nuScenes dataset \cite{qian2024nuscenes}, while Dolphin is fine-tuned on the BDD dataset \cite{yu2020bdd100k}. This distinction in fine-tuning datasets offers a unique opportunity to investigate the transferability of driving-specific VLMs across different datasets, as well as their answer reliability to visual corruption.

The main results are summarized in \cref{tab:benchmark} and \cref{tab:benchmark_robustness}. A key observation is that Dolphin \cite{ma2023dolphins}, which is primarily fine-tuned on the BDD \cite{yu2020bdd100k} dataset, demonstrates significant difficulty in answering questions from the nuScenes \cite{qian2024nuscenes} dataset. Given the general capabilities of VLMs to address questions across diverse domains, this result is both surprising and concerning, highlighting the limited generalizability of driving-specific VLMs when exposed to datasets or question formats that differ from their fine-tuning conditions. Regarding DriveLM \cite{sima2023drivelm}, we further investigate how the model benefits from in-distribution fine-tuning in the following section. This analysis aims to elucidate the potential advantages and limitations of fine-tuning on a specific language-annotated driving dataset.

\subsubsection{Metrics} 
\label{sec:exp-metric}

\noindent\textbf{Language Metrics.} The insights presented thus far rely primarily on accuracy and GPT-based scores. However, due to the prohibitive costs associated with large-scale evaluations using GPT APIs, traditional language metrics, such as BLEU \cite{papineni2002bleu} and ROUGE-L \cite{lin2004rouge} remain widely employed in existing benchmarks \cite{sima2023drivelm, kim2018textual}. To better understand their applicability, we evaluate the validity of these metrics in the context of language-based driving decision tasks. In \cref{fig:metric}, we present the prediction task performance under clean image inputs, evaluated across different metrics. As anticipated, the language metrics demonstrate high internal consistency: models with high BLEU scores also tend to achieve high ROUGE-L scores, as both metrics emphasize pattern-matching between predicted responses and ground-truth answers. Furthermore, we visualize how the same responses are scored under ROUGE-L and GPT scores at scale in \cref{fig:metric-scatter}. The results further reveal how the language score fails to reflect underlying key information: when ROUGE-L remains around $0.2$, GPT scores vary across a large range. DriveLM \cite{sima2023drivelm}, while surpassing other VLMs with large margins under ROUGE-L evaluation, still lags behind Qwen2-VL$_\text{72B}$ \cite{wang2024qwen2} and GPT-4o \cite{achiam2023gpt} in GPT evaluation. The observation indicates that the main improvement of in-distribution fine-tuning on the current small-scale driving dataset largely comes from the answering template.

\noindent\textbf{Accuracy.} In terms of accuracy, we also study how accuracy and GPT scores are related. The results are presented in Fig~\ref{fig:metric-mcq}. The GPT evaluation highly aligns with accuracy since we prompt the GPT to assign certain scores if the answer is correct. The divination is because GPT assigns another portion of scores to the coherence of explanation dimensions, capturing nuanced differences between answers.

\noindent\textbf{GPT Evaluation.} A critical question remains: is GPT evaluation currently the optimal approach? The answer is nuanced. GPT-based scoring can effectively capture human preferences and emphasize critical elements in driving scenarios, yet this capability is highly contingent on the provision of comprehensive driving contextual information. We empirically compare how the same response is scored given different information, shown in \cref{fig:metric-rubric}. When GPT evaluation is prompted solely with GT and model response, the resulting scores are highly homogeneous while the inclusion of specific rubrics, questions, and specific-driving context yields greater score diversity.

\begin{figure*}[t]
    \centering
    \includegraphics[width=\linewidth]{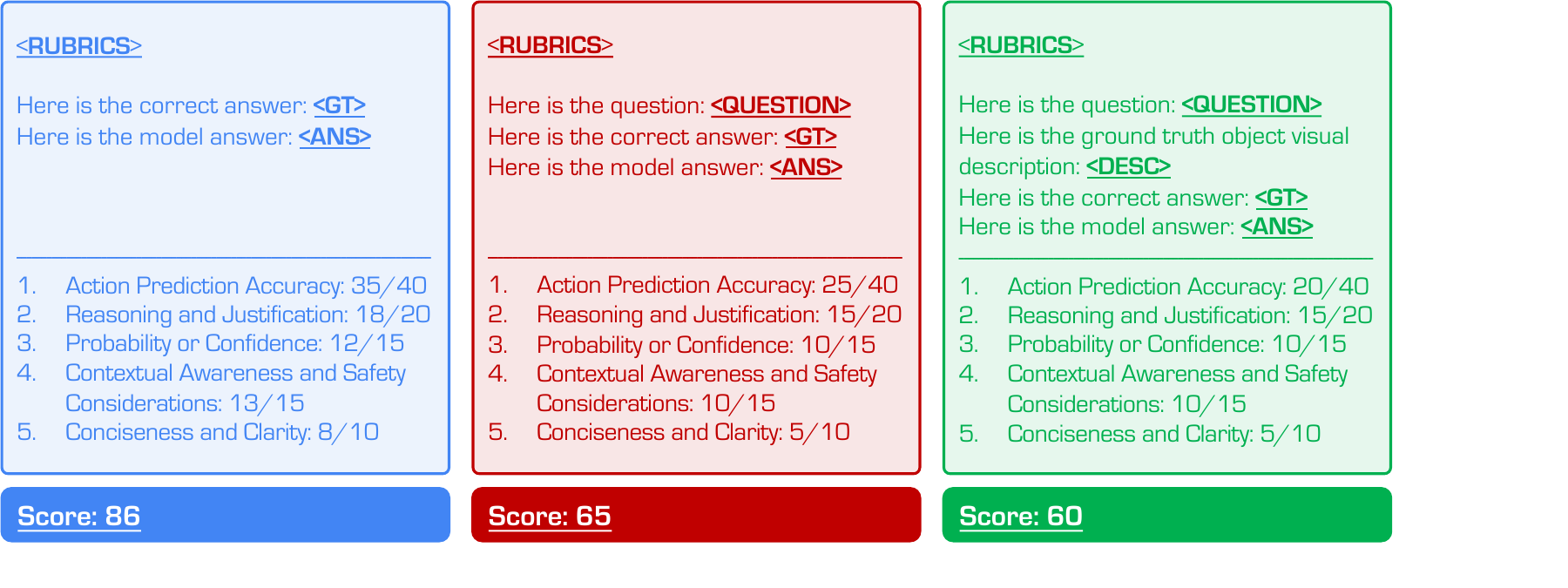}
    \vspace{-0.6cm}
    \caption{Examples of GPT evaluators with different information. The evaluator can revise the score and give a more accurate evaluation based on more contextual information in the driving scenarios. For more details please refer to Fig.~\ref{fig:exmaple-prompt-detailed}.}
    \label{fig:enter-label}
\end{figure*}
\section{Conclusion}
\label{sec:conclusion}

This work identifies and addresses key challenges in the deployment of Vision-Language Models (VLMs) for autonomous driving, with an emphasis on their visual-grounding reliability in complex real-world scenarios. Our findings reveal that VLMs frequently generate plausible yet unsupported responses when subjected to severe visual degradation, casting doubt on their reliability in critical decision-making tasks. Furthermore, imbalanced datasets and suboptimal evaluation protocols amplify these concerns, contributing to an overestimation of VLM reliabilities. To mitigate these challenges, we advocate for future efforts to prioritize the development of well-balanced, context-aware datasets and advanced evaluation metrics that rigorously assess the quality, contextual reasoning, and safety of driving decisions.

\appendix
\section*{Appendix}
\startcontents[appendices]
\printcontents[appendices]{l}{1}{\setcounter{tocdepth}{3}}

\section{Benchmark Setup}
\label{sec:benchmark_setup}
In this section, we elaborate in detail on the procedures and protocols used to establish the \textsf{\textcolor{robo_blue}{Drive}\textcolor{robo_red}{Bench}} in this work.

\subsection{Benchmark Construction}
\label{sec:bench_construction}
We detailed the benchmark construction process in this section. Our \textsf{\textcolor{robo_blue}{Drive}\textcolor{robo_red}{Bench}} is primarily inspired by DriveLM \cite{sima2023drivelm} given its impact and representativeness. Given its public availability, we subsample $200$ keyframes from the DriveLM \cite{sima2023drivelm} training dataset. These keyframes are selected to balance the ground truth distribution, which can more accurately reflect the model's performance and prevent bias in most common cases (\eg, \textit{``Going Ahead''}). Each keyframe has multiple questions related to different tasks, spanning perception, prediction, planning, and behavior. For each task, we follow the question type design in DriveLM~\cite{sima2023drivelm}, including multiple-choice questions (MCQs) and visual question answering (VQAs), as shown in \cref{tab:sup-data}. When evaluating corruption awareness, we add information about corruption context to the question and modify the answer accordingly if necessary. We generate the corruption-related question-answering pairs by prompting GPT4 based on original QA pairs, which we refer to the robustness dataset as shown in \cref{tab:sup-data-robust}.

\begin{table}[t]
\centering
\caption{\textbf{Detailed distribution} of the curated benchmark dataset.}
\vspace{-0.2cm}
\label{tab:sup-data}
\resizebox{\linewidth}{!}{
\begin{tabular}{c|l|c|c|c}
    \toprule
    \textbf{\#} & \textbf{Driving Task} & \textbf{Question Type} & \textbf{\# Samples} & \textbf{Total}
    \\
    \midrule\midrule
    ~\includegraphics[width=0.042\linewidth]{figures/icons/perception.png}~ & $\circ$ Perception & MCQ \& VQA & $400$ & \multirow{4}{*}{$\mathbf{1,261}$}  
    \\
    ~\includegraphics[width=0.042\linewidth]{figures/icons/prediction.png}~ & $\circ$ Prediction & VQA & $61$             
    \\
    ~\includegraphics[width=0.042\linewidth]{figures/icons/planning.png}~ & $\circ$ Planning & VQA & $600$             
    \\
    ~\includegraphics[width=0.042\linewidth]{figures/icons/behavior.png}~ & $\circ$ Behavior & MCQ & $200$             
    \\
    \bottomrule
\end{tabular}}
\end{table}

\begin{table}[t]
\centering
\caption{\textbf{Detailed distribution} of the proposed robustness benchmark dataset. The total number is summed across all the corruption types. ``Corrupt. Rec." represents corruption recognition, which asks the model to identify the current corruption types. ``Corrupt. Desc." represents questions related to the description of the current corrupted environment.}
\vspace{-0.2cm}
\label{tab:sup-data-robust}
\resizebox{\linewidth}{!}{
\begin{tabular}{c|l|c|c|c}
    \toprule
    \textbf{\#} & \textbf{Driving Task} & \textbf{Question Type} & \textbf{\# Samples} & \textbf{Total}
    \\
    \midrule\midrule
    ~~\includegraphics[width=0.044\linewidth]{figures/icons/human.png}~ & $\circ$ Corrupt. Rec. & MCQ & $4,000$ &
    \\
    ~\includegraphics[width=0.042\linewidth]{figures/icons/perception.png}~ & $\circ$ Perception & MCQ \& VQA & $5,475$ & \multirow{3}{*}{$\mathbf{19,237}$}  
    \\
    ~\includegraphics[width=0.042\linewidth]{figures/icons/prediction.png}~ & $\circ$ Prediction & VQA & $799$             
    \\
    ~\includegraphics[width=0.042\linewidth]{figures/icons/planning.png}~ & $\circ$ Planning & VQA & $5,999$             
    \\
    ~\includegraphics[width=0.042\linewidth]{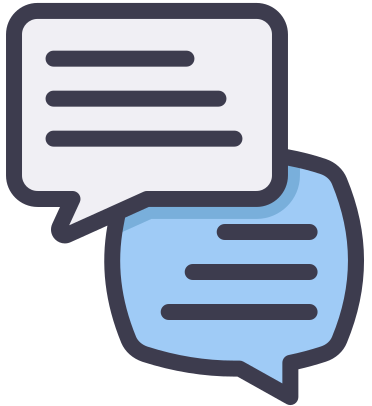}~ & $\circ$ Corrupt. Desc. & CAP & $3,000$             
    \\
    \bottomrule
\end{tabular}}
\end{table}

\subsection{Corruption Definitions}
In this section, we detailed our settings for generating image corruption. \textsf{\textcolor{robo_blue}{Drive}\textcolor{robo_red}{Bench}} encompasses five distinctive corruption categories, each with multiple different types of corruptions reflecting the real-world scenarios.
\begin{itemize}
    \item \includegraphics[width=0.05\linewidth]{figures/icons/weather.png}~\textbf{Weather \& Lighting Conditions ($\mathbf{5}$ Types)}:\\
    The simulations of diverse environmental weather and lighting conditions are used in the driving scenarios. In this benchmark, we include the $^1$\texttt{Brightness}, $^2$\texttt{Dark}, $^3$\texttt{Snow}, $^4$\texttt{Fog}, and $^5$\texttt{Rain} corruptions.
    
    \item \includegraphics[width=0.05\linewidth]{figures/icons/external.png}~\textbf{External Disturbances ($\mathbf{2}$ Types)}:\\
    The simulations of situations where camera lenses are occluded by external objects or stains. In this benchmark, we include the $^6$\texttt{Water Splash} and $^7$\texttt{Lens Obstacle} corruptions.
    
    \item \includegraphics[width=0.05\linewidth]{figures/icons/sensor.png}~\textbf{Sensor Failures ($\mathbf{3}$ Types)}:\\
    The simulations of sensor failures. In this benchmark, we include the $^8$\texttt{Camera Crash}, $^9$\texttt{Frame Lost}, and $^{10}$\texttt{Saturate} corruptions.
    
    \item \includegraphics[width=0.05\linewidth]{figures/icons/motion.png}~\textbf{Motion Blurs ($\mathbf{2}$ Types)}:\\
    The simulations of the blurs caused by the ego vehicle's high-speed motion. In this benchmark, we include the $^{11}$\texttt{Motion Blur} and $^{12}$\texttt{Zoom Blur} corruptions.
    
    \item \includegraphics[width=0.05\linewidth]{figures/icons/transmission.png}~\textbf{Data Transmission Errors ($\mathbf{3}$ Types)}: 
    The simulations of the errors happening during the video transmission process. In this benchmark, we include the $^{13}$\texttt{Bit Error}, $^{14}$\texttt{Color Quantization}, and $^{15}$\texttt{H.265 Compression} corruptions.
\end{itemize}

All the $15$ corruption types are generated by applying high-fidelity image processing algorithms developed in previous works \cite{kong2023robo3d, kong2024robodepth, xie2024benchmarking, yi2021benchmarking}. Here, we detail how each corruptions are synthesized as follows:
\begin{itemize}
    \item $^1$\texttt{Brightness}: Adjusts the brightness values of the camera images by scaling pixel intensity upwards.
    \item $^2$\texttt{Dark}: Simulates low-light conditions by scaling down the image's brightness using a gamma-adjusted mapping. Additionally, it introduces Poisson noise to mimic photon shot noise and Gaussian noise to simulate sensor noise.
    \item $^3$\texttt{Snow}: Generates a synthetic snow layer using random noise, applies motion blur to simulate falling snow, and blends it with the original image.
    \item $^4$\texttt{Fog}: Simulates fog by blending a fractal noise-based fog layer over the image.
    \item $^5$\texttt{Rain}: Adds streak-like artifacts to the image, created through line patterns combined with motion blur, to simulate rain.
    \item $^6$\texttt{Water Splash}: Simulates water splashes by overlaying transparent circular droplet patterns on the image, followed by a Gaussian blur to mimic water distortion effects.
    \item $^7$\texttt{Lens Obstacle}: Creates lens obstruction effects by blending blurred and unblurred regions of the image using a randomly placed and shaped elliptical mask to emulate obstructions on the lens surface.
    \item $^8$\texttt{Camera Crash}: Simulates a camera crash by replacing the affected image frames with black frames, representing a complete loss of data for specific viewpoints or cameras.
    \item $^9$\texttt{Frame Lost}: Emulates frame loss by randomly setting some frames to black, indicating partial data corruption or a temporary transmission failure.
    \item $^{10}$\texttt{Saturate}: Modifies the image's color saturation by manipulating the saturation channel in the HSV color space, either enhancing or reducing the vibrancy of colors.
    \item $^{11}$\texttt{Motion Blur}: Applies a linear motion blur to the image, simulating movement during exposure, with the blur radius and direction determined by severity.
    \item $^{12}$\texttt{Zoom Blur}: Applies a radial zoom effect to the image, creating a focused blur that simulates rapid movement toward or away from the lens, controlled by severity.
    \item $^{13}$\texttt{Bit Error}: Introduces random bit-level noise in the image data, mimicking digital corruption, with severity influencing the extent of errors.
    \item $^{14}$\texttt{Color Quantization}: Reduces the image's color palette to a limited set of levels, simulating low-quality color quantization, where severity controls the number of colors.
    \item $^{15}$\texttt{H.265 Compression}: Applies heavy H.265 video compression artifacts to the image, with severity increasing the compression level and artifact visibility.
\end{itemize}

\subsection{Overall Statistics}
The results in the main paper are based on the curated datasets, which contain $1,461$ questions, the detailed distribution of the questions is shown in \cref{tab:sup-data}. Specifically, each keyframe has two perception questions: one for MCQ and the other for VQA, four VQA questions for the planning task, and one behavior MCQ for the behavior task. In terms of the prediction task, not all keyframes have corresponding prediction questions.

\subsection{License}
The proposed benchmark is released under the Creative Commons Attribution-NonCommercial-ShareAlike 4.0 International License\footnote{\url{https://creativecommons.org/licenses/by-nc-sa/4.0}.}.
\section{Benchmark Study}
\label{sec:bench_study}
In this section, we include detailed information on the dataset distribution of the representative driving-with-language dataset. These datasets advance the development of driving with language models.

\subsection{DriveLM-nuScenes}
We visualize the dataset distribution in perception and behavior tasks in the DriveLM-nuScenes dataset \cite{sima2023drivelm}, as shown in \cref{fig:drivelm-distribution}. The majority of choices for vehicle behaviors are going straight, which is also shown in \cite{li2024ego}. Along with the DriveLM-Agent results in the main paper, we find that highly imbalanced data can cause several problems. When fine-tuning VLMs on this dataset, the model tends to memorize the majority of choices, and thus answer with it during inference even if the visual cues are absent, which prevents effective evaluation of the model reliability. The dataset mainly adopts language metrics \cite{papineni2002bleu, lin2004rouge} and naive GPT score: prompt with only the answer and ground truth, for evaluation. Our results in the main paper also show the limitations of these metrics in evaluating language-based driving decisions.

\subsection{BDD-X}

To show the general existence of the limitations of existing benchmarks, we also study the BDD-X~\cite{kim2018textual} dataset. We observe similar limitations as those in DriveLM~\cite{sima2023drivelm} dataset, where the data is highly imbalanced. Most actions of the car are \textit{``Stop"} or \textit{``Going Ahead"}, where the random guessing of VLMs can achieve high accuracy since we've shown they potentially guessed the answer based on common knowledge and general case in \cref{sec:exp-corrupt}. The observation shows that the limitations observed in our work are not individual but general drawbacks of existing driving-with-language benchmarks.

\begin{figure*}
    \centering
    \includegraphics[width=\linewidth]{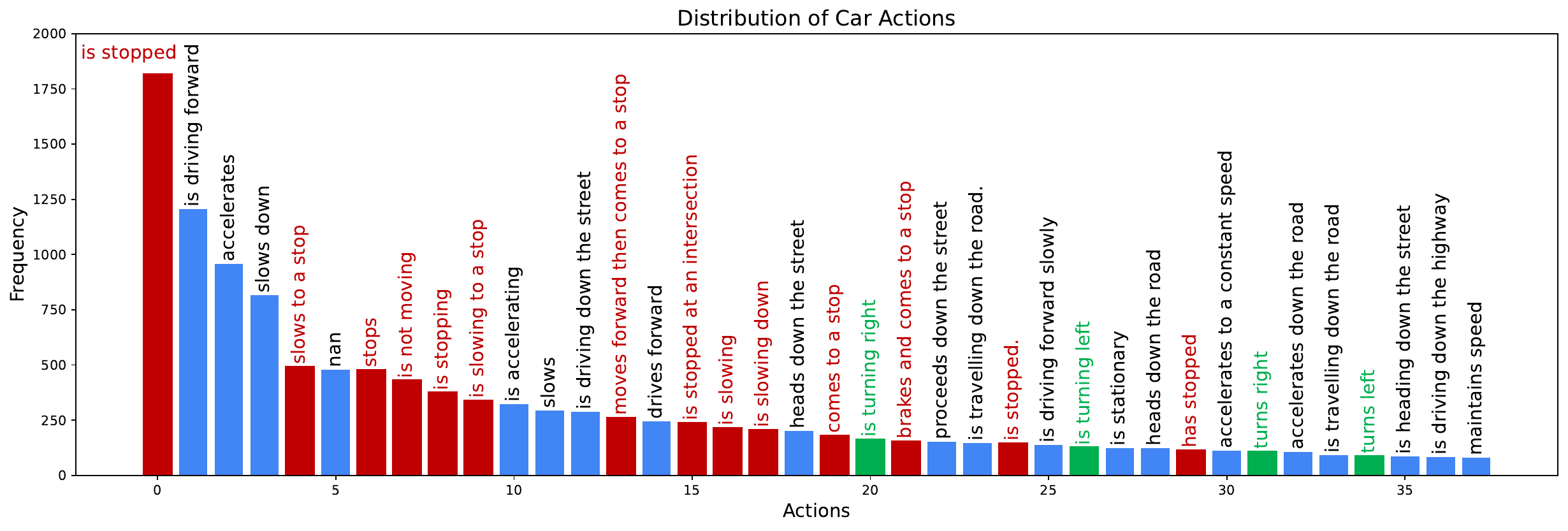}
    \vspace{-0.6cm}
    \caption{BDD-X dataset~\cite{kim2018textual}: detailed distribution of car actions. Only the actions with a frequency larger than $80$ are visualized. The \textcolor{robo_red}{\texttt{Stop}} actions and \textcolor{robo_green}{\texttt{Turn}} actions are highlighted. We observe similar data distribution in balance as those in DriveLM~\cite{sima2023drivelm}, where turning actions only account for a small portion of all actions.}
    \label{fig:bdd-x-dist}
\end{figure*}

\section{Additional Implementation Details}
In this section, we provide more details in terms of implementations and evaluation to facilitate the reproduction of this work.

\subsection{VLM Configurations}
\begin{itemize}
    \item \textbf{GPT-4o}~\cite{achiam2023gpt}, developed by OpenAI's, offers GPT-4-level intelligence with enhanced speed and multimodal capabilities, including voice, text, and image processing. It is designed to provide faster and more efficient responses across various applications.
    \item \textbf{Phi-3}~\cite{abdin2024phi} is a language model developed by Microsoft, focusing on efficiency and performance in natural language understanding and generation tasks. It is designed to handle a wide range of applications, from conversational agents to content creation.
    \item \textbf{Phi-3.5}~\cite{abdin2024phi} is an advanced version of Microsoft's Phi series, offering improved reasoning and mathematical capabilities comparable to larger models like GPT-4o. It maintains efficiency while managing complex AI tasks across different languages.
    \item \textbf{LLaVA-1.5}~\cite{liu2024llava1.5} is an open-source large multimodal model that integrates vision and language understanding. With 13 billion parameters, it achieves state-of-the-art performance across multiple benchmarks, rivaling models like GPT-4.
    \item \textbf{LLaVA-NeXT}~\cite{liu2024llavanext} is an evolution of the LLaVA series, enhancing multimodal capabilities by supporting multi-image, video, and 3D tasks within a unified large language model. It achieves state-of-the-art performance on a wide range of benchmarks, demonstrating strong video understanding through task transfer from images.
    \item \textbf{InterVL}~\cite{chen2023internvl} is an open-source multimodal dialogue model developed by OpenGVLab. It closely approximates the performance of proprietary models like GPT-4o, excelling in tasks that integrate visual and linguistic information, such as visual question answering and image captioning.
    \item \textbf{Oryx}~\cite{dong2024insight} is a unified multimodal architecture created by researchers from Tsinghua University and Tencent. It is designed for spatial-temporal understanding of images, videos, and multi-view 3D scenes, offering on-demand processing of visual inputs with arbitrary spatial sizes and temporal lengths.
    \item \textbf{Qwen2-VL}~\cite{qwen, wang2024qwen2} is a large language model developed by Alibaba Cloud, available in both chat and pretrained versions. It delivers high-quality language generation and understanding capabilities, optimized for tasks requiring nuanced comprehension and generation of human language.
    \item \textbf{DriveLM-Agent}~\cite{sima2023drivelm} is a model from OpenDriveLab tailored for autonomous driving applications, focusing on graph-based visual question answering. It addresses challenges in driving scenarios by integrating language understanding with visual perception, enhancing decision-making processes in autonomous systems.
    \item \textbf{Dolphins}~\cite{ma2023dolphins} is a multimodal language model developed by NVIDIA for driving applications. It adeptly processes inputs such as video data, text instructions, and historical control signals to generate informed outputs, facilitating a comprehensive understanding of complex driving scenarios.
\end{itemize}

\subsection{VLM Prompts}
\label{sec:vlm-prompt}

We use the same system prompt for all the candidate VLMs, as shown in \cref{fig:system-prompt}. We prompt the VLMs to explain for MCQs to facilitate the GPT evaluation based on their explanations.

\begin{figure}
\begin{codebox}
You are a smart autonomous driving assistant responsible for analyzing and responding to driving scenarios. You are provided with up to six camera images in the sequence [CAM FRONT, CAM FRONT LEFT, CAM FRONT RIGHT, CAM BACK, CAM BACK LEFT, CAM BACK RIGHT]. Each image has normalized coordinates from [0, 1], with (0,0) at the top left and (1,1) at the bottom right.
\\ \\
\textbf{Instructions:}
\\
1. \textbf{Answer Requirements}:
\begin{itemize}
    \item For multiple-choice questions, provide the selected answer choice along with an explanation.
    \item For “is” or “is not” questions, respond with a “Yes” or “No”, along with an explanation.
    \item For open-ended perception and prediction questions, related objects to which the camera.
\end{itemize}

2. \textbf{Key Information for Driving Context}:
\begin{itemize}
    \item When answering, focus on object attributes (\eg, categories, statuses, visual descriptions) and motions (\eg, speed, action, acceleration) relevant to driving decision-making
\end{itemize}
Use the images and coordinate information to respond accurately to questions related to perception, prediction, planning, or behavior, based on the question requirements.
\end{codebox}
\vspace{-0.4cm}
\caption{Inference system prompt.}
\label{fig:system-prompt}
\end{figure}

\subsection{GPT Evaluations}
\label{sec:gpt_evaluations}
We include the detailed prompt we used for GPT evaluation here, we use the following prompt to evaluate MCQ questions in the perception task as shown in \cref{fig:gpt-eval-prompt-mcq}. The \texttt{DESC} is used to prompt context information for accurate evaluation. Limited to the current drive-with-language dataset, we extract the natural language description of critical objects in the current environment to provide context information.

The prompt for the open-ended perception task is shown in \cref{fig:gpt-eval-prompt-open}. Since the ground truth for open-ended questions already included the visual description and moving status of important objects, we only prompt with \texttt{PRED} and \texttt{GT} with detailed rubrics.

\subsection{Human Evaluations}
\label{sec:human-eval}
In this subsection, we elaborate in more detail on how we conduct the human evaluation experiments in our benchmark.

\begin{figure*}[t]
    \centering
    \begin{subfigure}[t]{0.48\linewidth}
        \centering
        \includegraphics[width=\linewidth]{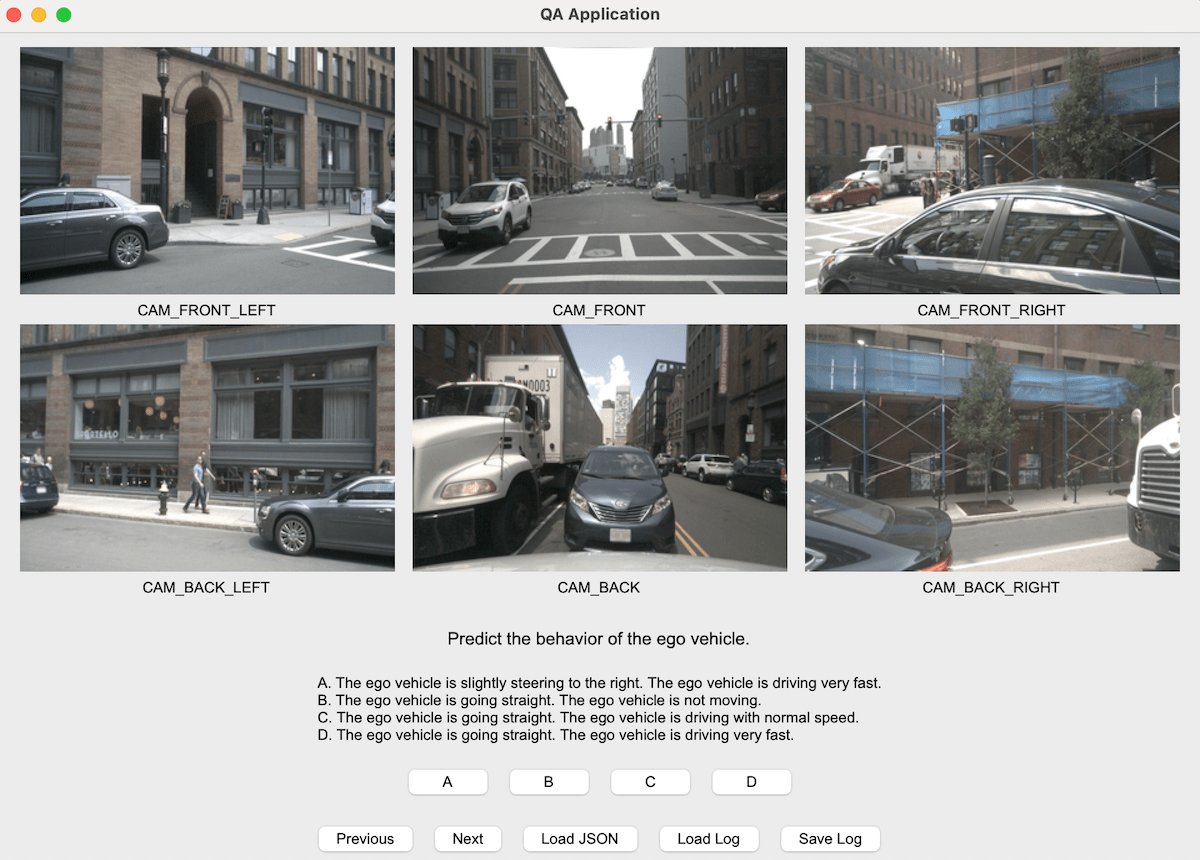}
        \vspace{-0.4cm}
        \caption{}
        \label{fig:human-eval-1}
    \end{subfigure}
    \vspace{0.2cm}
    \begin{subfigure}[t]{0.48\linewidth}
        \centering
        \includegraphics[width=\linewidth]{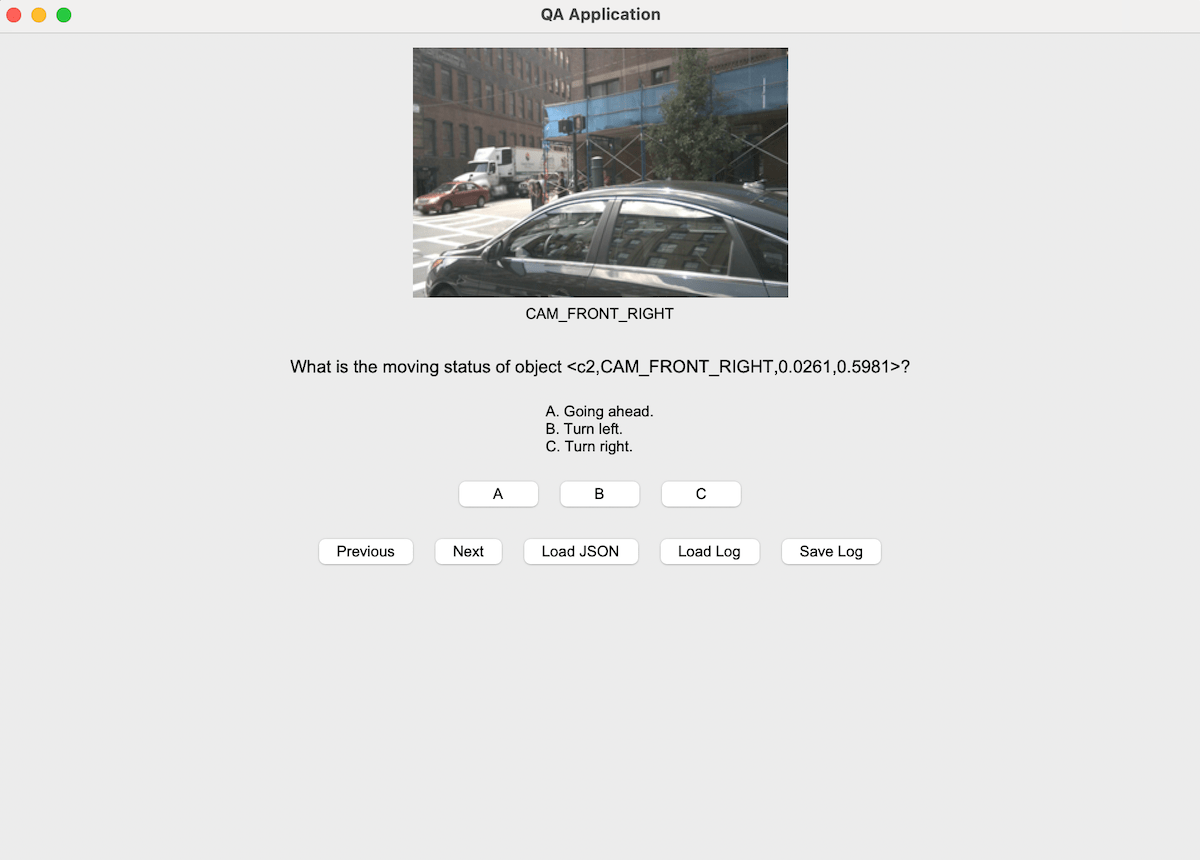}
        \vspace{-0.4cm}
        \caption{}
        \label{fig:human-eval-2}
    \end{subfigure}
    \vspace{0.2cm}
    \begin{subfigure}[t]{0.48\linewidth}
        \centering
        \includegraphics[width=\linewidth]{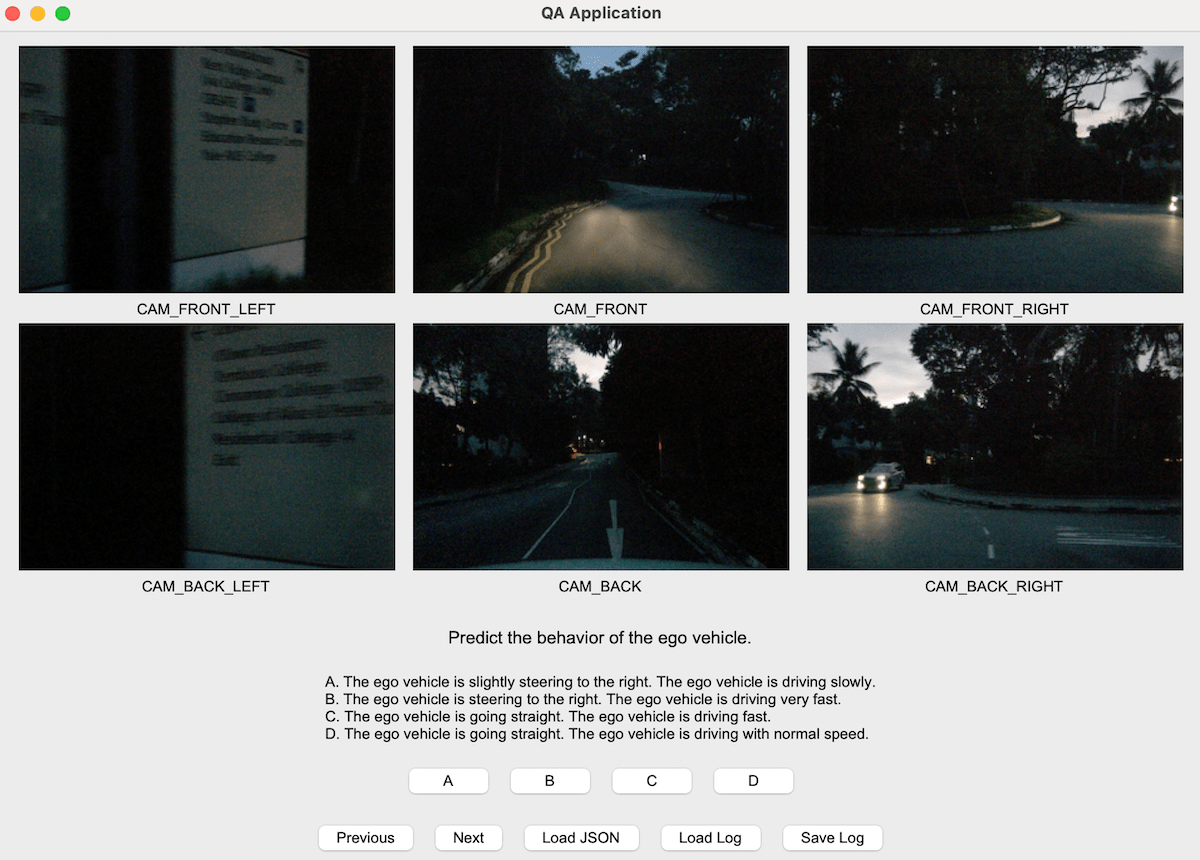}
        \vspace{-0.4cm}
        \caption{}
    \end{subfigure}
    \vspace{0.2cm}
    \begin{subfigure}[t]{0.48\linewidth}
        \centering
        \includegraphics[width=\linewidth]{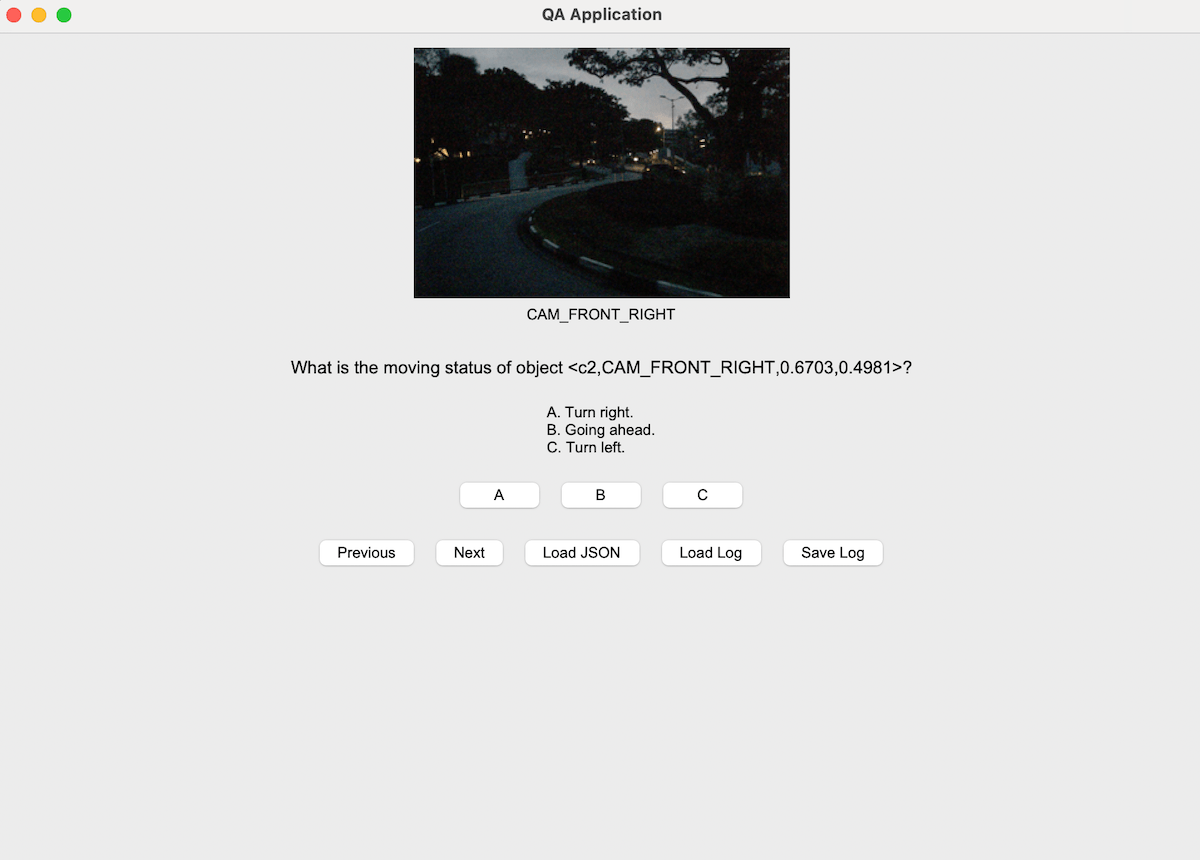}
        \vspace{-0.4cm}
        \caption{}
    \end{subfigure}
    \vspace{0.2cm}
    \begin{subfigure}[t]{0.48\linewidth}
        \centering
        \includegraphics[width=\linewidth]{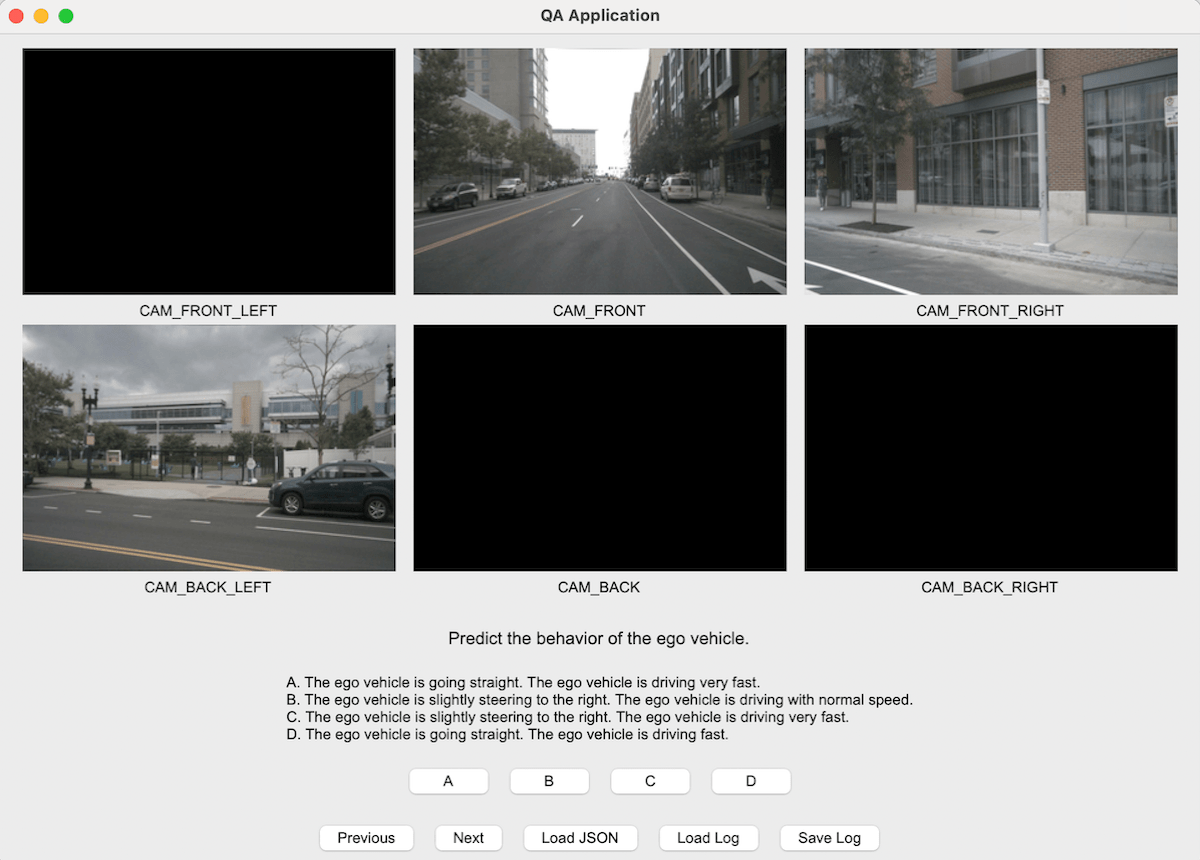}
        \vspace{-0.4cm}
        \caption{}
    \end{subfigure}
    \vspace{0.2cm}
    \begin{subfigure}[t]{0.48\linewidth}
        \centering
        \includegraphics[width=\linewidth]{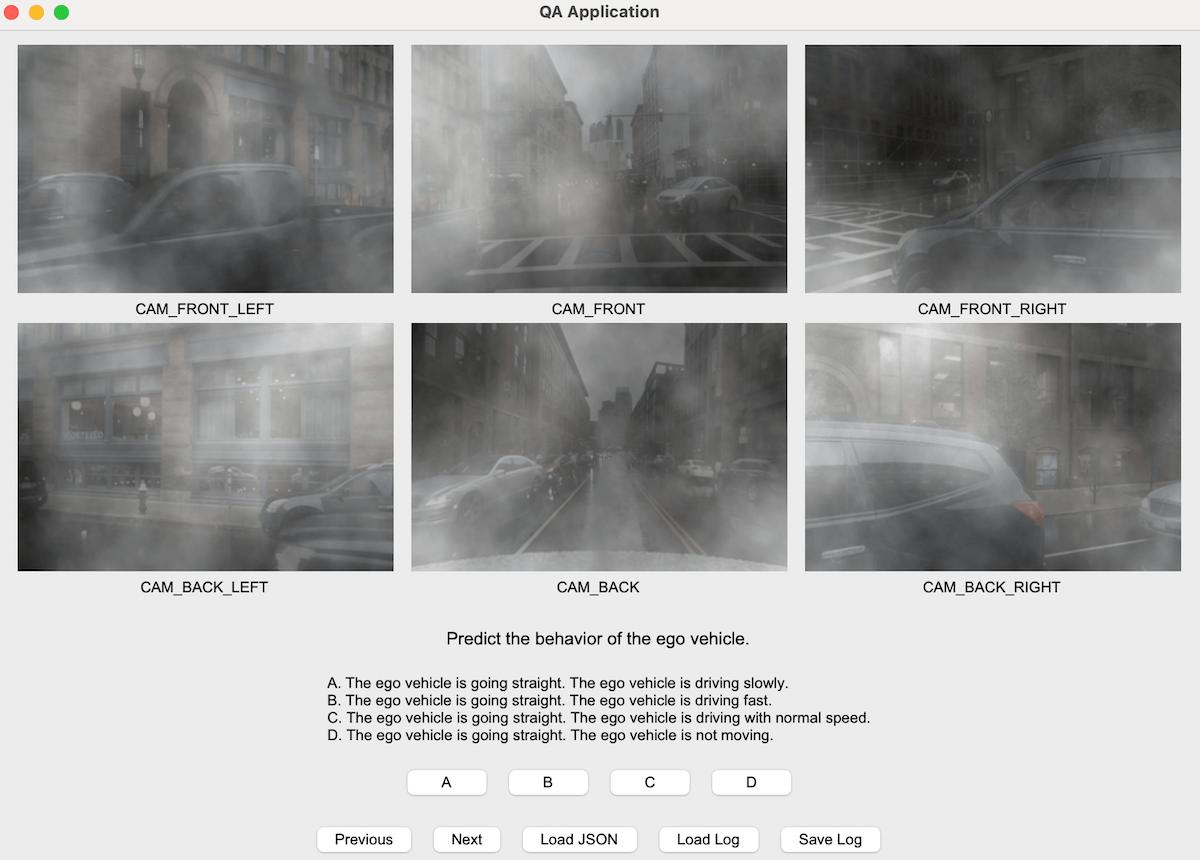}
        \vspace{-0.4cm}
        \caption{}
    \end{subfigure}
    \caption{Illustrative examples from our human evaluation interfaces.}
    \label{fig:human-eval}
\end{figure*}

\noindent\textbf{Procedures.}
Considering the large number of questions in the dataset, we subsample $15$ out of $200$ keyframes from our curated dataset. To ensure no overlaps between different corruptions, which might cause information leakage, we lower the probability if the same keyframes are sampled before. Then, we design a user interface for human evaluation, focusing on MCQs. The interface is shown in \cref{fig:human-eval}. As with evaluating VLMs, we only prompt single-view images if the questions are related to only one of the cameras.

\noindent\textbf{Ethic Declaration.}
According to the Federal Policy of Human Subjective Research\footnote{\url{https://www.federalregister.gov/d/2017-01058/p-1315}}, our research involves conducting anonymous visual recognition tasks, where participants respond to questions about visual stimuli without any interventions or demographic data collection. It qualifies for Exempt Research 2(i)\footnote{\url{https://www.federalregister.gov/d/2017-01058/p-1375}} because it solely involves survey-like procedures with no physical or psychological risks to participants. Specifically, it meets the requirements of Exempt Research 2(i) as the data is recorded in a manner ensuring that participants’ identities cannot be readily ascertained, directly or through linked identifiers. It does not fall under Exempt Research 2(ii) or 2(iii) because no identifiable information is recorded, and no IRB review is required to ensure these protections. We also submit an IRB review records to the corresponding institutions and receive the official confirmation of IRB review exemption.

\section{Detailed Experiment Results}
In this section, we include the detailed benchmark results evaluated by the GPT score metric. We also include the prediction spatial distribution in \cref{fig:camera_distribution} for each model.

\subsection{GPT Scores}

We include the detailed GPT scores in \cref{tab:perception-mcq}, \cref{tab:behavior-mcq}, \cref{tab:perception-open}, \cref{tab:prediction-open}, and \cref{tab:planning-open} for different tasks. The observation and conclusion in the main paper are primarily derived from GPT scores. Therefore, we focus more on the discussion of accuracy and language scores in the following sections.

\subsection{Accuracy Scores}

\begin{figure*}[t]
    \centering
    \begin{subfigure}[b]{0.48\linewidth}
        \centering
        \includegraphics[width=\linewidth]{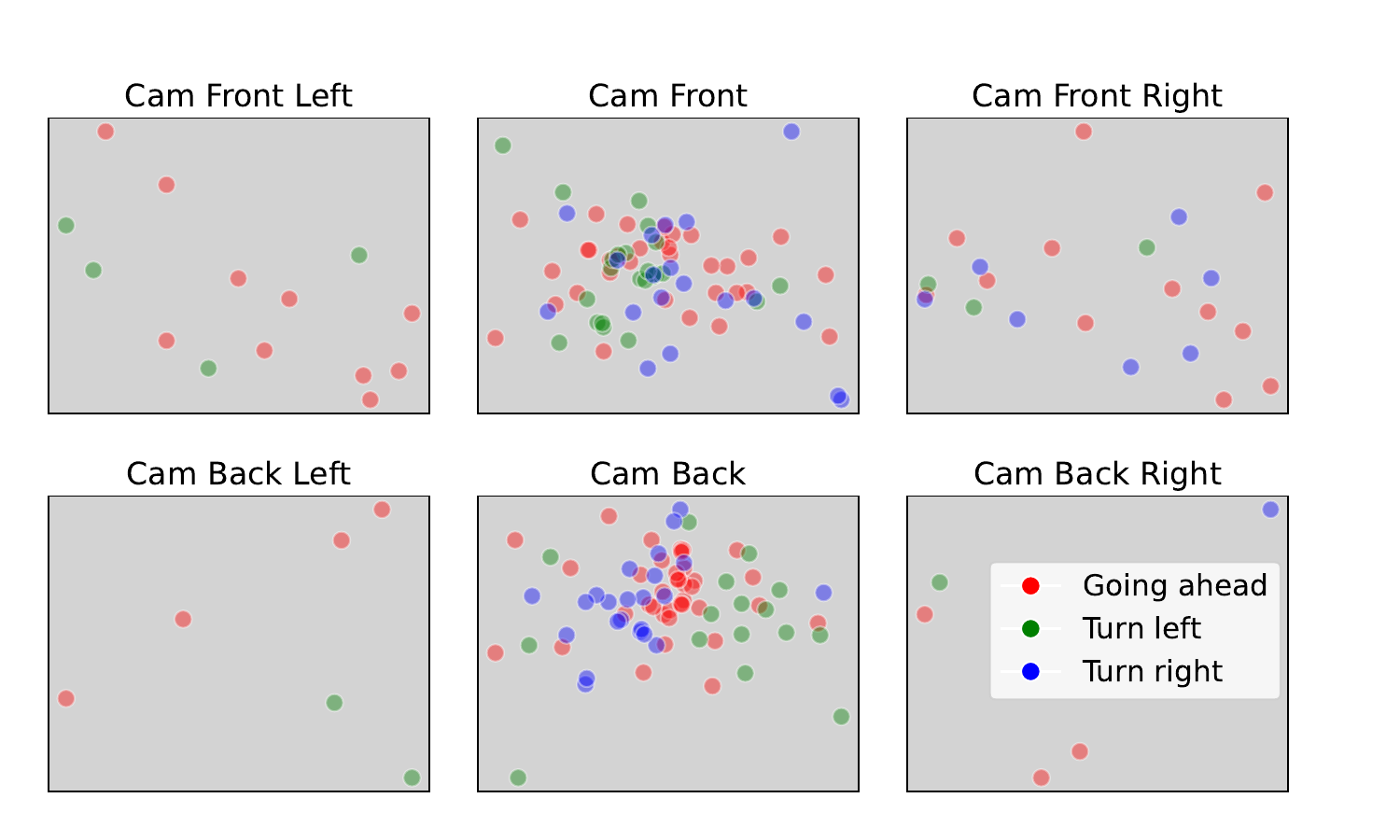}
        \vspace{-0.6cm}
        \caption{Ground Truth}
    \end{subfigure}
    \hfill
    \begin{subfigure}[b]{0.48\linewidth}
        \centering
        \includegraphics[width=\linewidth]{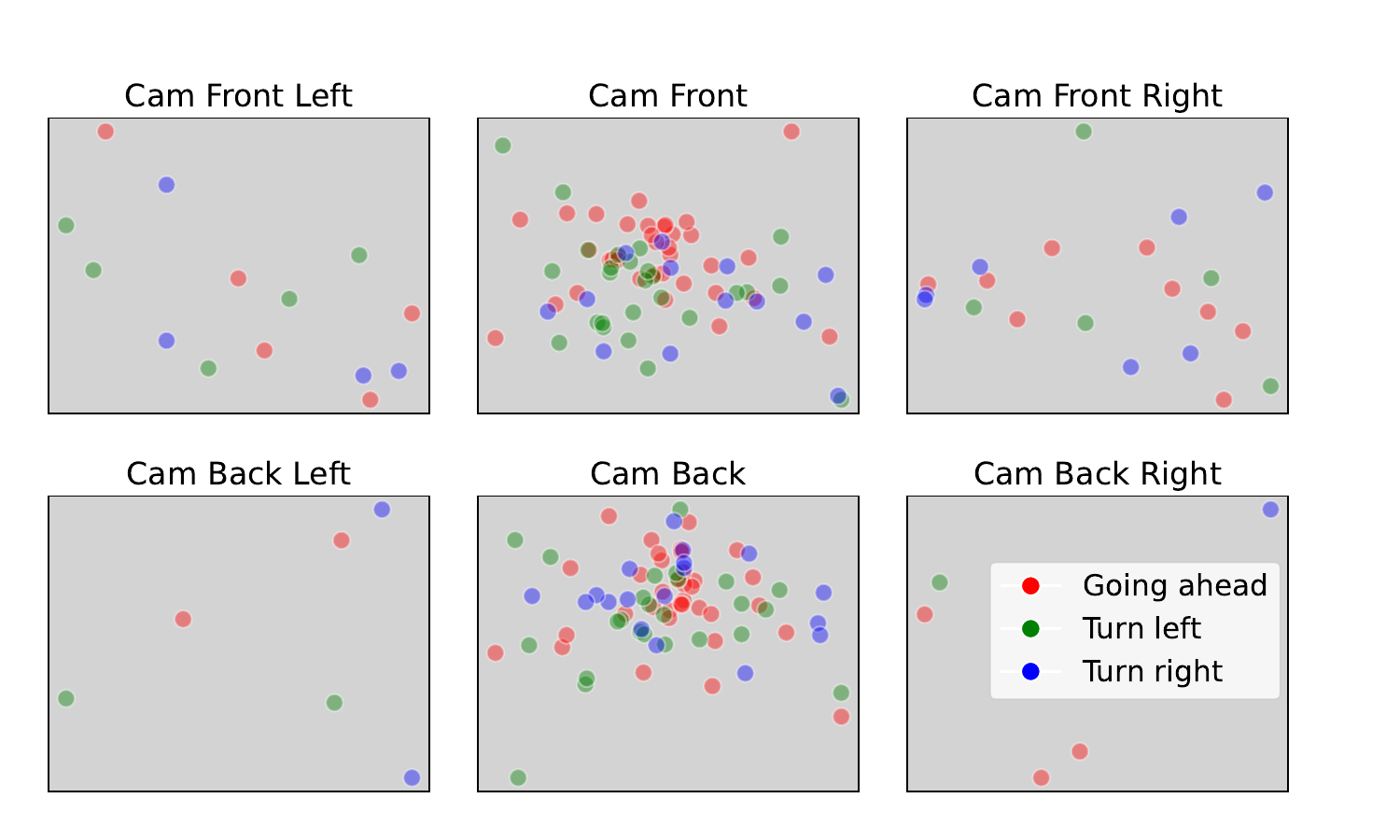}
        \vspace{-0.6cm}
        \caption{GPT-4o \cite{achiam2023gpt}}
    \end{subfigure}
    \hfill
    \begin{subfigure}[b]{0.48\linewidth}
        \centering
        \includegraphics[width=\linewidth]{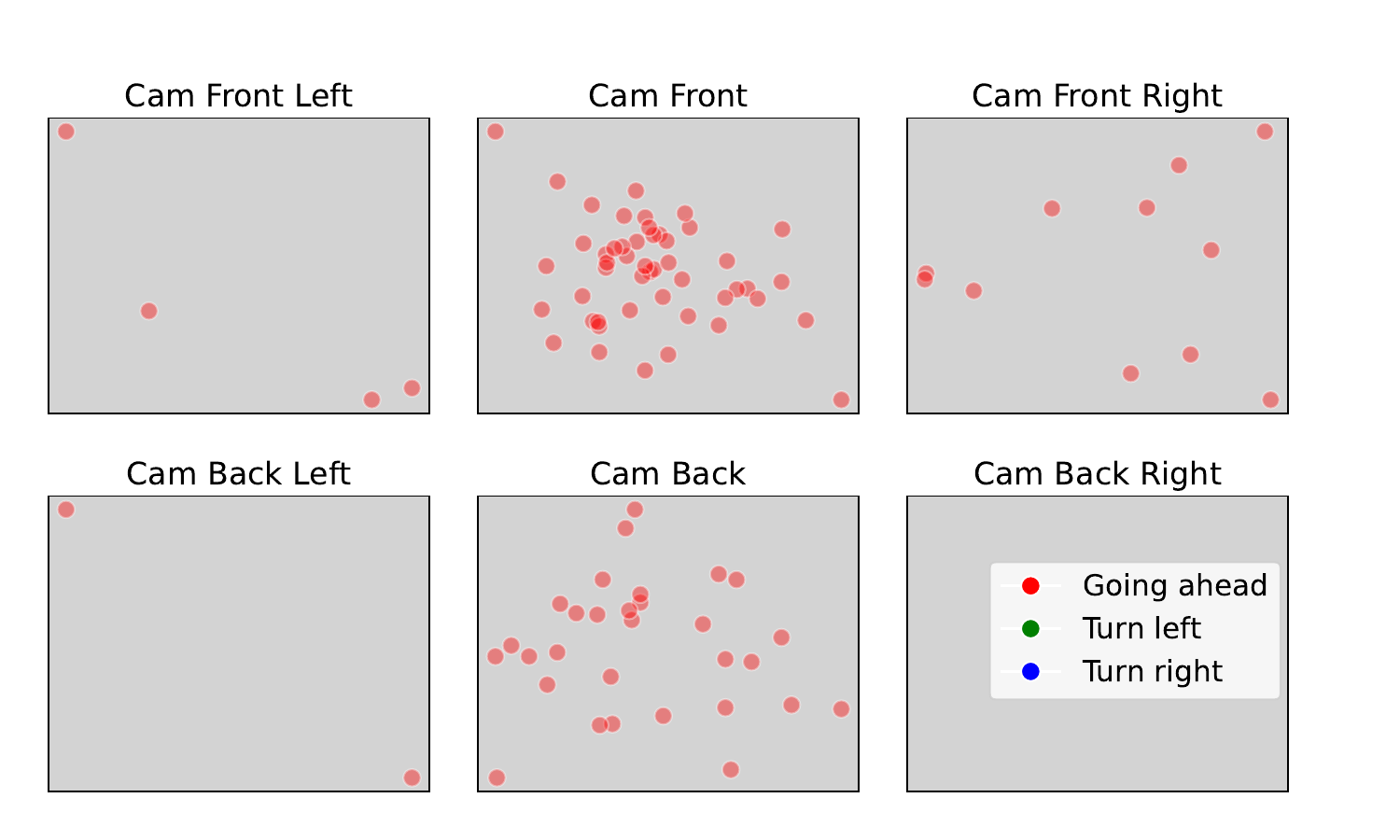}
        \vspace{-0.6cm}
        \caption{Phi-3 \cite{abdin2024phi}}
    \end{subfigure}
    \hfill
    \begin{subfigure}[b]{0.48\linewidth}
        \centering
        \includegraphics[width=\linewidth]{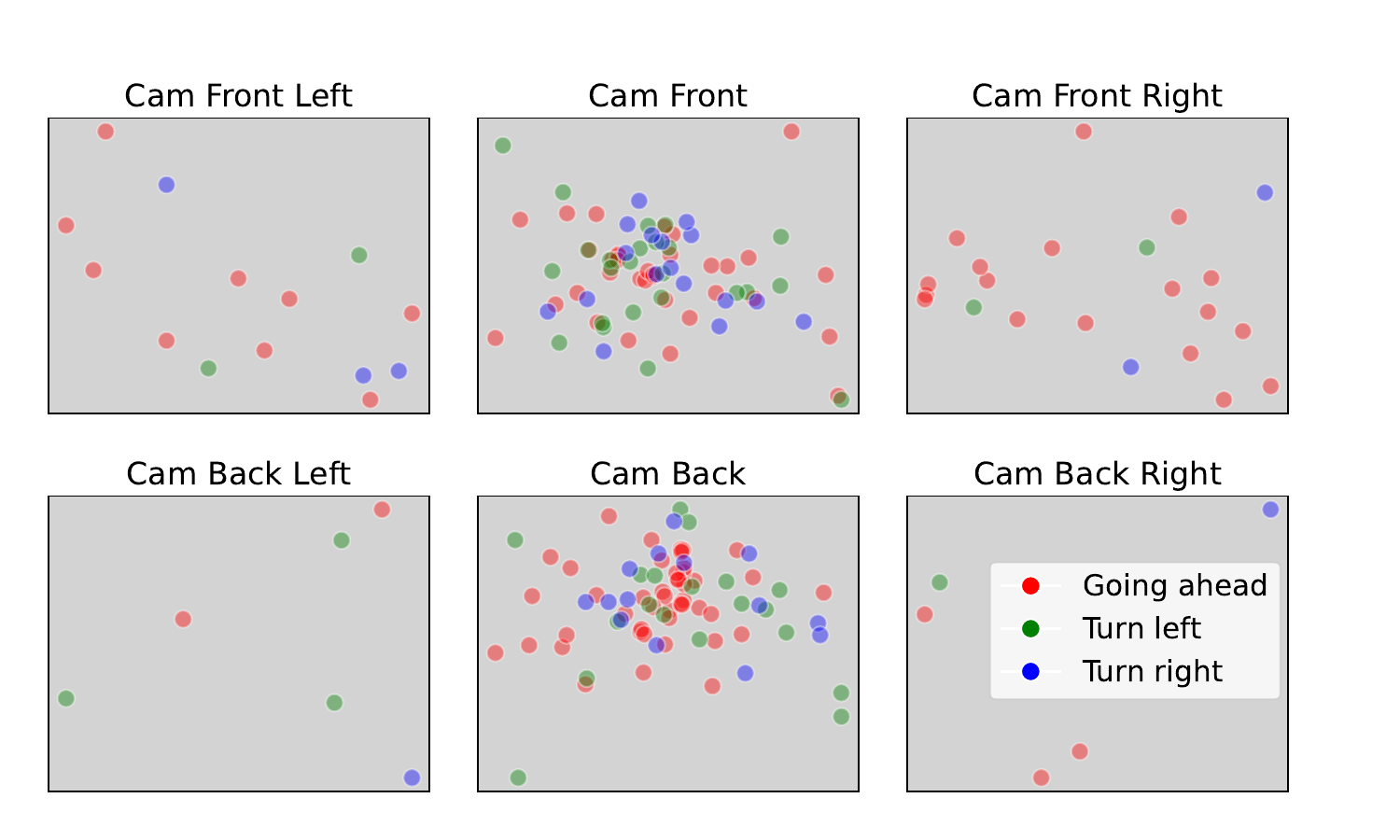}
        \vspace{-0.6cm}
        \caption{Phi-3.5 \cite{abdin2024phi}}
    \end{subfigure}
    \hfill
    \begin{subfigure}[b]{0.48\linewidth}
        \centering
        \includegraphics[width=\linewidth]{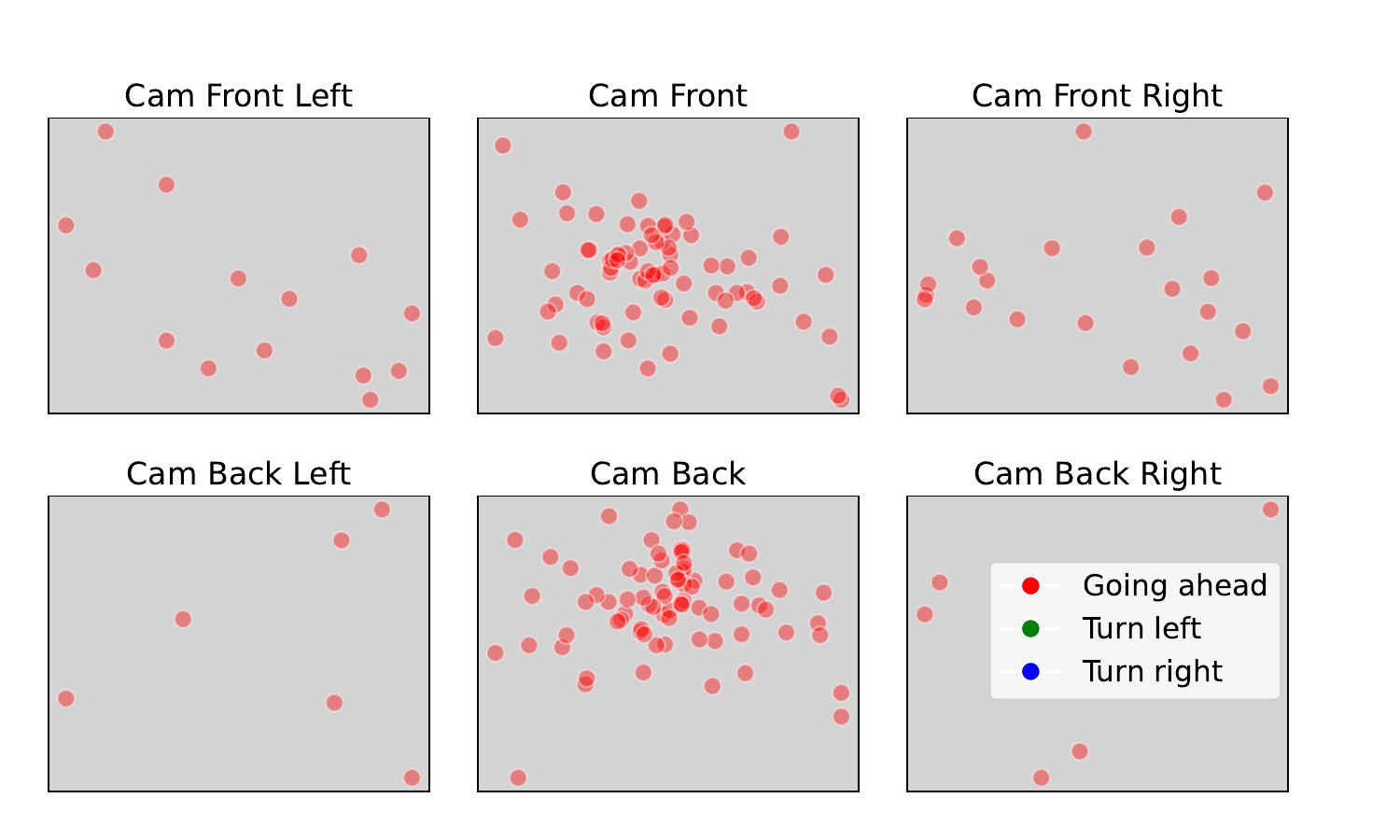}
        \vspace{-0.6cm}
        \caption{LLaVA-1.5$_\text{7B}$ \cite{liu2024llava1.5}}
        \label{fig:llava1.5-7b-spatial}
    \end{subfigure}
    \hfill
    \begin{subfigure}[b]{0.48\linewidth}
        \centering
        \includegraphics[width=\linewidth]{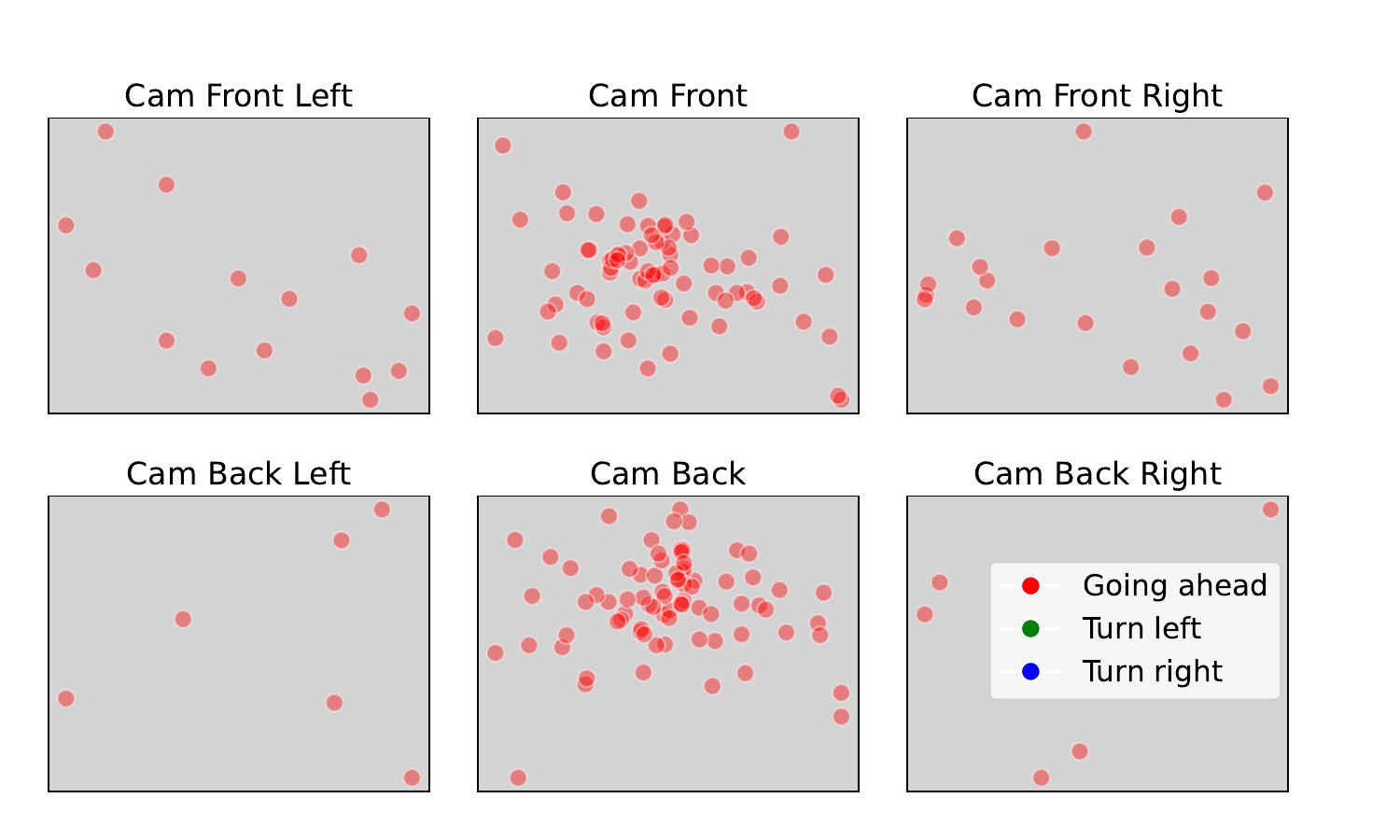}
        \vspace{-0.6cm}
        \caption{LLaVA-1.5$_\text{13B}$ \cite{liu2024llava1.5}}
        \label{fig:llava1.5-13b-spatial}
    \end{subfigure}
    \hfill
    \begin{subfigure}[b]{0.48\linewidth}
        \centering
        \includegraphics[width=\linewidth]{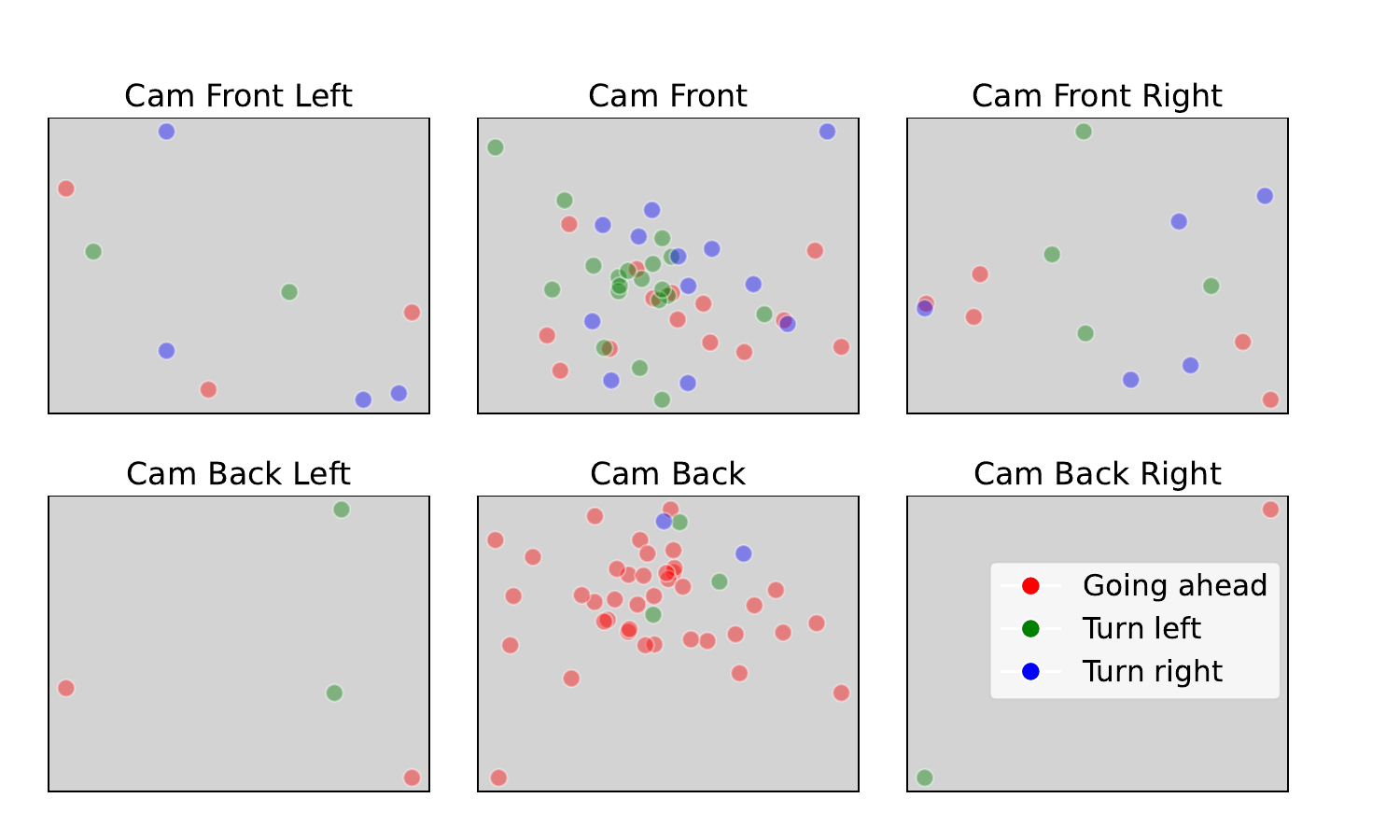}
        \vspace{-0.6cm}
        \caption{Qwen2-VL$_\text{72B}$ \cite{wang2024qwen2}}
    \end{subfigure}
    \hfill
    \begin{subfigure}[b]{0.48\linewidth}
        \centering
        \includegraphics[width=\linewidth]{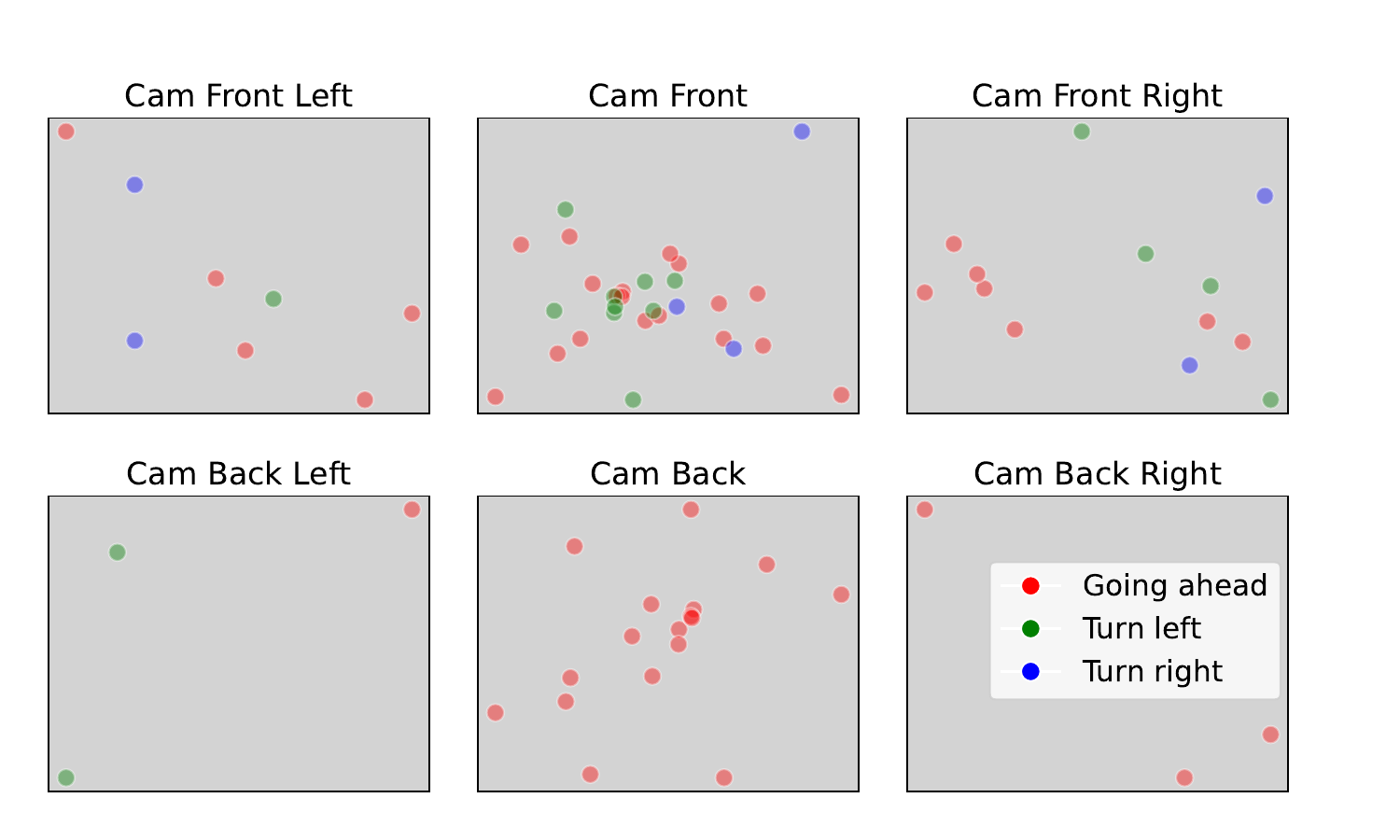}
        \vspace{-0.6cm}
        \caption{InterVL$_\text{8B}$ \cite{chen2023internvl}}
    \end{subfigure}
    \vspace{-0.2cm}
    \caption{Prediction spatial distributions from VLMs. The locations represent the object positions in the image within each camera, which is input to the model as a text description. We only visualize the data point where the model response aligns with the provided multiple choices (\eg, ``Going ahead'', ``Turn Left'', and ``Turn Right'').}
    \label{fig:camera_distribution}
\end{figure*}

We include accuracy scores for MCQs in addition to GPT scores in \cref{tab:perception-mcq-acc} and \cref{tab:behavior-mcq-acc}. Compared with GPT scores, we find that the accuracy score metric is more homogeneous. For example, the LLaVA-1.5 models have $50\%$ accuracy under all the input types, suggesting they are merely output \textit{``Going Ahead"} for perception MCQs, which is also observed in the prediction spatial distribution in Fig.~\ref{fig:llava1.5-7b-spatial} and \ref{fig:llava1.5-13b-spatial}. Moreover, we find that most models have no accuracy degradation under corruptions or even text-only inputs. This raises concerns about whether VLMs are indeed leveraging visual information to make decisions about the specified spatial location or naively guessing based on their general knowledge.

\subsection{ROUGE-L Scores}

We also present the detailed language scores, \ie, ROUGE-L~\cite{lin2004rouge} here for open-ended questions in \cref{tab:prediction-open-rougel} and \cref{tab:planning-open-rougel}. As discussed in the main paper, we find the fine-tuning process can significantly benefit the ROUGE-L score, as indicated by the fact that DriveLM \cite{sima2023drivelm} output performs other models with a large margin. On the contrary, GPT-4o \cite{achiam2023gpt}, which generates more detailed answers punished by the answer length. The ROUGE-L score of GPT-4o \cite{achiam2023gpt} is lower than most of the models even though the GPT scores are much higher.

\begin{figure*}[t]
    \begin{widecodebox}
    Please evaluate the multiple-choice answer on a scale from $0$ to $100$, where a higher score reflects precise alignment with the correct answer and well-supported reasoning. Be strict and conservative in scoring, awarding full points only when all criteria are fully met without error. Deduct points for minor inaccuracies, omissions, or lack of clarity. Distribute the \textbf{Total Score} across the following criteria:
    \\ \\
    \textbf{1. Answer Correctness ($\mathbf{50}$ points)}: \\
    ~~~~- Exact Match ($50$ points): Assign $50$ points if the predicted answer exactly matches the correct answer.
    \\
    - No Match ($0$ points): Assign $0$ points if the predicted answer does not match the correct answer, regardless of explanation quality.
    \\ \\
    \textbf{2. Object Recognition ($\mathbf{10}$ points):}\\
    - Award up to $5$ points for accurately identifying all relevant object(s) in the scene.
    \\
    - Award up to $5$ points for correct descriptions of the identified object(s), including attributes like colors, materials, sizes, or shapes.
    \\
    - Guideline: Deduct points for any missing, misidentified, or irrelevant objects, particularly if they are crucial to the driving context. Deduct points if any important visual details are missing, incorrect, or overly generalized, especially if they affect comprehension or recognition.
    \\ \\    
    \textbf{3. Object Location and Orientation ($\mathbf{15}$ points):} \\
    - Score up to $5$ points for a precise description of the object's location, orientation, or position relative to the ego vehicle.
    \\
    - Award up to $5$ points for acknowledging environmental factors, such as lighting, visibility, and other conditions that influence perception.
    \\
    - Score up to $5$ points based on how well the answer reflects an understanding of situational context, such as obstacles, traffic flow, or potential hazards.
    \\
    - Guideline: Deduct points for inaccuracies or omissions in spatial information that could affect scene understanding. Deduct points if the answer fails to consider factors impacting object visibility or situational awareness. Deduct points for overlooked or misinterpreted contextual factors that may impact driving decisions.
    \\ \\ 
    \textbf{4. Environmental Condition Awareness ($\mathbf{15}$ points):} \\
    - Award up to $15$ points if the explanation considers environmental conditions (e.g., weather or sensor limitations) that could impact perception.
    \\
    - Guideline: Deduct points if relevant environmental conditions are ignored or inadequately addressed.
    \\ \\
    \textbf{5. Clarity of Reasoning ($\mathbf{10}$ points):} \\
    - Award up to $5$ points for clear, logically structured reasoning that is easy to understand.
    \\
    - Assign up to $5$ points for grammatical accuracy and coherent structure.
    \\
    - Guideline: Deduct points for vague or confusing explanations that hinder comprehension. Deduct points for grammar or syntax issues that impact clarity or logical flow.
    \\ \\
    Assign $0$ points from criteria $2$ to $5$ if no explanation is provided.
    \\ \\ 
    Here is the multiple-choice question: \textcolor{robo_red}{\texttt{\textbf{QUESTION}}}
    \\
    Here is the ground truth object visual description: \textcolor{robo_red}{\texttt{\textbf{DESC}}}
    \\
    Here is the correct answer: \textcolor{robo_red}{\texttt{\textbf{GT}}}
    \\
    Here is the predicted answer and explanation (if any): \textcolor{robo_red}{\texttt{\textbf{PRED}}}
    \\ \\
    \textbf{Please fill in the following scoring sheet, and then provide a brief summary supporting the score:}\\ \\
    1. Answer Correctness ($50$ points):\\ 
    2. Object Recognition ($10$ points):\\
    3. Object Location and Orientation ($15$ points):\\
    4. Environmental Condition Awareness ($15$ points):\\
    5. Clarity of Reasoning ($10$ points):
    \\ \\
    \textcolor{robo_blue}{\textbf{Total Score}}: 
    \\
    \textcolor{robo_blue}{\textbf{Brief Summary}}:
    \end{widecodebox}
    \vspace{-0.2cm}
\caption{GPT evaluation prompts for \textbf{MCQs} in our benchmark.}
\label{fig:gpt-eval-prompt-mcq}
\vspace{0.2cm}
\end{figure*}
\begin{figure*}[t]
    \begin{widecodebox}
    Please evaluate the predicted answer on a scale from $0$ to $100$, where a higher score reflects precise alignment with the correct answer and well-supported reasoning. Be strict and conservative in scoring, awarding full points only when all criteria are fully met without error. Deduct points for minor inaccuracies, omissions, or lack of clarity. Distribute the \textbf{Total Score} across the following criteria:
    \\ \\
    \textbf{1. Action Alignment ($\mathbf{20}$ points):} \\
    - Assign up to $20$ points based on how accurately the predicted action (e.g., forward, turn left, turn right) matches the correct answer.\\
    - Guideline: Award full points only for exact matches or highly similar actions. Deduct points for any inaccuracies or missing elements. Assign $0$ points if no action prediction is provided.
    \\ \\
    \textbf{2. Motion Precision ($\mathbf{20}$ points):} \\
    - Award up to $20$ points based on how closely the predicted motion (e.g., speed up, decelerate) aligns with the correct motion in the answer.\\
    - Guideline: Deduct points if the predicted motion fails to match the type or intensity of the correct answer. Ensure that the intended speed or deceleration aligns accurately with the driving context. Assign $0$ points if no motion prediction is provided.
    \\ \\
    \textbf{3. Driving Context Appropriateness ($\mathbf{15}$ points):} \\
    - Score up to $15$ points for the relevance of the predicted answer to the driving context implied by the correct answer, emphasizing logical alignment with the situation.\\
    - Guideline: Award higher scores only if the answer fully reflects an accurate understanding of the driving context. Deduct points if the action or motion is illogical or does not align with the scenario's requirements.
    \\ \\
    \textbf{4. Situational Awareness ($\mathbf{15}$ points):} \\
    - Award up to $15$ points for demonstrated awareness of environmental factors (e.g., traffic participants, obstacles) relevant to the action or motion.\\
    - Guideline: Deduct points if the answer misses key situational details that may lead to unsafe or incorrect predictions.
    \\ \\
    \textbf{5. Conciseness and Clarity ($\mathbf{20}$ points):} \\
    - Assess the clarity and brevity of the predicted answer. Answers should be concise, clear, and easy to understand, effectively communicating the intended actions and motions.\\
    - Guideline: Deduct points for verbosity, ambiguity, or lack of focus that could hinder quick comprehension.
    \\ \\
    \textbf{6. Grammar ($\mathbf{10}$ points):} \\
    - Evaluate the grammatical accuracy and structure of the answer. Assign up to $5$ points for clarity and logical flow, and up to $5$ points for grammatical accuracy.\\
    - Guideline: Deduct points for grammar or syntax issues that reduce readability or coherence.
    \\ \\
    Here is the predicted answer: \textcolor{robo_red}{\texttt{\textbf{PRED}}}
    \\ \\
    Here is the correct answer: \textcolor{robo_red}{\texttt{\textbf{GT}}}
    \\ \\
    \textbf{Please fill in the following scoring sheet, and then provide a brief summary supporting the score:}\\ \\
    1. Action Alignment ($20$ points): \\ 
    2. Motion Precision ($20$ points): \\
    3. Driving Context Appropriateness ($15$ points): \\
    4. Situational Awareness ($15$ points): \\
    5. Conciseness and Clarity ($20$ points): \\ 
    6. Grammar ($10$ points): \\ \\
    \textcolor{robo_blue}{\textbf{Total Score}}: 
    \\
    \textcolor{robo_blue}{\textbf{Brief Summary}}:
    \end{widecodebox}
    \vspace{-0.2cm}
\caption{GPT evaluation prompts for \textbf{Open-Ended Questions} in our benchmark.}
\label{fig:gpt-eval-prompt-open}
\vspace{1cm}
\end{figure*}

\clearpage

\begin{figure*}[t]
    \centering
    \begin{subfigure}[t]{0.33\linewidth}
        \centering
        \includegraphics[width=\linewidth]{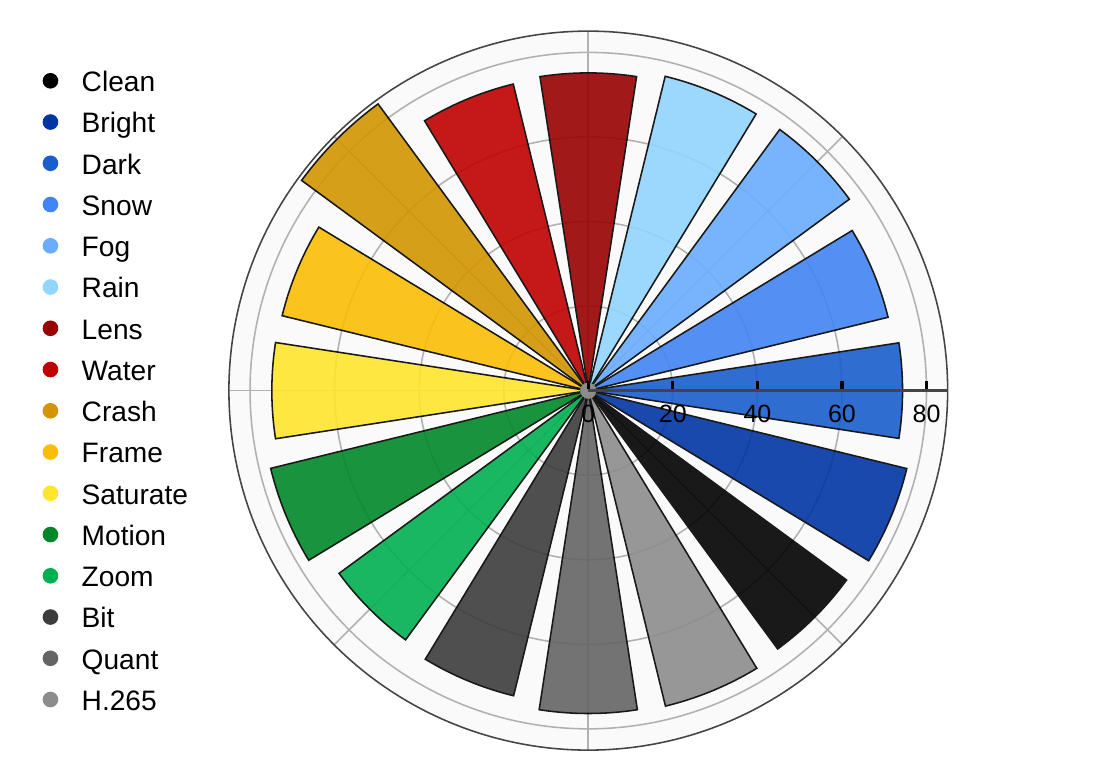}
        \caption{GPT-4o}
        \label{fig:gpt4o-radar}
    \end{subfigure}
    \begin{subfigure}[t]{0.33\linewidth}
        \centering
        \includegraphics[width=\linewidth]{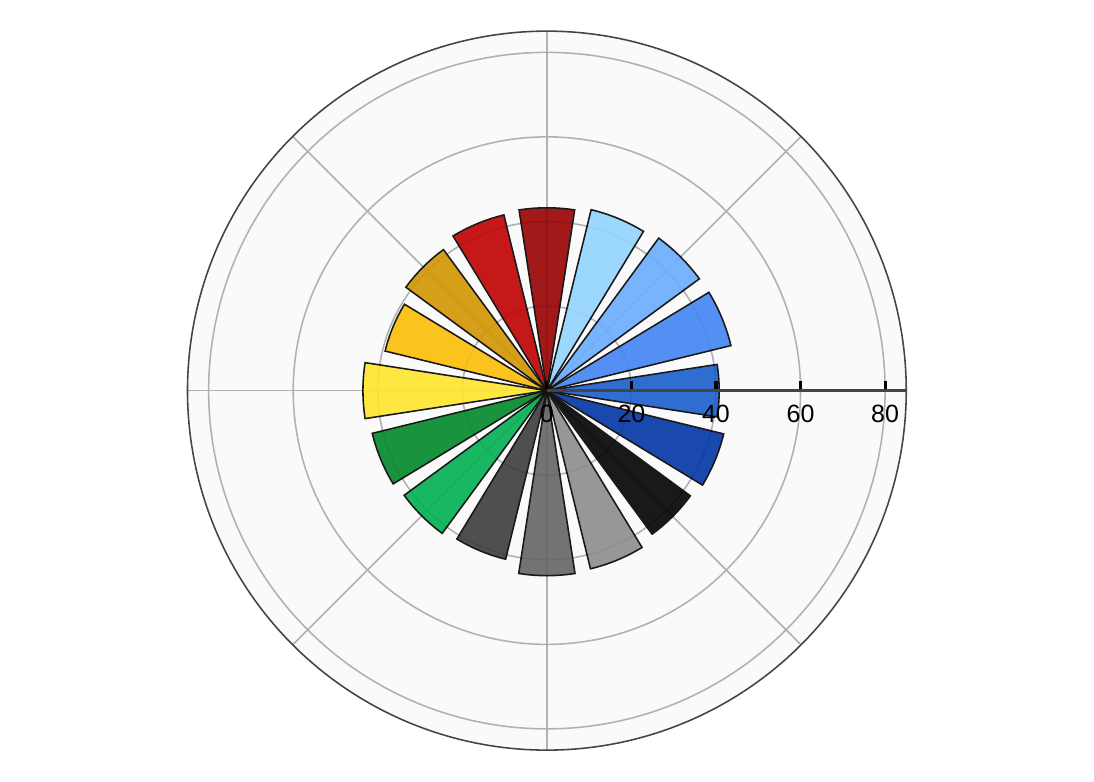}
        \caption{Phi3}
        \label{fig:phi3-radar}
    \end{subfigure}
    \begin{subfigure}[t]{0.33\linewidth}
        \centering
        \includegraphics[width=\linewidth]{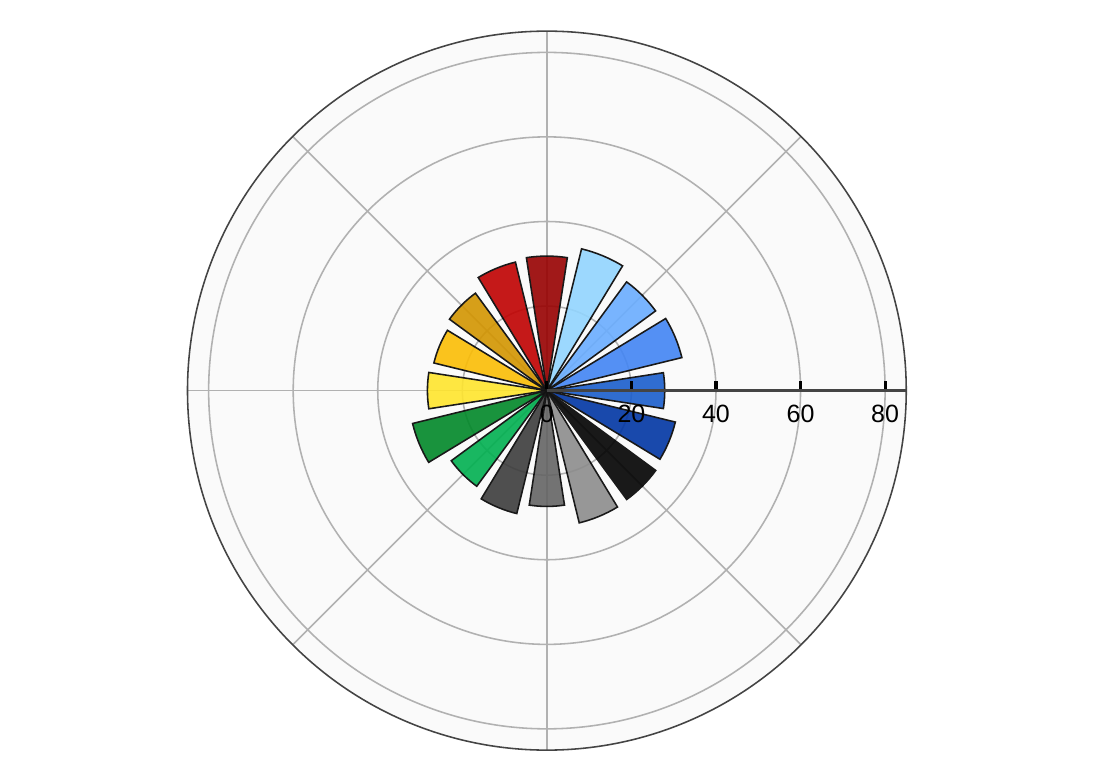}
        \caption{Phi-3.5}
        \label{fig:phi3.5-radar}
    \end{subfigure}

    \vspace{0.5cm} 
    
    \begin{subfigure}[t]{0.33\linewidth}
        \centering
        ~~~\includegraphics[width=\linewidth]{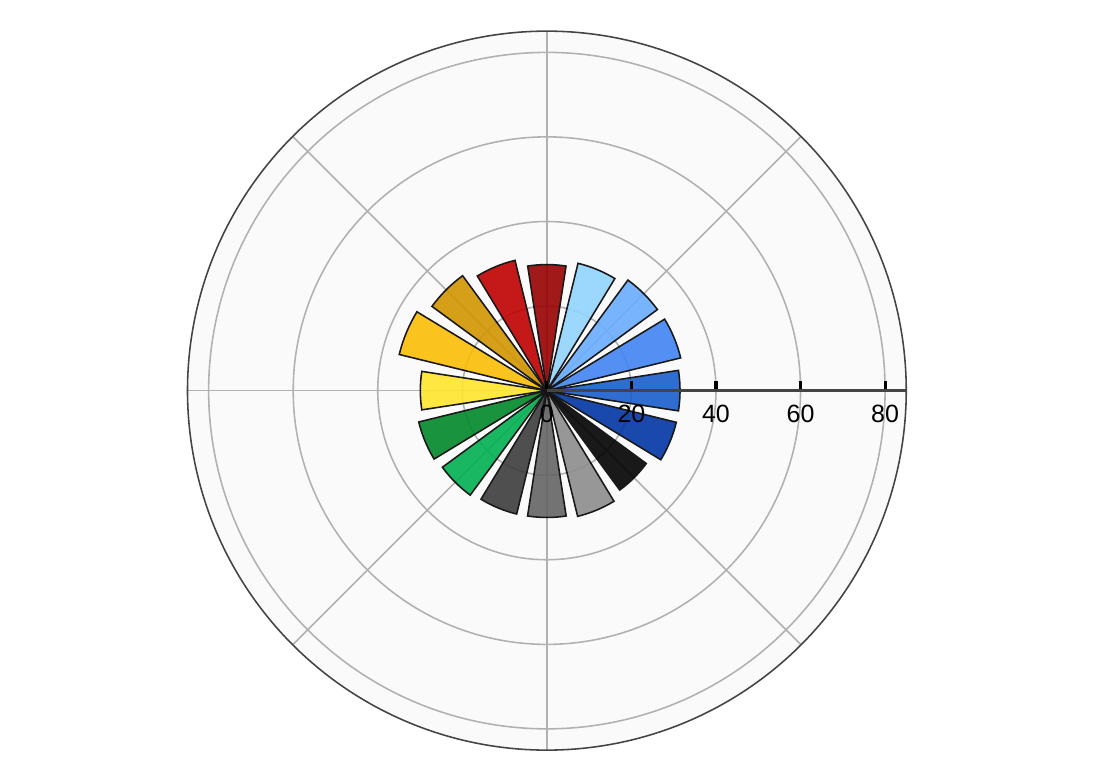}
        \caption{LLaVA-1.5\textsubscript{7B}}
        \label{fig:llava1.5-7b-radar}
    \end{subfigure}
    \begin{subfigure}[t]{0.33\linewidth}
        \centering
        \includegraphics[width=\linewidth]{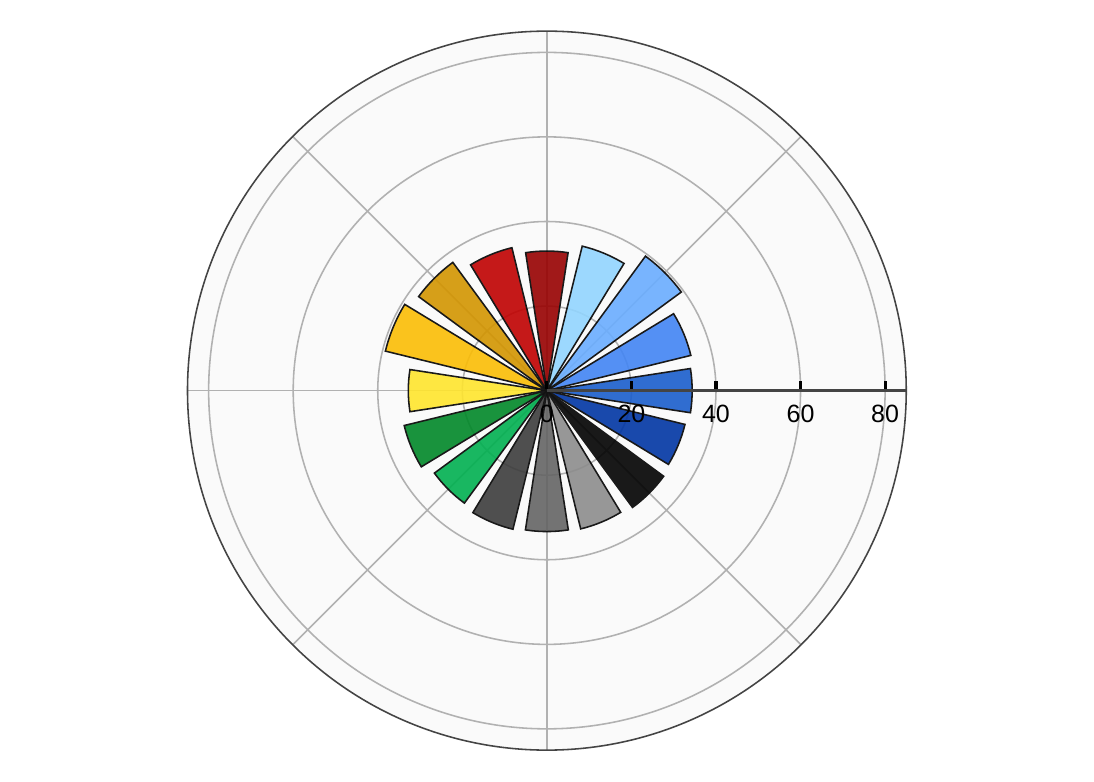}
        \caption{LLaVA-1.5\textsubscript{13B}}
        \label{fig:llava1.5-13b-radar}
    \end{subfigure}
    \begin{subfigure}[t]{0.33\linewidth}
        \centering
        \includegraphics[width=\linewidth]{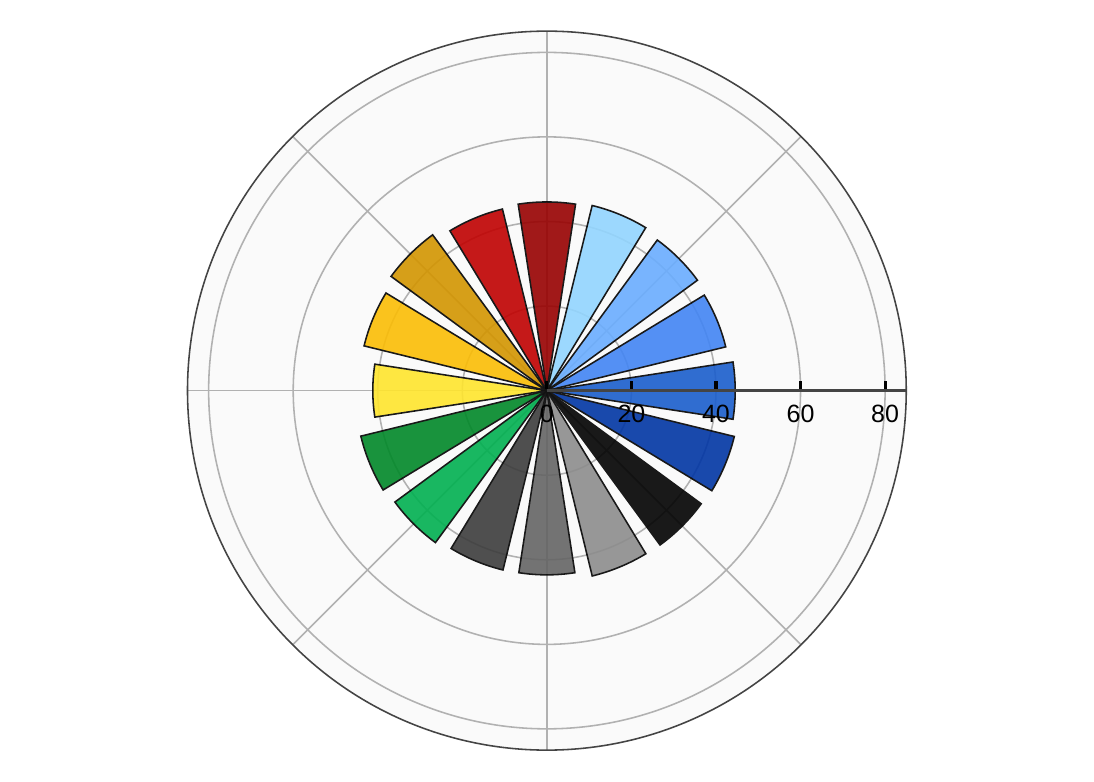}
        \caption{LLaVA-NeXT}
        \label{fig:llava1.6-7b-radar}
    \end{subfigure}

    \vspace{0.5cm} 
    
    \begin{subfigure}[t]{0.33\linewidth}
        \centering
        ~~~\includegraphics[width=\linewidth]{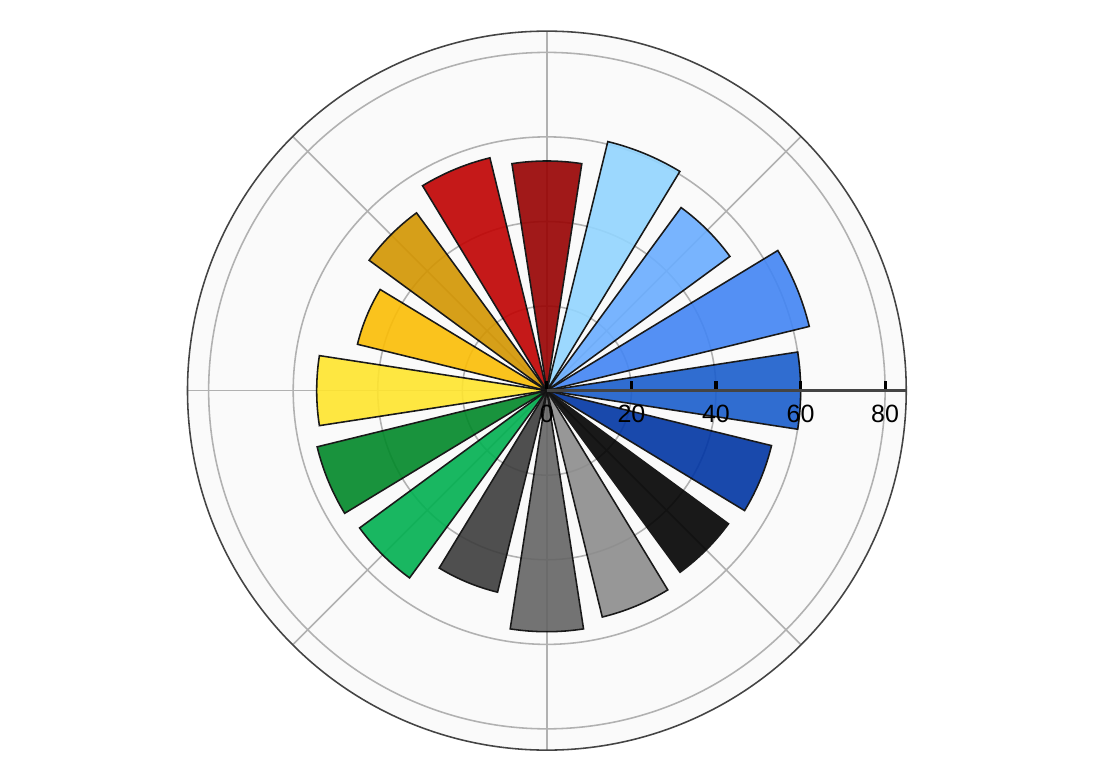}
        \caption{InternVL\textsubscript{8B}}
        \label{fig:internvl-radar}
    \end{subfigure}
    \begin{subfigure}[t]{0.33\linewidth}
        \centering
        \includegraphics[width=\linewidth]{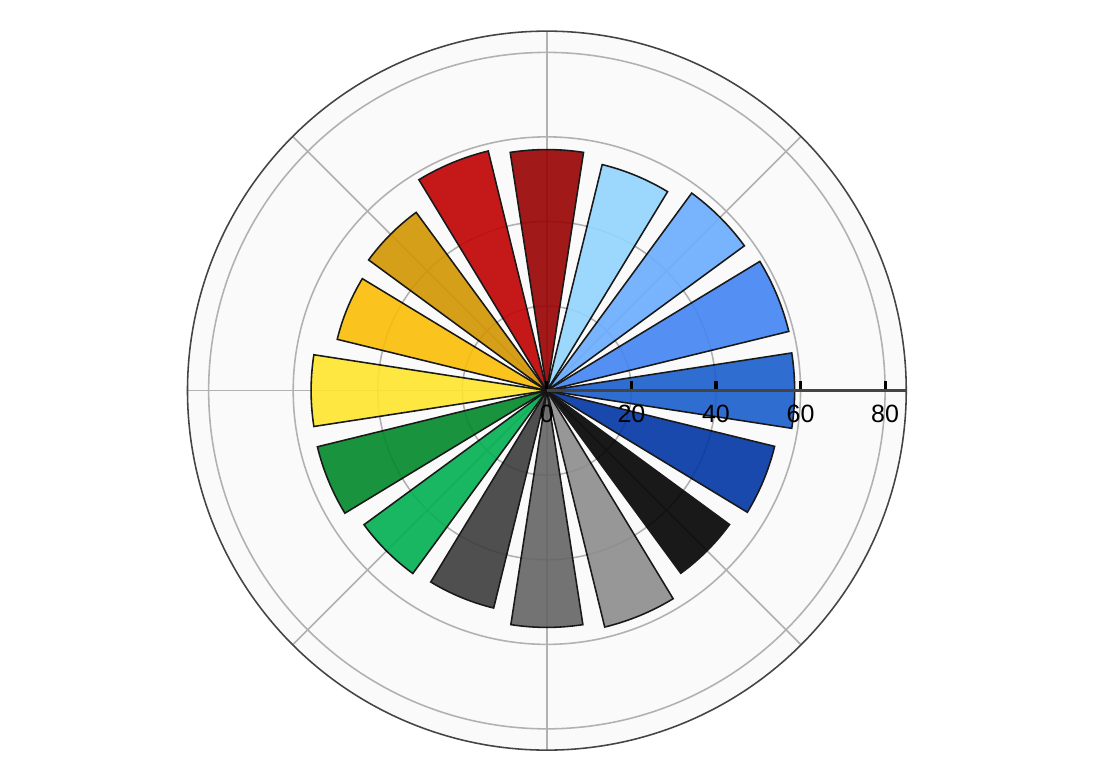}
        \caption{Oryx}
        \label{fig:oryx-radar}
    \end{subfigure}
    \begin{subfigure}[t]{0.33\linewidth}
        \centering
        \includegraphics[width=\linewidth]{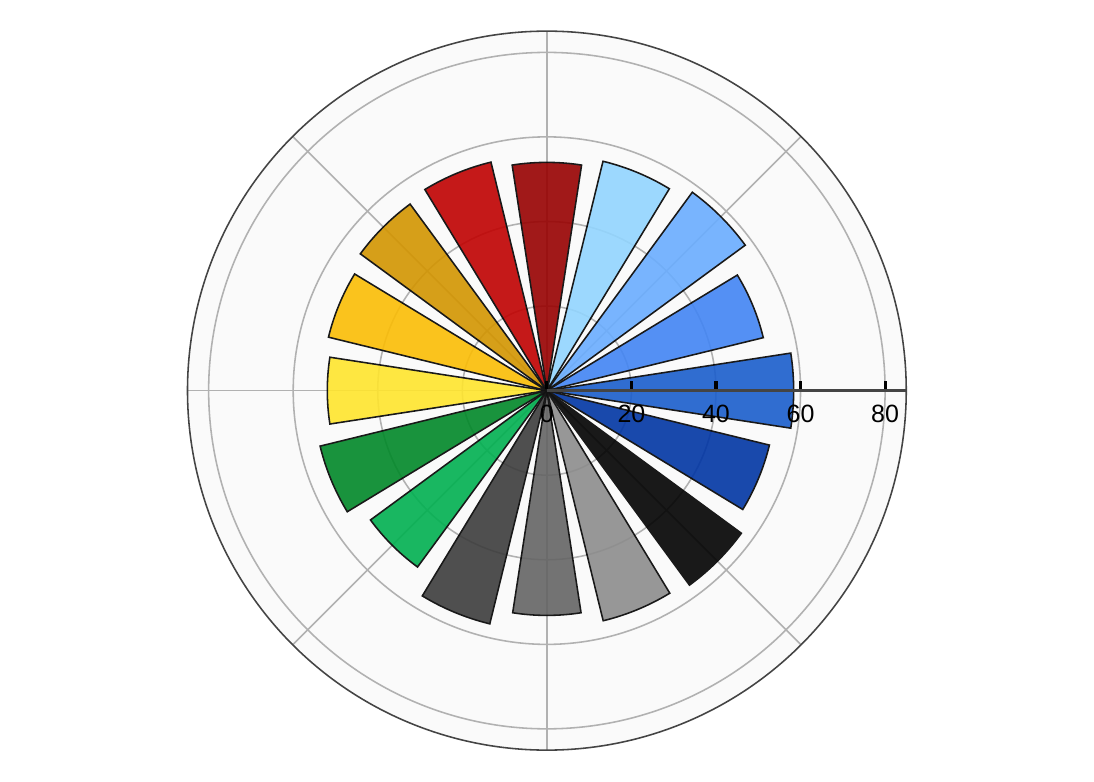}
        \caption{Qwen2VL\textsubscript{7B}}
        \label{fig:qwen2vl-7b-radar}
    \end{subfigure}

    \vspace{0.5cm} 

    \begin{subfigure}[t]{0.33\linewidth}
        \centering
        ~~~\includegraphics[width=\linewidth]{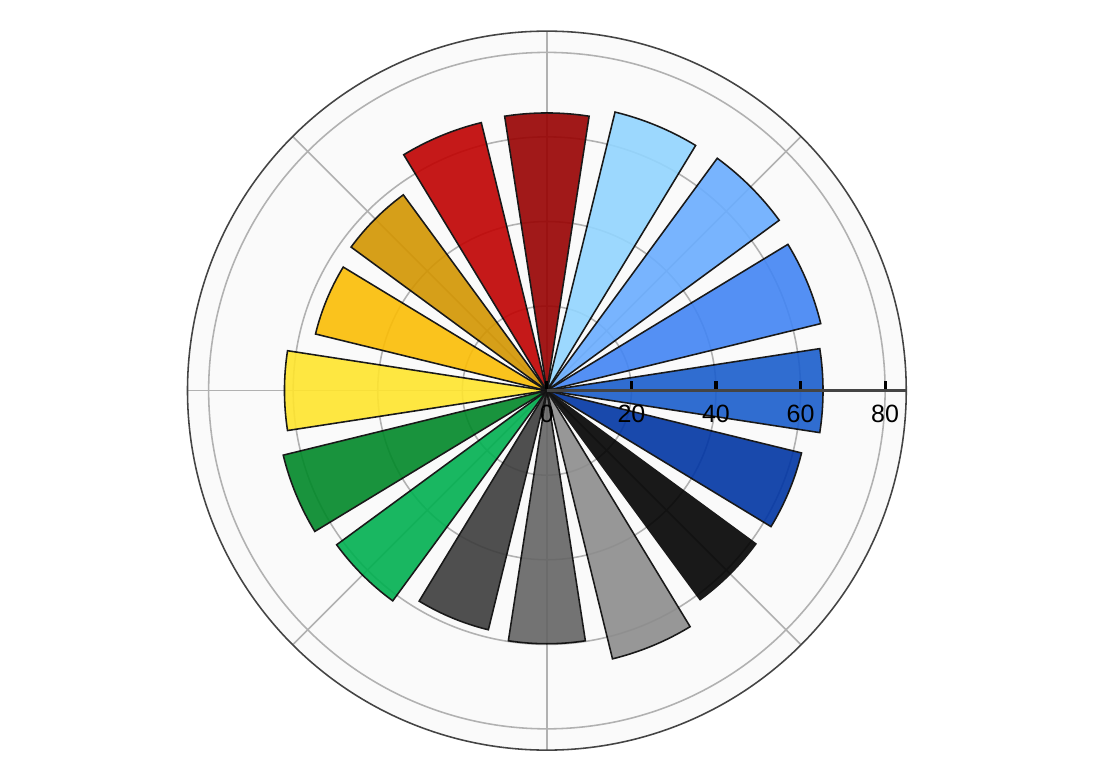}
        \caption{Qwen2VL\textsubscript{72B}}
        \label{fig:qwen2vl-72b-radar}
    \end{subfigure}
    \begin{subfigure}[t]{0.33\linewidth}
        \centering
        \includegraphics[width=\linewidth]{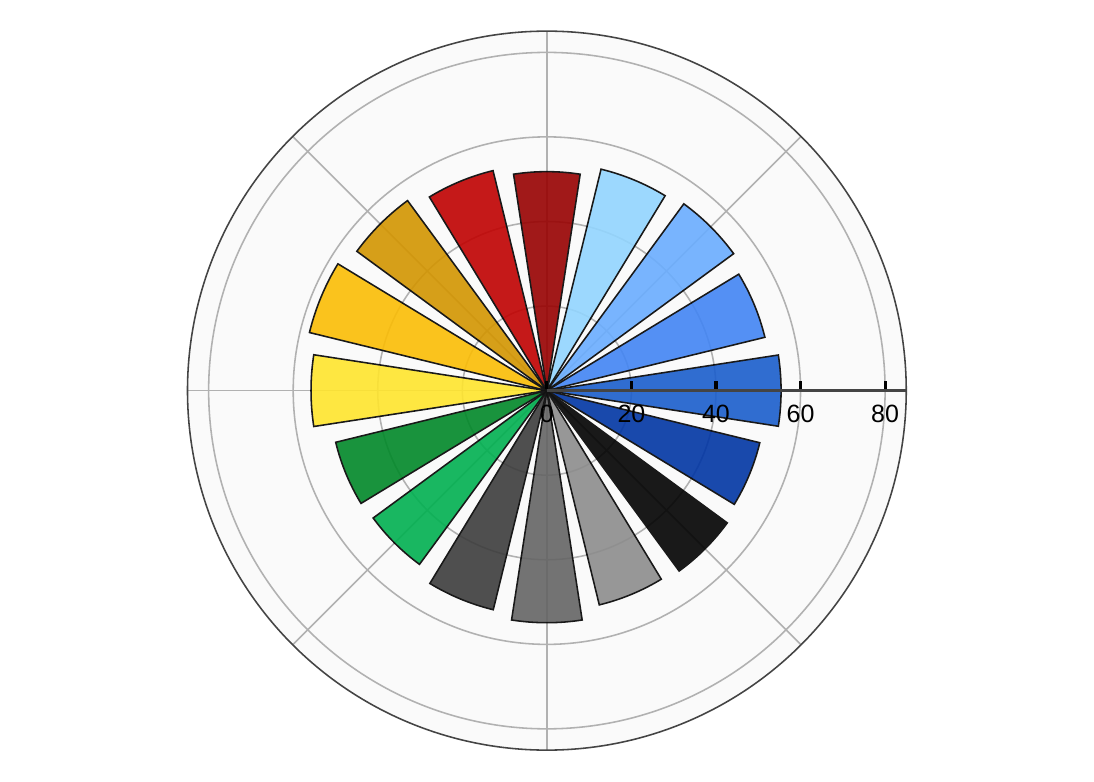}
        \caption{Dolphin}
        \label{fig:dolphin-radar}
    \end{subfigure}
    \begin{subfigure}[t]{0.33\linewidth}
        \centering
        \includegraphics[width=\linewidth]{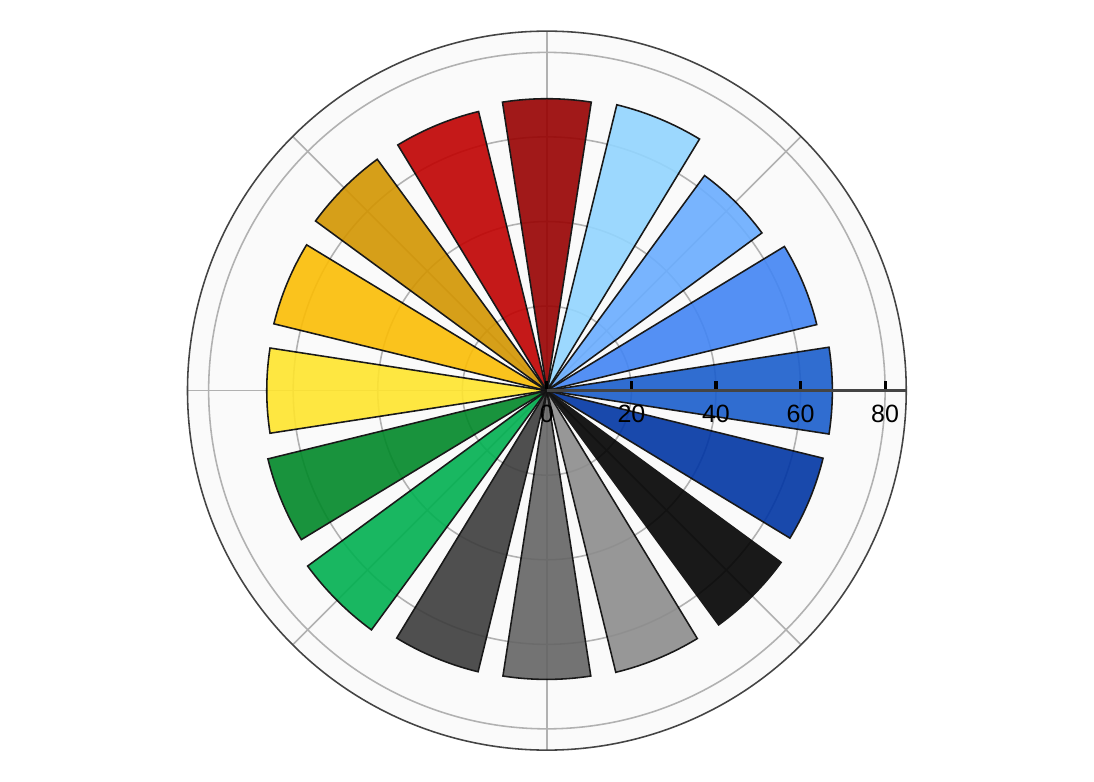}
        \caption{DriveLM}
        \label{fig:drivelm-radar}
    \end{subfigure}
    \caption{\textbf{Model performance comparisons using radar graphs}. The performance for each input corruption type is averaged across all the $1,261$ questions spanning four different tasks using GPT scores. The gray dash line represents the performance of \textcolor{robo_red}{\textbf{text-only}} input. We observe VLMs have subtle performance changes under corruptions. For some models, the GPT scores under only text input are even higher than the performance when the visual information is available.}
    \label{fig:model-comparison}
\end{figure*}




\clearpage
\begin{figure*}
    \centering
    \includegraphics[width=\linewidth]{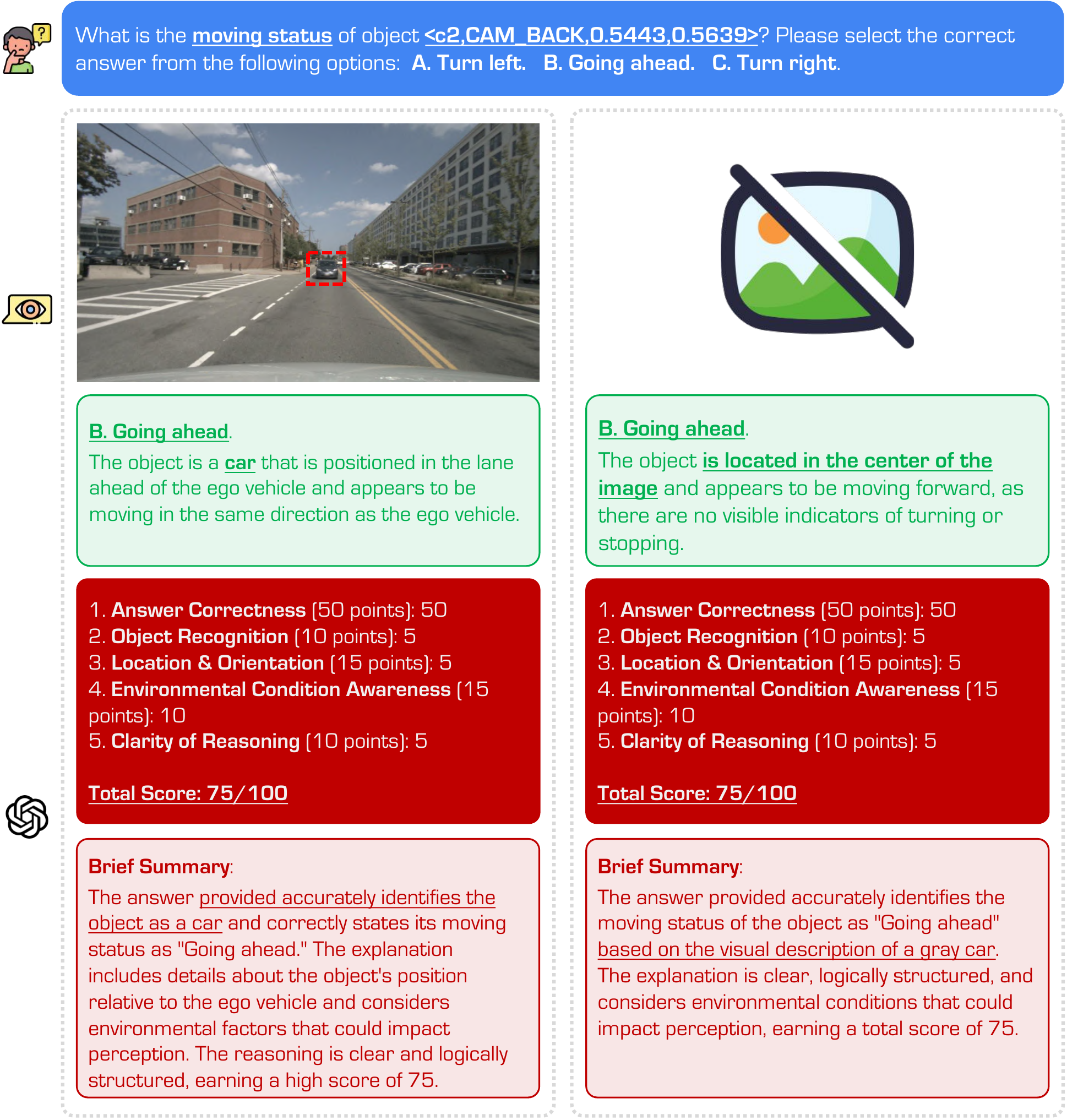}
    \vspace{-0.6cm}
    \caption{Example of Qwen2-VL$_\text{72B}$ for the \includegraphics[width=0.026\linewidth]{figures/icons/perception.png}~\textbf{Perception} task. Under the \textbf{\textcolor{robo_red}{text-only condition}} (right example), the model can fabricate answers based on the coordinate information in the question. The GPT evaluator can not reflect the fabricated answer even with detailed rubrics, resulting in the same score as prompted with a clean image.}
    \label{fig:example-1}
\end{figure*}

\clearpage
\begin{figure*}
    \centering
    \includegraphics[width=\linewidth]{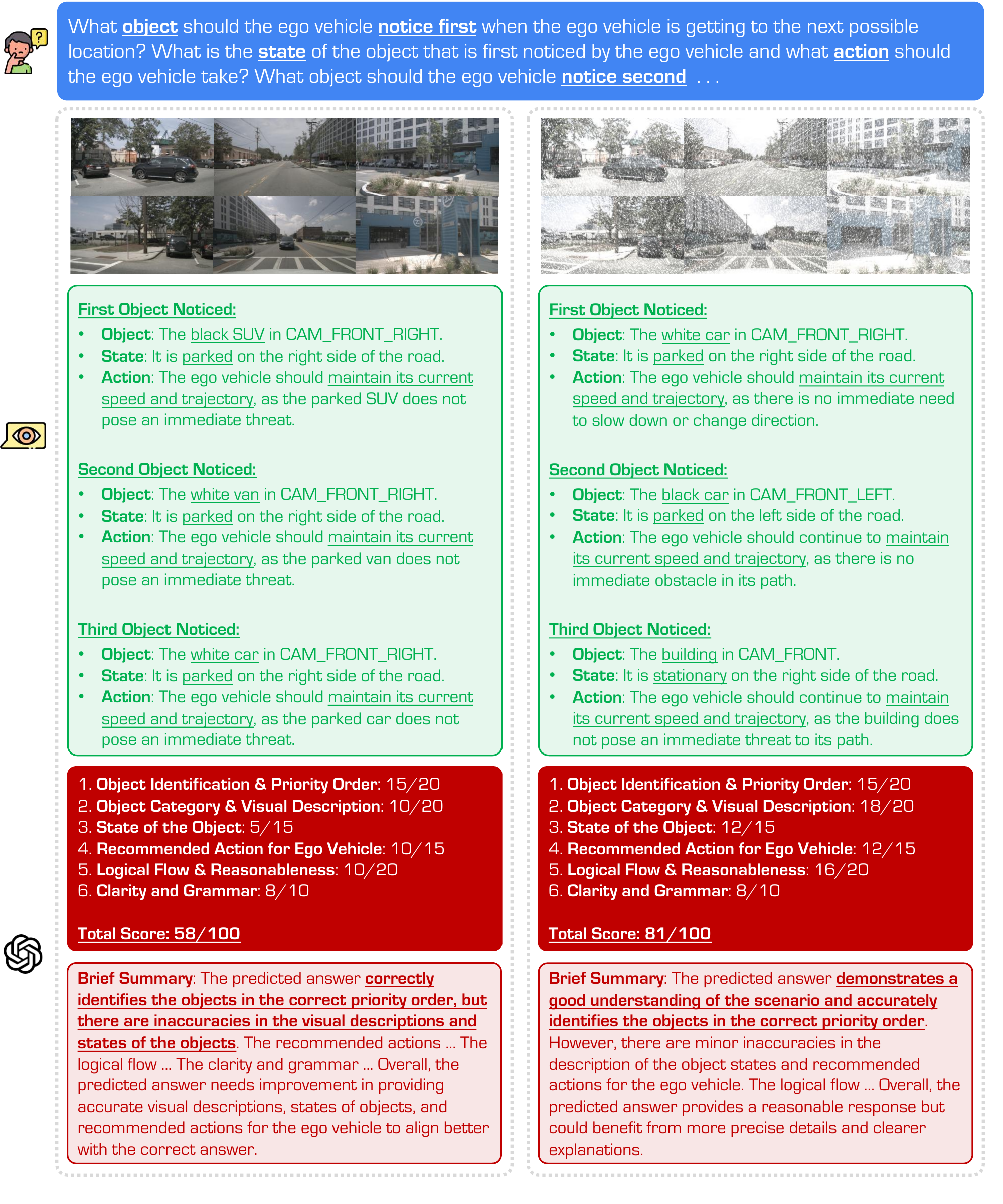}
    \vspace{-0.6cm}
    \caption{Example of Qwen2-VL$_\text{72B}$ for the \includegraphics[width=0.026\linewidth]{figures/icons/prediction.png}~\textbf{Prediction} task. Under the \textbf{\textcolor{robo_red}{snow condition}} (right example), the model can still observe some visible objects in the images. Since most objects are severely occluded, the third object noticed becomes more general: the building, compared to the clean inputs, where all the objects are visible vehicles. However, the GPT score is even higher. The example illustrates that even the GPT evaluator can not accurately reflect the language quality generated by VLMs.}
    \label{fig:example-2}
\end{figure*}

\clearpage
\begin{figure*}
    \centering
    \includegraphics[width=\linewidth]{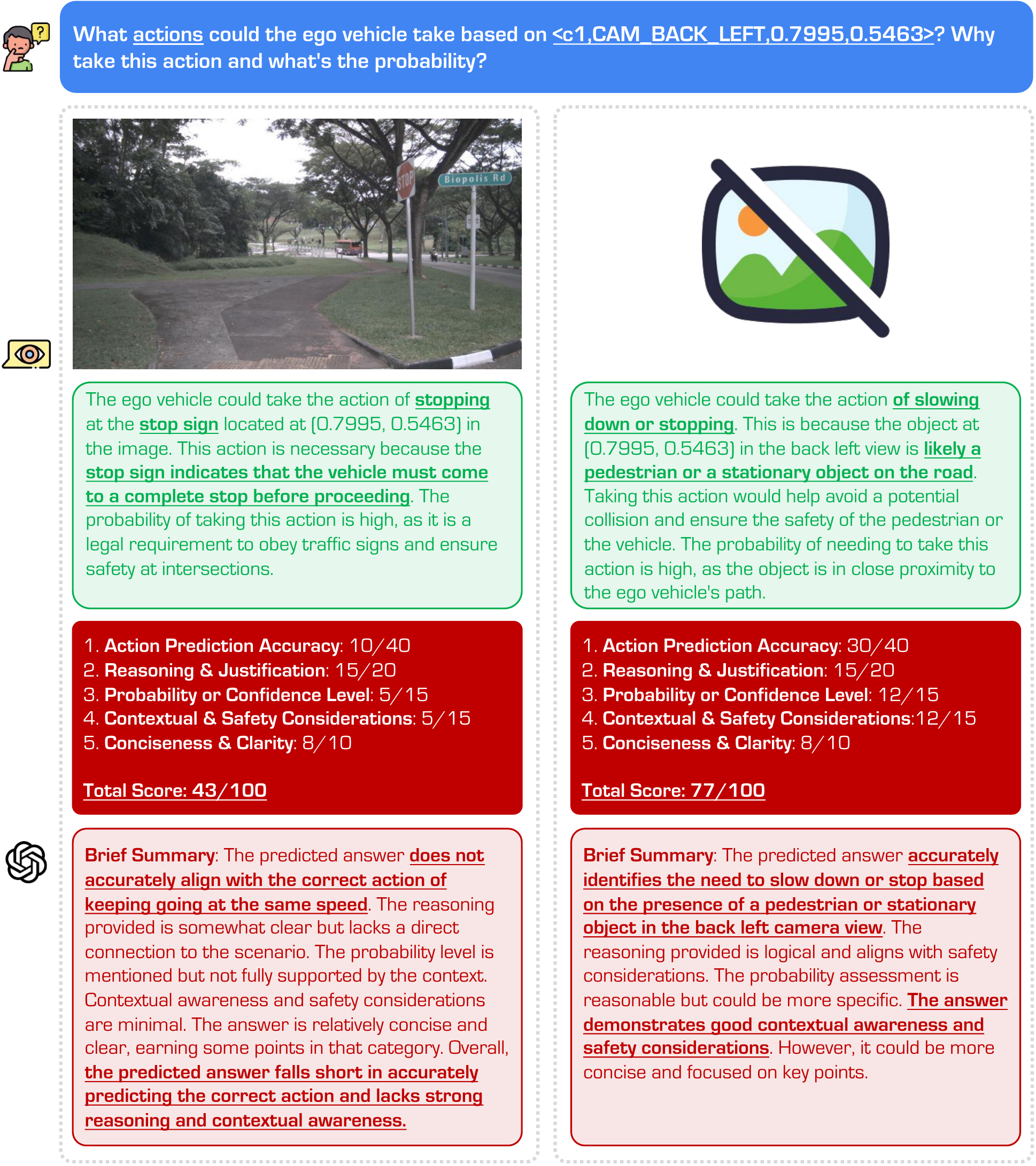}
    \vspace{-0.6cm}
    \caption{Example of Qwen2-VL$_\text{72B}$ for \includegraphics[width=0.026\linewidth]{figures/icons/planning.png}~\textbf{Planning} task. Under the \textbf{\textcolor{robo_red}{text-only condition}} (right example), the model fails to perceive the motorcycle to stop sign, giving their close distance on the image. When the image is absent, the model guesses the object to be a pedestrian or a stationary object on the road, unexpectedly leading to an answer with higher scores.}
    \label{fig:example-3}
\end{figure*}

\clearpage
\begin{figure*}
    \centering
    \includegraphics[width=\linewidth]{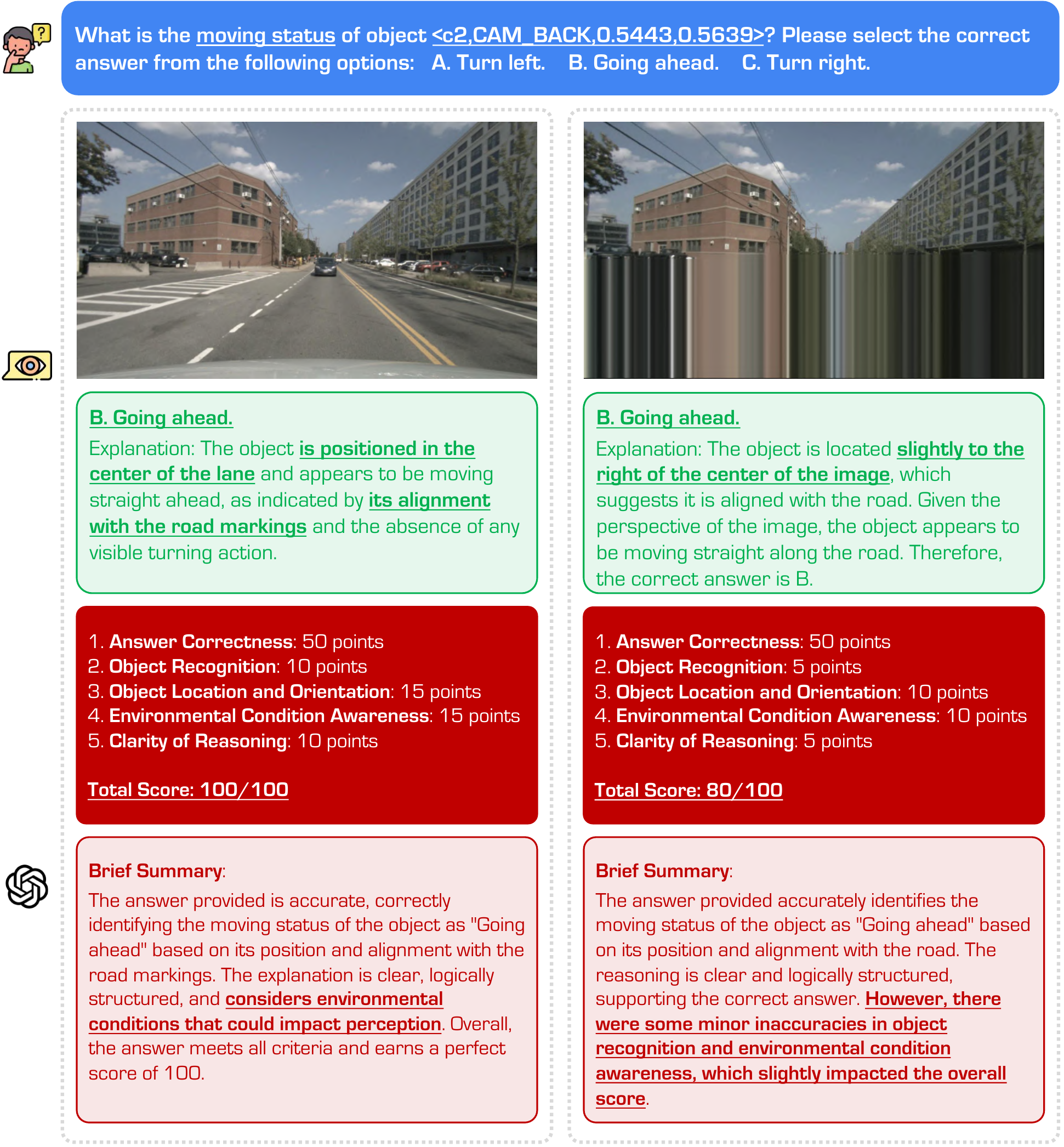}
    \vspace{-0.6cm}
    \caption{Example of GPT-4o for \includegraphics[width=0.026\linewidth]{figures/icons/perception.png}~\textbf{Perception} task. Under the \textbf{\textcolor{robo_red}{bit error condition}} (right example), the visual information for the target object is completely corrupted but the model can guess the answer correctly. However, the GPT evaluator here captures the nuance in the explanation and assigns a higher score to the more detailed answer with clean image inputs.}
    \label{fig:example-4}
\end{figure*}

\clearpage
\begin{figure*}
    \centering
    \includegraphics[width=\linewidth]{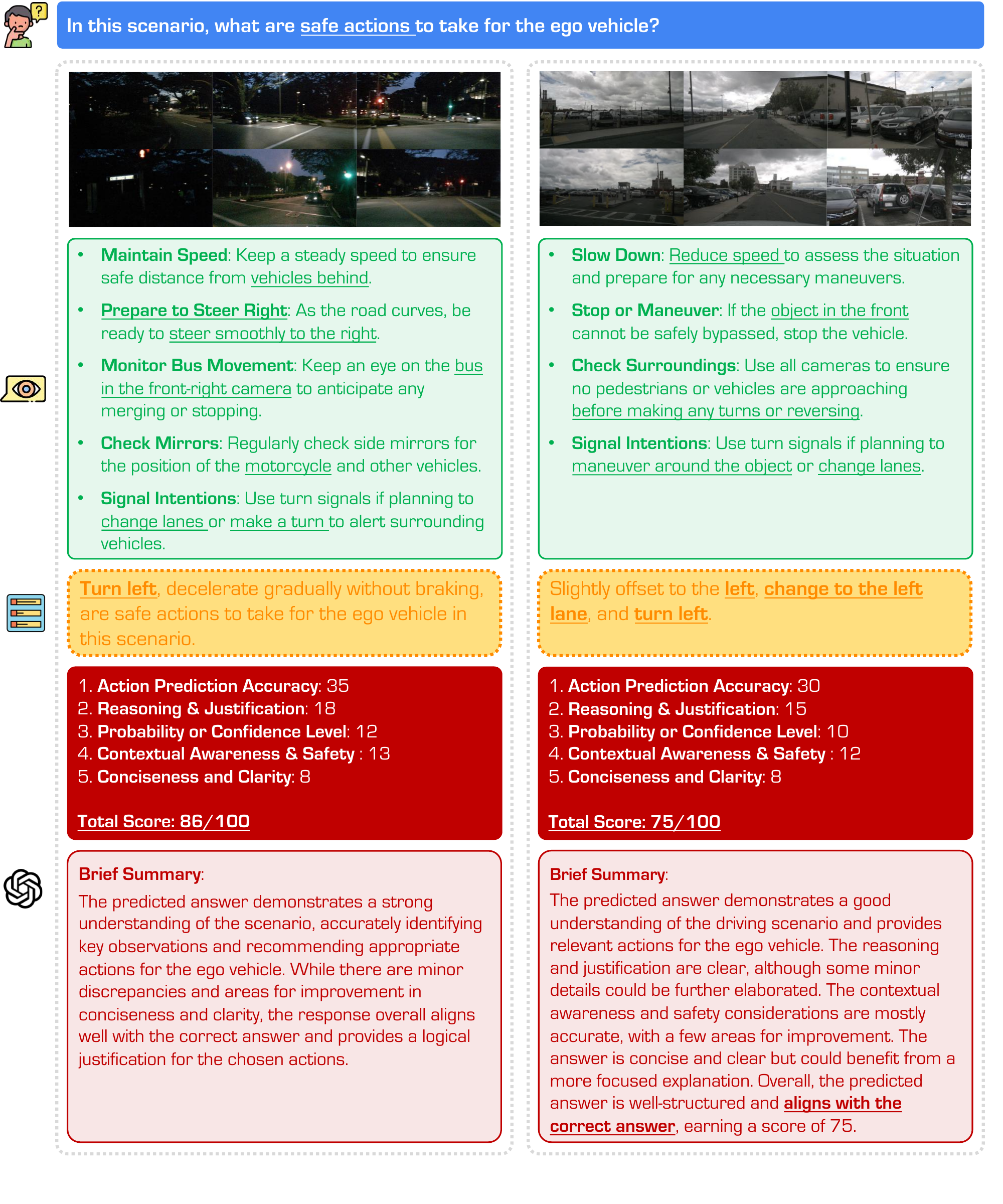}
    \vspace{-0.6cm}
    \caption{Example of GPT-4o for \includegraphics[width=0.026\linewidth]{figures/icons/planning.png}~\textbf{Planning} task. In both \textbf{\textcolor{robo_red}{low-light}} (left example) and \textbf{\textcolor{robo_green}{normal}} (right example) conditions, the ground truth is to turn left. However, in the left case, the vehicle is prepared to turn left. Turning right will cause a potential collision with the white sedan on the front right camera. On the right image, the road is empty and the predicted slow-down action will not cause safety issues. However, the GPT evaluator assigns a higher score on the left case, ignoring the potential safety consequences as there are no high-fidelity physical context prompts.}
\end{figure*}

\clearpage
\begin{figure*}
    \centering
    \includegraphics[width=\linewidth]{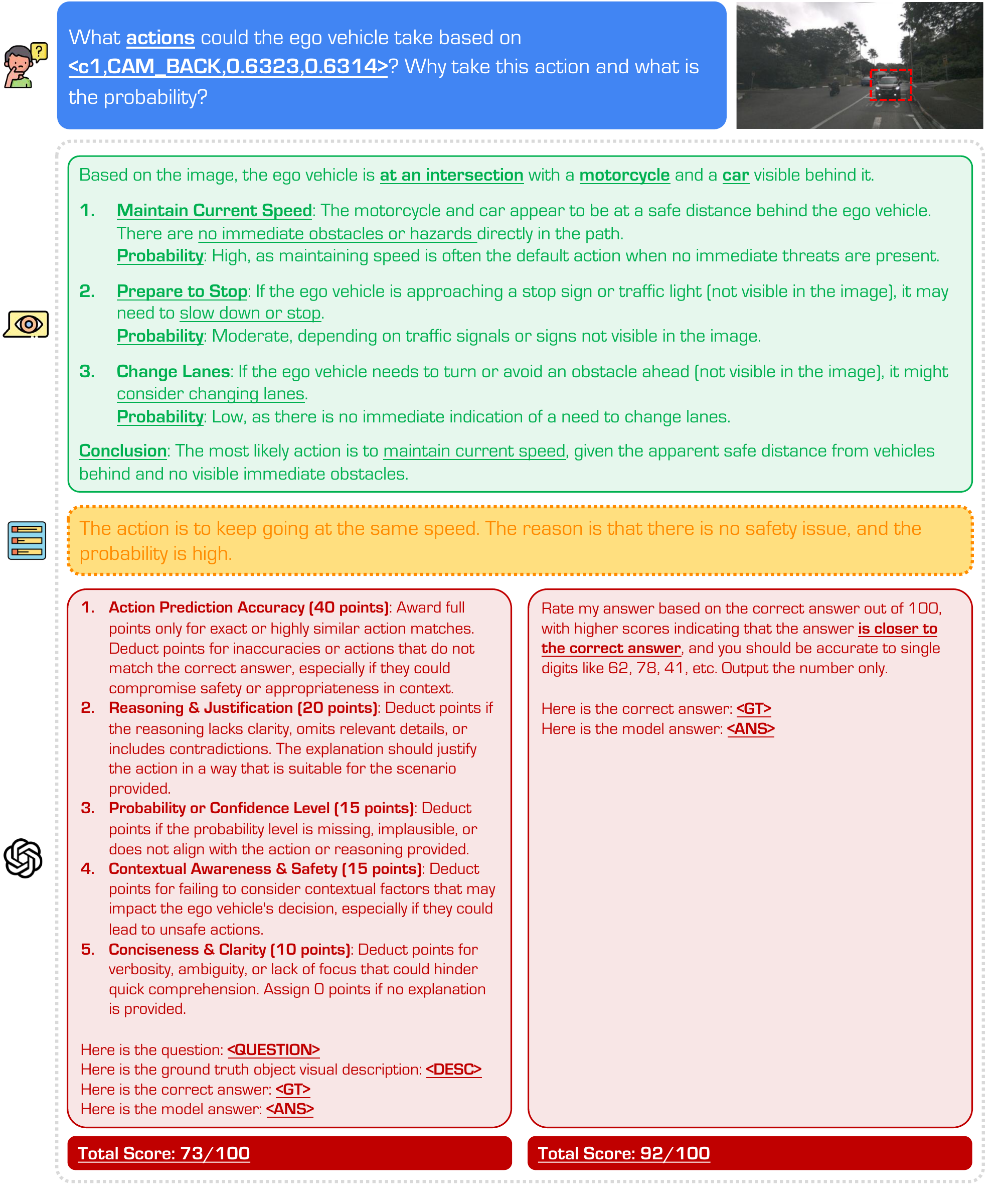}
    \vspace{-0.6cm}
    \caption{Examples of the GPT evaluator with different prompts. With the same question and answer, the GPT evaluator assigns different scores based on the rubrics. Without detailed information on the physical context, the scoring point, the evaluator fails to capture critical information for driving decisions in the answer but simply based on semantic similarities.}
    \label{fig:example-6}
\end{figure*}

\clearpage
\begin{figure*}
    \centering
    \includegraphics[width=\linewidth]{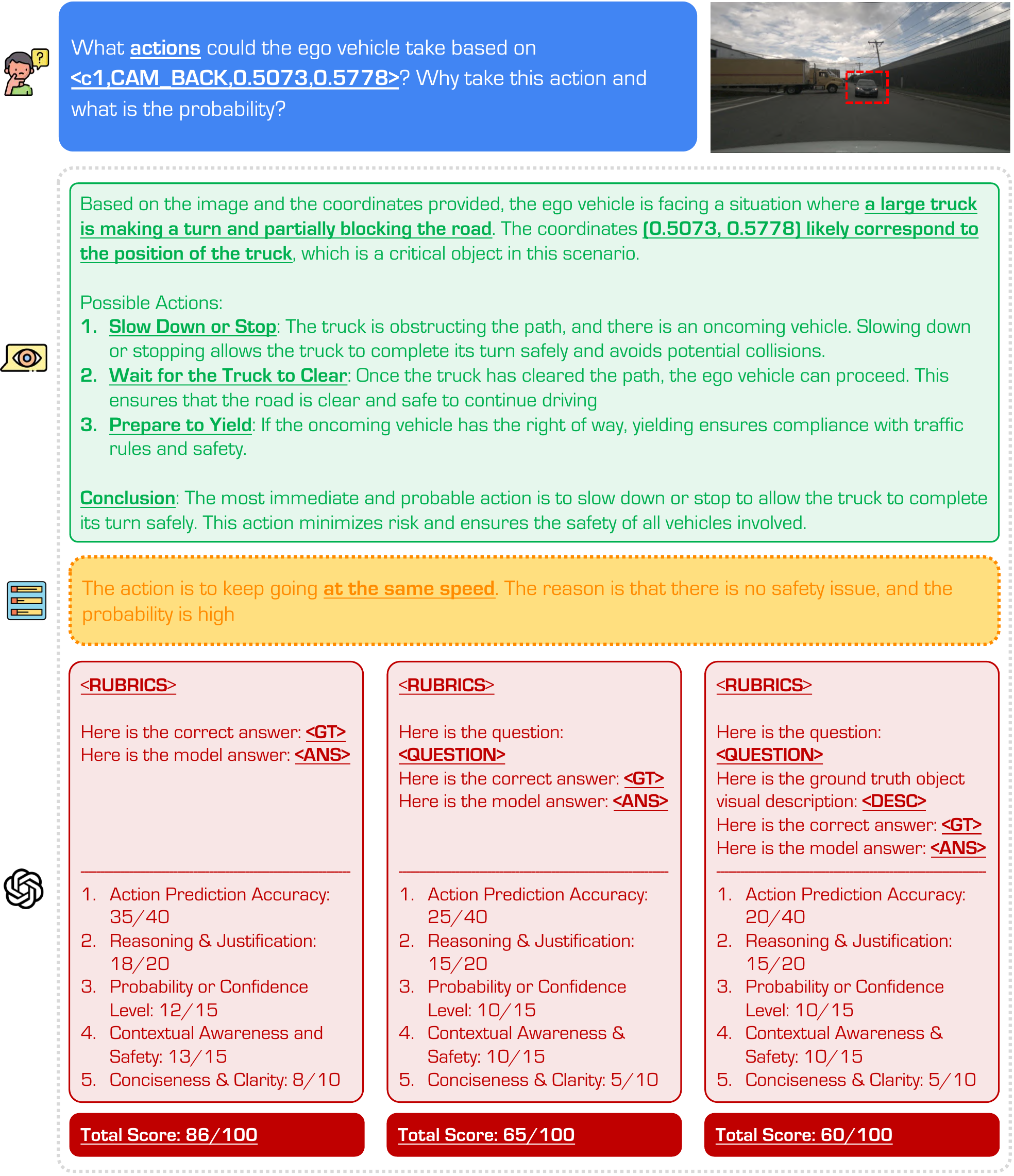}
    \vspace{-0.6cm}
    \caption{Examples of the GPT evaluators with different information. We gradually add more information about the question and the visual description of the target objects. The evaluator gives a more accurate score based on more information.}
    \label{fig:exmaple-prompt-detailed}
\end{figure*}

\clearpage
\begin{table*}[t]
\centering
\caption{Detailed GPT score results of \textbf{MCQs} for the \includegraphics[width=0.026\linewidth]{figures/icons/perception.png}~\textbf{Perception} task. ``\textcolor{robo_green}{Clean}'' represents clean image inputs. ``\textcolor{robo_blue}{T.O.}'' represents text-only evaluation. The ``\textcolor{robo_red}{Corrupt}'' settings range from weather conditions, external disturbances, sensor failures, motion blur, and transmission errors. The benchmarked VLMs include commercial, open-sourced, and driving specialist models, respectively.}
\vspace{-0.1cm}
\resizebox{\linewidth}{!}{
    \begin{tabular}{r|c|c|ccccc|cc|ccc|cc|ccc}
    \toprule
    \textbf{Method} & \rotatebox{90}{\textcolor{robo_green}{$\bullet$~Clean~}} & \rotatebox{90}{\textcolor{robo_blue}{$\bullet$~T.O.}} & \rotatebox{90}{\textcolor{robo_red}{$\bullet$}~Brightness~} & \rotatebox{90}{\textcolor{robo_red}{$\bullet$}~Dark~} & \rotatebox{90}{\textcolor{robo_red}{$\bullet$}~Snow~} & \rotatebox{90}{\textcolor{robo_red}{$\bullet$}~Fog~} & \rotatebox{90}{\textcolor{robo_red}{$\bullet$}~Rain~} & \rotatebox{90}{\textcolor{robo_red}{$\bullet$}~Lens Obstacle~} & \rotatebox{90}{\textcolor{robo_red}{$\bullet$}~Water Splash~} & \rotatebox{90}{\textcolor{robo_red}{$\bullet$}~Camera Crash~} & \rotatebox{90}{\textcolor{robo_red}{$\bullet$}~Frame Lost~} & \rotatebox{90}{\textcolor{robo_red}{$\bullet$}~Saturate~} & \rotatebox{90}{\textcolor{robo_red}{$\bullet$}~Motion Blur~} & \rotatebox{90}{\textcolor{robo_red}{$\bullet$}~Zoom Blur~} & \rotatebox{90}{\textcolor{robo_red}{$\bullet$}~Bit Error~} & \rotatebox{90}{\textcolor{robo_red}{$\bullet$}~Color Quant~} & \rotatebox{90}{\textcolor{robo_red}{$\bullet$}~H.265 Compression~~}
    \\
    \midrule\midrule
    \rowcolor{robo_green!10} \includegraphics[width=0.025\linewidth]{figures/icons/human.png}~\textcolor{robo_green}{Human} & \textcolor{robo_green}{$47.67$} & \textcolor{robo_green}{-} & \textcolor{robo_green}{$43.33$} & \textcolor{robo_green}{$26.67$} & \textcolor{robo_green}{$45.00$} & \textcolor{robo_green}{$18.33$} & \textcolor{robo_green}{$40.00$} & \textcolor{robo_green}{$35.33$} & \textcolor{robo_green}{$37.33$} & \textcolor{robo_green}{$42.00$} & \textcolor{robo_green}{$20.00$} & \textcolor{robo_green}{$25.00$} & \textcolor{robo_green}{$43.00$} & \textcolor{robo_green}{$20.00$} & \textcolor{robo_green}{$33.67$} & \textcolor{robo_green}{$25.33$} & \textcolor{robo_green}{$31.33$}
    \\
    \midrule
    \textcolor{gray}{GPT-4o} & \textcolor{gray}{$41.87$} & \textcolor{gray}{$43.59$} & \textcolor{gray}{$43.84$} & \textcolor{gray}{$44.82$} & \textcolor{gray}{$45.18$} & \textcolor{gray}{$44.30$} & \textcolor{gray}{$46.20$} & \textcolor{gray}{$45.69$} & \textcolor{gray}{$44.10$} & \textcolor{gray}{$38.12$} & \textcolor{gray}{$40.08$} & \textcolor{gray}{$39.32$} & \textcolor{gray}{$41.40$} & \textcolor{gray}{$36.67$} & \textcolor{gray}{$37.54$} & \textcolor{gray}{$38.22$} & \textcolor{gray}{$39.37$}
    \\
    \midrule
    \rowcolor{robo_gray!10}Phi-3 & $35.51$ & $32.65$ & $35.28$ & $34.15$ & $38.88$ & $38.22$ & $37.70$ & $38.39$ & $36.75$ & $34.93$ & $33.89$ & $37.53$ & $37.80$ & $39.72$ & $36.49$ & $37.15$ & $37.03$ 
    \\
    Phi-3.5 & $40.22$ & $38.33$ & $39.22$ & $36.46$ & $41.41$ & $43.04$ & $41.40$ & $40.66$ & $40.83$ & $37.59$ & $36.91$ & $39.86$ & $40.33$ & $42.30$ & $36.71$ & $41.49$ & $41.48$ \\
    \rowcolor{robo_gray!10}LLaVA-1.5$_\text{7B}$ & $32.40$ & $32.68$ & $32.48$ & $32.95$ & $31.95$ & $32.43$ & $32.30$ & $32.88$ & $32.18$ & $32.93$ & $31.63$ & $32.50$ & $32.43$ & $32.93$ & $32.48$ & $32.18$ & $32.63$ 
    \\
    LLaVA-1.5$_\text{13B}$ & $33.58$ & $33.25$ & $33.25$ & $33.25$ & $33.15$ & $33.50$ & $33.53$ & $32.95$ & $32.93$ & $33.48$ & $33.40$ & $33.25$ & $33.45$ & $33.68$ & $33.33$ & $33.25$ & $33.38$ 
    \\
    \rowcolor{robo_gray!10}LLaVA-NeXT & $32.98$ & $4.20$ & $33.85$ & $20.43$ & $11.62$ & $16.58$ & $27.33$ & $18.24$ & $32.30$ & $26.80$ & $20.83$ & $23.50$ & $34.00$ & $17.75$ & $24.50$ & $18.03$ & $26.24$
    \\
    InternVL$_\text{8B}$ & $46.60$ & $52.46$ & $43.65$ & $44.15$ & $43.58$ & $46.02$ & $42.38$ & $41.48$ & $43.38$ & $45.32$ & $49.08$ & $43.98$ & $41.30$ & $41.50$ & $38.25$ & $44.84$ & $42.13$ 
    \\
    \rowcolor{robo_gray!10}Oryx & $17.98$ & $20.87$ & $16.48$ & $16.88$ & $16.63$ & $16.79$ & $14.31$ & $16.35$ & $15.85$ & $16.49$ & $21.44$ & $21.38$ & $16.36$ & $21.04$ & $17.65$ & $18.13$ & $19.51$ 
    \\
    Qwen2VL$_\text{7B}$ & $42.64$ & $37.76$ & $43.08$ & $37.29$ & $39.72$ & $41.67$ & $40.87$ & $40.69$ & $39.89$ & $39.75$ & $39.17$ & $40.85$ & $41.32$ & $39.62$ & $34.28$ & $39.90$ & $41.20$ 
    \\
    \rowcolor{robo_gray!10}Qwen2VL$_\text{72B}$ & $38.15$ & $21.53$ & $36.24$ & $37.77$ & $35.91$ & $35.78$ & $37.13$ & $38.14$ & $38.97$ & $29.48$ & $25.63$ & $36.87$ & $36.91$ & $37.01$ & $30.90$ & $36.05$ & $41.48$ 
    \\
    \midrule
    Dolphin & $6.50$ & $8.35$ & $10.18$ & $11.08$ & $10.70$ & $9.53$ & $10.58$ & $9.93$ & $9.80$ & $10.08$ & $9.95$ & $11.20$ & $9.85$ & $10.10$ & $8.80$ & $10.00$ & $11.10$ 
    \\
    \rowcolor{robo_gray!10}DriveLM & $22.38$ & $12.45$ & $20.78$ & $25.30$ & $18.98$ & $24.43$ & $25.95$ & $22.03$ & $21.03$ & $21.95$ & $16.28$ & $19.38$ & $22.98$ & $20.93$ & $19.90$ & $16.25$ & $26.48$ 
    \\
    \bottomrule
\end{tabular}}
\label{tab:perception-mcq}
\end{table*}
\begin{table*}[t]
\centering
\caption{Detailed Accuracy score results of \textbf{MCQs} for the \includegraphics[width=0.026\linewidth]{figures/icons/perception.png}~\textbf{Perception} task. ``\textcolor{robo_green}{Clean}'' represents clean image inputs. ``\textcolor{robo_blue}{T.O.}'' represents text-only evaluation. The ``\textcolor{robo_red}{Corrupt}'' settings range from weather conditions, external disturbances, sensor failures, motion blur, and transmission errors. The benchmarked VLMs include commercial, open-sourced, and driving specialist models, respectively.}
\vspace{-0.1cm}
\resizebox{\linewidth}{!}{
    \begin{tabular}{r|c|c|ccccc|cc|ccc|cc|ccc}
    \toprule
    \textbf{Method} & \rotatebox{90}{\textcolor{robo_green}{$\bullet$~Clean~}} & \rotatebox{90}{\textcolor{robo_blue}{$\bullet$~T.O.}} & \rotatebox{90}{\textcolor{robo_red}{$\bullet$}~Brightness~} & \rotatebox{90}{\textcolor{robo_red}{$\bullet$}~Dark~} & \rotatebox{90}{\textcolor{robo_red}{$\bullet$}~Snow~} & \rotatebox{90}{\textcolor{robo_red}{$\bullet$}~Fog~} & \rotatebox{90}{\textcolor{robo_red}{$\bullet$}~Rain~} & \rotatebox{90}{\textcolor{robo_red}{$\bullet$}~Lens Obstacle~} & \rotatebox{90}{\textcolor{robo_red}{$\bullet$}~Water Splash~} & \rotatebox{90}{\textcolor{robo_red}{$\bullet$}~Camera Crash~} & \rotatebox{90}{\textcolor{robo_red}{$\bullet$}~Frame Lost~} & \rotatebox{90}{\textcolor{robo_red}{$\bullet$}~Saturate~} & \rotatebox{90}{\textcolor{robo_red}{$\bullet$}~Motion Blur~} & \rotatebox{90}{\textcolor{robo_red}{$\bullet$}~Zoom Blur~} & \rotatebox{90}{\textcolor{robo_red}{$\bullet$}~Bit Error~} & \rotatebox{90}{\textcolor{robo_red}{$\bullet$}~Color Quant~} & \rotatebox{90}{\textcolor{robo_red}{$\bullet$}~H.265 Compression~~}
    \\
    \midrule\midrule
    \rowcolor{robo_green!10} \includegraphics[width=0.025\linewidth]{figures/icons/human.png}~\textcolor{robo_green}{Human} & \textcolor{robo_green}{$93.33$} & \textcolor{robo_green}{-} & \textcolor{robo_green}{$80.00$} & \textcolor{robo_green}{$53.33$} & \textcolor{robo_green}{$80.00$} & \textcolor{robo_green}{$33.33$} & \textcolor{robo_green}{$80.00$} & \textcolor{robo_green}{$66.67$} & \textcolor{robo_green}{$73.33$} & \textcolor{robo_green}{$80.00$} & \textcolor{robo_green}{$40.00$} & \textcolor{robo_green}{$46.67$} & \textcolor{robo_green}{$80.00$} & \textcolor{robo_green}{$40.00$} & \textcolor{robo_green}{$60.00$} & \textcolor{robo_green}{$46.67$} & \textcolor{robo_green}{$53.33$}
    \\
    \midrule
     \textcolor{gray}{GPT-4o} & \textcolor{gray}{$59.00$} & \textcolor{gray}{$59.50$} & \textcolor{gray}{$60.50$} & \textcolor{gray}{$63.50$} & \textcolor{gray}{$59.00$} & \textcolor{gray}{$61.00$} & \textcolor{gray}{$59.50$} & \textcolor{gray}{$58.00$} & \textcolor{gray}{$59.00$} & \textcolor{gray}{$50.00$} & \textcolor{gray}{$51.50$} & \textcolor{gray}{$56.50$} & \textcolor{gray}{$57.50$} & \textcolor{gray}{$52.00$} & \textcolor{gray}{$51.00$} & \textcolor{gray}{$54.00$} & \textcolor{gray}{$57.50$}
    \\
    \midrule
    \rowcolor{robo_gray!10} Phi3 & $54.50$ & $17.50$ & $55.00$ & $33.00$ & $50.00$ & $55.00$ & $59.00$ & $56.00$ & $57.50$ & $32.50$ & $23.50$ & $58.00$ & $57.50$ & $49.00$ & $36.00$ & $49.50$ & $57.50$ 
    \\
    Phi-3.5 & $56.50$ & $58.50$ & $59.00$ & $58.00$ & $58.00$ & $59.00$ & $58.50$ & $60.00$ & $59.50$ & $58.50$ & $57.00$ & $59.50$ & $59.00$ & $59.50$ & $57.50$ & $60.50$ & $58.50$ 
    \\
    \rowcolor{robo_gray!10} LLaVA-1.5$_\text{7B}$ & $50.00$ & $50.00$ & $50.00$ & $50.00$ & $50.00$ & $50.00$ & $50.00$ & $50.00$ & $50.00$ & $50.00$ & $50.00$ & $50.00$ & $50.00$ & $50.00$ & $50.00$ & $50.00$ & $50.00$ 
    \\
    LLaVA-1.5$_\text{13B}$ & $50.00$ & $50.00$ & $50.00$ & $50.00$ & $50.00$ & $50.00$ & $50.00$ & $50.00$ & $50.00$ & $50.00$ & $50.00$ & $50.00$ & $50.00$ & $50.00$ & $50.00$ & $50.00$ & $50.00$ 
    \\
    \rowcolor{robo_gray!10} LLaVA-NeXT & $55.00$ & $34.50$ & $53.50$ & $42.50$ & $41.50$ & $44.00$ & $55.00$ & $42.00$ & $52.00$ & $31.00$ & $26.50$ & $50.50$ & $53.50$ & $39.00$ & $44.50$ & $37.50$ & $49.00$ 
    \\    
     InternVL$_\text{8B}$ & $56.50$ & $23.50$ & $57.50$ & $61.00$ & $62.00$ & $59.50$ & $59.50$ & $59.50$ & $60.00$ & $50.50$ & $31.00$ & $56.50$ & $56.00$ & $59.00$ & $60.00$ & $58.50$ & $58.50$ 
    \\
    \rowcolor{robo_gray!10} Oryx & $51.00$ & $19.00$ & $53.50$ & $50.50$ & $52.00$ & $52.50$ & $52.50$ & $50.00$ & $48.00$ & $32.50$ & $29.00$ & $52.00$ & $48.00$ & $36.00$ & $50.50$ & $52.00$ & $52.00$ \\
    
    Qwen2VL$_\text{7B}$ & $59.00$ & $56.50$ & $60.00$ & $59.00$ & $60.00$ & $59.50$ & $59.00$ & $59.00$ & $59.00$ & $58.00$ & $56.00$ & $59.00$ & $59.50$ & $55.00$ & $54.50$ & $57.00$ & $59.00$ 
    \\
    \rowcolor{robo_gray!10} Qwen2VL$_\text{72B}$ & $60.00$ & $23.50$ & $58.00$ & $60.00$ & $58.50$ & $59.50$ & $61.50$ & $60.00$ & $59.50$ & $39.00$ & $29.50$ & $60.00$ & $59.00$ & $55.50$ & $55.50$ & $58.50$ & $61.00$ 
    \\
    \midrule
    Dolphin & $7.00$ & $9.00$ & $5.50$ & $5.50$ & $5.50$ & $6.00$ & $4.50$ & $6.50$ & $5.50$ & $5.50$ & $7.50$ & $6.50$ & $6.00$ & $5.50$ & $6.00$ & $5.50$ & $5.50$ 
    \\
    \rowcolor{robo_gray!10} DriveLM & $40.00$ & $23.00$ & $37.00$ & $45.00$ & $35.00$ & $46.50$ & $45.50$ & $40.50$ & $37.50$ & $39.50$ & $29.50$ & $34.00$ & $42.00$ & $36.50$ & $36.00$ & $30.50$ & $47.00$ 
    \\
    \bottomrule
    \end{tabular}}
\label{tab:perception-mcq-acc}
\end{table*}
\begin{table*}[ht]
\centering
\caption{Detailed GPT score results of \textbf{MCQs} for the \includegraphics[width=0.026\linewidth]{figures/icons/behavior.png}~\textbf{Behavior} task. ``\textcolor{robo_green}{Clean}'' represents clean image inputs. ``\textcolor{robo_blue}{T.O.}'' represents text-only evaluation. The ``\textcolor{robo_red}{Corrupt}'' settings range from weather conditions, external disturbances, sensor failures, motion blur, and transmission errors. The benchmarked VLMs include commercial, open-sourced, and driving specialist models, respectively.}
\vspace{-0.1cm}
\resizebox{\linewidth}{!}{
    \begin{tabular}{r|c|c|ccccc|cc|ccc|cc|ccc}
    \toprule
    \textbf{Method} & \rotatebox{90}{\textcolor{robo_green}{$\bullet$~Clean~}} & \rotatebox{90}{\textcolor{robo_blue}{$\bullet$~T.O.}} & \rotatebox{90}{\textcolor{robo_red}{$\bullet$}~Brightness~} & \rotatebox{90}{\textcolor{robo_red}{$\bullet$}~Dark~} & \rotatebox{90}{\textcolor{robo_red}{$\bullet$}~Snow~} & \rotatebox{90}{\textcolor{robo_red}{$\bullet$}~Fog~} & \rotatebox{90}{\textcolor{robo_red}{$\bullet$}~Rain~} & \rotatebox{90}{\textcolor{robo_red}{$\bullet$}~Lens Obstacle~} & \rotatebox{90}{\textcolor{robo_red}{$\bullet$}~Water Splash~} & \rotatebox{90}{\textcolor{robo_red}{$\bullet$}~Camera Crash~} & \rotatebox{90}{\textcolor{robo_red}{$\bullet$}~Frame Lost~} & \rotatebox{90}{\textcolor{robo_red}{$\bullet$}~Saturate~} & \rotatebox{90}{\textcolor{robo_red}{$\bullet$}~Motion Blur~} & \rotatebox{90}{\textcolor{robo_red}{$\bullet$}~Zoom Blur~} & \rotatebox{90}{\textcolor{robo_red}{$\bullet$}~Bit Error~} & \rotatebox{90}{\textcolor{robo_red}{$\bullet$}~Color Quant~} & \rotatebox{90}{\textcolor{robo_red}{$\bullet$}~H.265 Compression~~}
    \\
    \midrule\midrule
    \rowcolor{robo_green!10}\includegraphics[width=0.025\linewidth]{figures/icons/human.png}~\textcolor{robo_green}{Human} & \textcolor{robo_green}{$65.00$} & \textcolor{robo_green}{-} & \textcolor{robo_green}{$40.00$} & \textcolor{robo_green}{$40.67$} & \textcolor{robo_green}{$66.67$} & \textcolor{robo_green}{$51.00$} & \textcolor{robo_green}{$73.33$} & \textcolor{robo_green}{$53.33$} & \textcolor{robo_green}{$56.00$} & \textcolor{robo_green}{$60.00$} & \textcolor{robo_green}{$36.67$} & \textcolor{robo_green}{$54.00$} & \textcolor{robo_green}{$67.67$} & \textcolor{robo_green}{$34.67$} & \textcolor{robo_green}{$53.33$} & \textcolor{robo_green}{$73.33$} & \textcolor{robo_green}{$53.33$}  
    \\
    \midrule
    \textcolor{gray}{GPT-4o} & \textcolor{gray}{$45.40$} & \textcolor{gray}{$50.03$} & \textcolor{gray}{$46.27$} & \textcolor{gray}{$49.20$} & \textcolor{gray}{$42.54$} & \textcolor{gray}{$41.79$} & \textcolor{gray}{$46.55$} & \textcolor{gray}{$45.30$} & \textcolor{gray}{$47.78$} & \textcolor{gray}{$45.28$} & \textcolor{gray}{$40.44$} & \textcolor{gray}{$47.89$} & \textcolor{gray}{$39.91$} & \textcolor{gray}{$32.25$} & \textcolor{gray}{$48.40$} & \textcolor{gray}{$50.35$} & \textcolor{gray}{$41.07$}
    \\
    \midrule
    \rowcolor{robo_gray!10}Phi-3 & $45.20$ & $40.91$ & $45.98$ & $44.48$ & $47.91$ & $45.17$ & $47.45$ & $44.22$ & $44.02$ & $43.65$ & $44.01$ & $43.51$ & $42.48$ & $41.05$ & $43.83$ & $46.60$ & $44.30$ 
    \\
    Phi-3.5 & $36.75$ & $39.16$ & $37.15$ & $38.14$ & $37.19$ & $39.53$ & $38.40$ & $36.79$ & $37.36$ & $36.83$ & $37.98$ & $39.09$ & $37.70$ & $38.91$ & $38.27$ & $38.23$ & $36.85$ \\
    \rowcolor{robo_gray!10}LLaVA-1.5$_\text{7B}$ & $13.60$ & $14.91$ & $12.79$ & $12.83$ & $15.57$ & $12.63$ & $14.06$ & $13.99$ & $12.79$ & $14.68$ & $13.65$ & $13.12$ & $13.55$ & $13.83$ & $13.98$ & $13.44$ & $13.48$ \\
    LLaVA-1.5$_\text{13B}$ & $32.99$ & $32.79$ & $33.34$ & $33.10$ & $33.10$ & $31.96$ & $32.44$ & $32.56$ & $32.49$ & $31.87$ & $31.55$ & $31.84$ & $33.17$ & $31.14$ & $33.40$ & $31.78$ & $33.72$ 
    \\
    \rowcolor{robo_gray!10}LLaVA-NeXT & $48.16$ & $11.92$ & $48.84$ & $38.82$ & $15.90$ & $39.13$ & $47.07$ & $20.72$ & $47.02$ & $48.20$ & $36.67$ & $39.69$ & $47.36$ & $39.60$ & $46.99$ & $28.13$ & $47.55$ 
    \\
    InternVL$_\text{8B}$ & $54.58$ & $20.14$ & $32.54$ & $36.95$ & $42.10$ & $56.72$ & $31.53$ & $31.09$ & $41.65$ & $50.17$ & $32.77$ & $43.66$ & $34.82$ & $34.90$ & $50.41$ & $50.78$ & $41.66$ 
    \\
    \rowcolor{robo_gray!10}Oryx & $33.92$ & $23.94$ & $34.19$ & $37.77$ & $33.02$ & $32.89$ & $32.56$ & $34.16$ & $34.83$ & $34.51$ & $34.82$ & $34.05$ & $33.95$ & $29.61$ & $35.25$ & $33.27$ & $32.33$ 
    \\
    Qwen2VL$_\text{7B}$ & $49.07$ & $46.93$ & $46.81$ & $48.75$ & $48.04$ & $47.64$ & $48.45$ & $46.80$ & $49.24$ & $47.95$ & $47.19$ & $49.58$ & $48.83$ & $41.27$ & $49.72$ & $47.07$ & $47.90$ 
    \\
    \rowcolor{robo_gray!10}Qwen2VL$_\text{72B}$ & $51.26$ & $39.46$ & $52.13$ & $51.24$ & $51.64$ & $49.75$ & $53.18$ & $52.46$ & $50.81$ & $51.25$ & $47.44$ & $51.22$ & $48.87$ & $35.72$ & $52.76$ & $49.77$ & $48.52$ 
    \\
    \midrule
    Dolphin & $8.81$ & $7.11$ & $7.17$ & $9.54$ & $9.02$ & $6.48$ & $8.05$ & $7.95$ & $7.10$ & $9.29$ & $8.94$ & $8.02$ & $8.02$ & $9.42$ & $8.37$ & $10.07$ & $6.32$ 
    \\
    \rowcolor{robo_gray!10}DriveLM & $42.78$ & $27.83$ & $47.18$ & $36.30$ & $40.70$ & $39.18$ & $40.93$ & $43.30$ & $40.98$ & $39.95$ & $38.23$ & $40.08$ & $45.68$ & $38.88$ & $41.10$ & $33.50$ & $39.65$ 
    \\
    \bottomrule
    \end{tabular}}
    \label{tab:behavior-mcq}
\end{table*}
\begin{table*}[t]
\centering
\caption{Detailed Accuracy score results of \textbf{MCQs} for the \includegraphics[width=0.026\linewidth]{figures/icons/behavior.png}~\textbf{Behavior} task. ``\textcolor{robo_green}{Clean}'' represents clean image inputs. ``\textcolor{robo_blue}{T.O.}'' represents text-only evaluation. The ``\textcolor{robo_red}{Corrupt}'' settings range from weather conditions, external disturbances, sensor failures, motion blur, and transmission errors. The benchmarked VLMs include commercial, open-sourced, and driving specialist models, respectively.}
\vspace{-0.1cm}
\resizebox{\linewidth}{!}{
    \begin{tabular}{r|c|c|ccccc|cc|ccc|cc|ccc}
    \toprule
    \textbf{Method} & \rotatebox{90}{\textcolor{robo_green}{$\bullet$~Clean~}} & \rotatebox{90}{\textcolor{robo_blue}{$\bullet$~T.O.}} & \rotatebox{90}{\textcolor{robo_red}{$\bullet$}~Brightness~} & \rotatebox{90}{\textcolor{robo_red}{$\bullet$}~Dark~} & \rotatebox{90}{\textcolor{robo_red}{$\bullet$}~Snow~} & \rotatebox{90}{\textcolor{robo_red}{$\bullet$}~Fog~} & \rotatebox{90}{\textcolor{robo_red}{$\bullet$}~Rain~} & \rotatebox{90}{\textcolor{robo_red}{$\bullet$}~Lens Obstacle~} & \rotatebox{90}{\textcolor{robo_red}{$\bullet$}~Water Splash~} & \rotatebox{90}{\textcolor{robo_red}{$\bullet$}~Camera Crash~} & \rotatebox{90}{\textcolor{robo_red}{$\bullet$}~Frame Lost~} & \rotatebox{90}{\textcolor{robo_red}{$\bullet$}~Saturate~} & \rotatebox{90}{\textcolor{robo_red}{$\bullet$}~Motion Blur~} & \rotatebox{90}{\textcolor{robo_red}{$\bullet$}~Zoom Blur~} & \rotatebox{90}{\textcolor{robo_red}{$\bullet$}~Bit Error~} & \rotatebox{90}{\textcolor{robo_red}{$\bullet$}~Color Quant~} & \rotatebox{90}{\textcolor{robo_red}{$\bullet$}~H.265 Compression~~}
    \\
    \midrule\midrule
    \rowcolor{robo_green!10} \includegraphics[width=0.025\linewidth]{figures/icons/human.png}~\textcolor{robo_green}{Human} & \textcolor{robo_green}{$66.67$} & \textcolor{robo_green}{-} & \textcolor{robo_green}{$40.00$} & \textcolor{robo_green}{$46.67$} & \textcolor{robo_green}{$66.67$} & \textcolor{robo_green}{$53.33$} & \textcolor{robo_green}{$73.33$} & \textcolor{robo_green}{$53.33$} & \textcolor{robo_green}{$53.33$} & \textcolor{robo_green}{$60.00$} & \textcolor{robo_green}{$40.00$} & \textcolor{robo_green}{$53.33$} & \textcolor{robo_green}{$66.67$} & \textcolor{robo_green}{$33.33$} & \textcolor{robo_green}{$53.33$} & \textcolor{robo_green}{$73.33$} & \textcolor{robo_green}{$53.33$}
    \\
    \midrule
    \textcolor{gray}{GPT-4o} & \textcolor{gray}{$25.50$} & \textcolor{gray}{$24.00$} & \textcolor{gray}{$25.50$} & \textcolor{gray}{$25.00$} & \textcolor{gray}{$21.50$} & \textcolor{gray}{$23.50$} & \textcolor{gray}{$25.00$} & \textcolor{gray}{$26.00$} & \textcolor{gray}{$24.00$} & \textcolor{gray}{$26.50$} & \textcolor{gray}{$28.50$} & \textcolor{gray}{$24.50$} & \textcolor{gray}{$23.50$} & \textcolor{gray}{$24.00$} & \textcolor{gray}{$26.00$} & \textcolor{gray}{$22.50$} & \textcolor{gray}{$21.50$}
    \\
    \midrule
    \rowcolor{robo_gray!10} Phi3 & $26.50$ & $30.00$ & $29.50$ & $29.50$ & $28.00$ & $29.50$ & $28.50$ & $27.00$ & $30.00$ & $32.50$ & $31.00$ & $28.50$ & $29.50$ & $23.50$ & $27.50$ & $31.50$ & $27.50$ \\
    Phi-3.5 & $36.50$ & $40.00$ & $37.00$ & $36.00$ & $37.00$ & $38.50$ & $38.00$ & $36.50$ & $35.50$ & $37.00$ & $39.00$ & $39.00$ & $37.50$ & $36.50$ & $39.00$ & $36.50$ & $36.00$ \\
    \rowcolor{robo_gray!10} LLaVA1.5$_\text{7B}$ & $10.00$ & $9.50$ & $8.50$ & $8.00$ & $8.00$ & $7.50$ & $8.00$ & $8.50$ & $8.00$ & $11.00$ & $10.00$ & $7.50$ & $9.00$ & $7.50$ & $11.00$ & $8.50$ & $8.00$ \\
    LLaVA1.5$_\text{13B}$ & $32.50$ & $33.00$ & $33.00$ & $33.00$ & $32.50$ & $32.50$ & $32.00$ & $32.50$ & $32.50$ & $32.00$ & $32.50$ & $32.50$ & $33.00$ & $31.00$ & $34.00$ & $31.00$ & $33.50$ \\
    \rowcolor{robo_gray!10} LLaVA-NeXT & $23.00$ & $15.00$ & $23.00$ & $24.00$ & $22.00$ & $23.00$ & $22.50$ & $24.50$ & $25.50$ & $27.50$ & $24.50$ & $26.50$ & $24.00$ & $23.00$ & $24.00$ & $21.50$ & $25.50$ \\
    InternVL$_\text{8B}$ & $27.50$ & $21.50$ & $9.00$ & $14.50$ & $20.50$ & $25.50$ & $13.00$ & $11.50$ & $15.00$ & $25.00$ & $17.50$ & $21.00$ & $11.50$ & $12.00$ & $28.00$ & $23.50$ & $18.50$ \\
    \rowcolor{robo_gray!10} Oryx & $21.00$ & $21.00$ & $21.50$ & $21.50$ & $21.50$ & $20.50$ & $21.50$ & $21.50$ & $21.50$ & $21.50$ & $22.00$ & $21.50$ & $22.00$ & $21.00$ & $22.50$ & $21.50$ & $21.50$ \\
    Qwen2VL$_\text{7B}$ & $30.00$ & $23.00$ & $29.00$ & $28.00$ & $25.00$ & $28.50$ & $27.50$ & $25.00$ & $28.50$ & $31.50$ & $28.50$ & $33.50$ & $26.00$ & $21.50$ & $27.00$ & $28.50$ & $30.00$ \\
    \rowcolor{robo_gray!10} Qwen2VL$_\text{72B}$ & $23.00$ & $36.50$ & $25.50$ & $24.50$ & $25.50$ & $22.00$ & $29.50$ & $26.00$ & $22.50$ & $27.00$ & $25.00$ & $26.00$ & $22.50$ & $22.00$ & $28.50$ & $23.50$ & $23.50$ \\
    \midrule
    Dolphin & $0.50$ & $3.50$ & $1.50$ & $0.00$ & $1.00$ & $0.00$ & $0.00$ & $1.00$ & $1.00$ & $1.50$ & $2.50$ & $0.50$ & $1.00$ & $1.00$ & $1.00$ & $0.50$ & $0.50$ \\
    \rowcolor{robo_gray!10} DriveLM & $44.00$ & $25.50$ & $48.50$ & $37.00$ & $41.50$ & $40.00$ & $42.50$ & $43.50$ & $41.00$ & $41.00$ & $39.50$ & $40.50$ & $46.50$ & $40.00$ & $43.00$ & $35.00$ & $41.50$ \\
    \bottomrule
    \end{tabular}}
\label{tab:behavior-mcq-acc}
\end{table*}
\begin{table*}[t]
\centering
\caption{Detailed GPT score results of \textbf{open-ended questions} for the \includegraphics[width=0.026\linewidth]{figures/icons/perception.png}~\textbf{Perception} task. ``\textcolor{robo_green}{Clean}'' represents clean image inputs. ``\textcolor{robo_blue}{T.O.}'' represents text-only evaluation. The ``\textcolor{robo_red}{Corrupt}'' settings range from weather conditions, external disturbances, sensor failures, motions, and transmission errors. The benchmarked VLMs include commercial, open-sourced, and driving specialist models, respectively.}
\vspace{-0.1cm}
\resizebox{\linewidth}{!}{
    \begin{tabular}{r|c|c|ccccc|cc|ccc|cc|cccc}
    \toprule
    \textbf{Method} & \rotatebox{90}{\textcolor{robo_green}{$\bullet$~Clean~}} & \rotatebox{90}{\textcolor{robo_blue}{$\bullet$~T.O.}} & \rotatebox{90}{\textcolor{robo_red}{$\bullet$}~Brightness~} & \rotatebox{90}{\textcolor{robo_red}{$\bullet$}~Dark~} & \rotatebox{90}{\textcolor{robo_red}{$\bullet$}~Snow~} & \rotatebox{90}{\textcolor{robo_red}{$\bullet$}~Fog~} & \rotatebox{90}{\textcolor{robo_red}{$\bullet$}~Rain~} & \rotatebox{90}{\textcolor{robo_red}{$\bullet$}~Lens Obstacle~} & \rotatebox{90}{\textcolor{robo_red}{$\bullet$}~Water Splash~} & \rotatebox{90}{\textcolor{robo_red}{$\bullet$}~Camera Crash~} & \rotatebox{90}{\textcolor{robo_red}{$\bullet$}~Frame Lost~} & \rotatebox{90}{\textcolor{robo_red}{$\bullet$}~Saturate~} & \rotatebox{90}{\textcolor{robo_red}{$\bullet$}~Motion Blur~} & \rotatebox{90}{\textcolor{robo_red}{$\bullet$}~Zoom Blur~} & \rotatebox{90}{\textcolor{robo_red}{$\bullet$}~Bit Error~} & \rotatebox{90}{\textcolor{robo_red}{$\bullet$}~Color Quant~} & \rotatebox{90}{\textcolor{robo_red}{$\bullet$}~H.265 Compression~~}
    \\
    \midrule\midrule
    \textcolor{gray}{GPT-4o} & \textcolor{gray}{$28.87$} & \textcolor{gray}{$29.37$} & \textcolor{gray}{$29.51$} & \textcolor{gray}{$28.15$} & \textcolor{gray}{$30.19$} & \textcolor{gray}{$28.89$} & \textcolor{gray}{$28.63$} & \textcolor{gray}{$28.49$} & \textcolor{gray}{$29.42$} & \textcolor{gray}{$27.12$} & \textcolor{gray}{$28.16$} & \textcolor{gray}{$27.96$} & \textcolor{gray}{$29.82$} & \textcolor{gray}{$32.79$} & \textcolor{gray}{$27.25$} & \textcolor{gray}{$26.44$} & \textcolor{gray}{$29.89$}
    \\
    \midrule
    \rowcolor{robo_gray!10}Phi-3 & $10.26$ & $10.38$ & $10.44$ & $11.27$ & $10.97$ & $10.81$ & $10.99$ & $10.64$ & $11.32$ & $10.67$ & $10.01$ & $11.03$ & $11.37$ & $12.31$ & $11.38$ & $10.33$ & $10.58$ 
    \\
    Phi-3.5 & $14.83$ & $18.20$ & $14.24$ & $13.08$ & $15.36$ & $16.09$ & $18.04$ & $16.53$ & $14.39$ & $16.42$ & $14.47$ & $12.88$ & $14.97$ & $12.96$ & $14.74$ & $13.27$ & $18.66$ 
    \\
    \rowcolor{robo_gray!10}LLaVA-1.5$_\text{7B}$ & $14.03$ & $11.94$ & $13.53$ & $13.31$ & $13.31$ & $13.61$ & $13.75$ & $13.91$ & $13.48$ & $13.95$ & $12.90$ & $12.98$ & $13.35$ & $13.05$ & $13.36$ & $12.83$ & $14.28$ 
    \\
    LLaVA-1.5$_\text{13B}$ & $13.13$ & $11.50$ & $12.72$ & $13.39$ & $12.86$ & $13.38$ & $13.05$ & $13.23$ & $13.59$ & $15.13$ & $14.98$ & $12.39$ & $13.60$ & $13.85$ & $12.72$ & $13.43$ & $12.26$ 
    \\
    \rowcolor{robo_gray!10}LLaVA-NeXT & $15.33$ & $23.53$ & $15.49$ & $14.95$ & $16.61$ & $15.62$ & $16.05$ & $15.66$ & $15.66$ & $15.19$ & $16.04$ & $15.16$ & $16.06$ & $17.92$ & $15.27$ & $15.10$ & $15.86$ 
    \\
    InternVL$_\text{8B}$ & $18.12$ & $14.73$ & $20.05$ & $18.82$ & $23.14$ & $14.28$ & $27.75$ & $23.43$ & $26.67$ & $11.88$ & $17.11$ & $26.26$ & $22.02$ & $27.25$ & $11.54$ & $29.57$ & $29.77$ 
    \\
    \rowcolor{robo_gray!10}Oryx & $16.07$ & $16.07$ & $15.72$ & $13.46$ & $11.98$ & $15.54$ & $13.73$ & $17.75$ & $15.86$ & $13.35$ & $13.29$ & $14.45$ & $13.74$ & $13.90$ & $13.17$ & $13.40$ & $14.66$ 
    \\
    Qwen2VL$_\text{7B}$ & $15.33$ & $32.56$ & $16.75$ & $14.95$ & $15.16$ & $15.66$ & $15.02$ & $15.66$ & $15.48$ & $17.31$ & $17.65$ & $14.52$ & $15.29$ & $15.20$ & $15.17$ & $14.63$ & $17.89$ 
    \\
    \rowcolor{robo_gray!10}Qwen2VL$_\text{72B}$ & $22.10$ & $13.88$ & $20.59$ & $17.00$ & $14.64$ & $18.77$ & $20.00$ & $24.19$ & $20.34$ & $17.55$ & $15.45$ & $19.13$ & $16.10$ & $18.58$ & $15.97$ & $18.78$ & $16.36$ 
    \\
    \midrule
    Dolphin & $12.68$ & $13.67$ & $12.07$ & $10.34$ & $12.37$ & $11.46$ & $11.73$ & $12.33$ & $11.04$ & $12.34$ & $11.39$ & $11.47$ & $11.34$ & $10.17$ & $11.40$ & $11.35$ & $11.65$ 
    \\
    \rowcolor{robo_gray!10}DriveLM & $11.32$ & $5.05$ & $11.30$ & $10.03$ & $11.21$ & $9.71$ & $10.22$ & $10.71$ & $11.01$ & $11.14$ & $10.13$ & $9.38$ & $10.97$ & $9.03$ & $11.39$ & $10.50$ & $10.73$ 
    \\
    \bottomrule
    \end{tabular}}
\label{tab:perception-open}
\end{table*}
\begin{table*}[t]
\centering
\caption{Detailed GPT score results of the \textbf{open-ended questions} for \includegraphics[width=0.026\linewidth]{figures/icons/prediction.png}~\textbf{Prediction}. ``\textcolor{robo_green}{Clean}'' represents clean image inputs. ``\textcolor{robo_blue}{T.O.}'' represents text-only evaluation. The ``\textcolor{robo_red}{Corrupt}'' settings range from weather conditions, external disturbances, sensor failures, motions, and transmission errors. The benchmarked VLMs include commercial, open-sourced, and driving specialist models, respectively.}
\vspace{-0.1cm}
\resizebox{\linewidth}{!}{
    \begin{tabular}{r|c|c|ccccc|cc|ccc|cc|ccc}
    \toprule
    \textbf{Method} & \rotatebox{90}{\textcolor{robo_green}{$\bullet$~Clean~}} & \rotatebox{90}{\textcolor{robo_blue}{$\bullet$~T.O.}} & \rotatebox{90}{\textcolor{robo_red}{$\bullet$}~Brightness~} & \rotatebox{90}{\textcolor{robo_red}{$\bullet$}~Dark~} & \rotatebox{90}{\textcolor{robo_red}{$\bullet$}~Snow~} & \rotatebox{90}{\textcolor{robo_red}{$\bullet$}~Fog~} & \rotatebox{90}{\textcolor{robo_red}{$\bullet$}~Rain~} & \rotatebox{90}{\textcolor{robo_red}{$\bullet$}~Lens Obstacle~} & \rotatebox{90}{\textcolor{robo_red}{$\bullet$}~Water Splash~} & \rotatebox{90}{\textcolor{robo_red}{$\bullet$}~Camera Crash~} & \rotatebox{90}{\textcolor{robo_red}{$\bullet$}~Frame Lost~} & \rotatebox{90}{\textcolor{robo_red}{$\bullet$}~Saturate~} & \rotatebox{90}{\textcolor{robo_red}{$\bullet$}~Motion Blur~} & \rotatebox{90}{\textcolor{robo_red}{$\bullet$}~Zoom Blur~} & \rotatebox{90}{\textcolor{robo_red}{$\bullet$}~Bit Error~} & \rotatebox{90}{\textcolor{robo_red}{$\bullet$}~Color Quant~} & \rotatebox{90}{\textcolor{robo_red}{$\bullet$}~H.265 Compression~~}
    \\
    \midrule\midrule
    \textcolor{gray}{GPT-4o} & \textcolor{gray}{$51.30$} & \textcolor{gray}{$49.05$} & \textcolor{gray}{$52.15$} & \textcolor{gray}{$50.28$} & \textcolor{gray}{$47.97$} & \textcolor{gray}{$49.66$} & \textcolor{gray}{$51.39$} & \textcolor{gray}{$50.49$} & \textcolor{gray}{$53.30$} & \textcolor{gray}{$46.62$} & \textcolor{gray}{$45.95$} & \textcolor{gray}{$49.18$} & \textcolor{gray}{$51.90$} & \textcolor{gray}{$49.75$} & \textcolor{gray}{$48.59$} & \textcolor{gray}{$47.56$} & \textcolor{gray}{$54.36$}
    \\
    \midrule
    \rowcolor{robo_gray!10}Phi-3 & $40.11$ & $22.61$ & $45.21$ & $26.28$ & $35.54$ & $44.05$ & $34.87$ & $40.90$ & $33.56$ & $44.26$ & $38.74$ & $41.39$ & $35.08$ & $36.46$ & $32.25$ & $32.82$ & $37.57$ 
    \\
    Phi-3.5 & $45.13$ & $4.92$ & $46.57$ & $9.67$ & $48.02$ & $45.08$ & $52.16$ & $42.75$ & $47.02$ & $47.18$ & $24.66$ & $23.26$ & $49.16$ & $28.38$ & $41.39$ & $18.02$ & $49.87$ 
    \\
    \rowcolor{robo_gray!10}LLaVA-1.5$_\text{7B}$ & $22.02$ & $14.64$ & $24.79$ & $20.95$ & $15.97$ & $15.30$ & $18.98$ & $16.28$ & $24.16$ & $6.11$ & $13.90$ & $20.30$ & $25.20$ & $10.56$ & $11.10$ & $15.61$ & $23.92$ 
    \\
    LLaVA-1.5$_\text{13B}$ & $36.98$ & $23.98$ & $36.00$ & $35.59$ & $40.51$ & $39.23$ & $38.90$ & $38.11$ & $38.92$ & $36.25$ & $36.54$ & $37.57$ & $38.10$ & $39.74$ & $34.28$ & $37.16$ & $36.31$ 
    \\
    \rowcolor{robo_gray!10}LLaVA-NeXT & $35.07$ & $28.36$ & $37.15$ & $35.31$ & $37.59$ & $37.62$ & $35.44$ & $37.00$ & $35.87$ & $36.25$ & $30.10$ & $40.56$ & $34.66$ & $39.36$ & $31.74$ & $34.07$ & $35.66$ 
    \\
    InternVL$_\text{8B}$ & $45.52$ & $48.89$ & $45.73$ & $40.71$ & $35.75$ & $38.43$ & $33.18$ & $38.69$ & $40.71$ & $45.00$ & $45.75$ & $30.03$ & $39.52$ & $36.55$ & $40.12$ & $28.40$ & $30.31$ 
    \\
    \rowcolor{robo_gray!10}Oryx & $48.13$ & $12.77$ & $49.52$ & $44.33$ & $47.67$ & $45.77$ & $47.20$ & $45.90$ & $50.30$ & $42.18$ & $44.59$ & $42.26$ & $48.21$ & $52.54$ & $45.56$ & $43.77$ & $49.61$ 
    \\
    Qwen2VL$_\text{7B}$ & $37.89$ & $37.77$ & $40.82$ & $35.90$ & $38.92$ & $44.15$ & $40.15$ & $41.89$ & $41.57$ & $36.61$ & $35.87$ & $41.25$ & $40.89$ & $39.23$ & $36.69$ & $40.52$ & $38.84$ 
    \\
    \rowcolor{robo_gray!10}Qwen2VL$_\text{72B}$ & $49.35$ & $5.57$ & $43.89$ & $43.25$ & $43.74$ & $44.49$ & $45.57$ & $41.61$ & $47.46$ & $40.89$ & $34.75$ & $39.38$ & $48.21$ & $51.15$ & $40.49$ & $40.82$ & $46.67$ 
    \\
    \midrule
    Dolphin & $32.66$ & $39.98$ & $29.85$ & $32.31$ & $24.64$ & $29.92$ & $31.38$ & $33.41$ & $31.79$ & $29.05$ & $30.93$ & $30.49$ & $31.59$ & $26.38$ & $30.13$ & $25.64$ & $30.62$ 
    \\
    \rowcolor{robo_gray!10}DriveLM & $44.33$ & $4.70$ & $46.82$ & $43.90$ & $42.33$ & $35.84$ & $44.13$ & $44.00$ & $42.59$ & $46.25$ & $33.56$ & $29.69$ & $42.15$ & $19.00$ & $38.20$ & $44.33$ & $42.87$ 
    \\
    \bottomrule
    \end{tabular}}
\label{tab:prediction-open}
\end{table*}
\begin{table*}[t]
\centering
\caption{Detailed ROUGE-L score results of \textbf{open-ended questions} for the \includegraphics[width=0.026\linewidth]{figures/icons/prediction.png}~\textbf{Predicion} task. ``\textcolor{robo_green}{Clean}'' represents clean image inputs. ``\textcolor{robo_blue}{T.O.}'' represents text-only evaluation. The ``\textcolor{robo_red}{Corrupt}'' settings range from weather conditions, external disturbances, sensor failures, motion blur, and transmission errors. The benchmarked VLMs include commercial, open-sourced, and driving specialist models, respectively.}
\vspace{-0.1cm}
\resizebox{\linewidth}{!}{
    \begin{tabular}{r|c|c|ccccc|cc|ccc|cc|ccc}
    \toprule
    \textbf{Method} & \rotatebox{90}{\textcolor{robo_green}{$\bullet$~Clean~}} & \rotatebox{90}{\textcolor{robo_blue}{$\bullet$~T.O.}} & \rotatebox{90}{\textcolor{robo_red}{$\bullet$}~Brightness~} & \rotatebox{90}{\textcolor{robo_red}{$\bullet$}~Dark~} & \rotatebox{90}{\textcolor{robo_red}{$\bullet$}~Snow~} & \rotatebox{90}{\textcolor{robo_red}{$\bullet$}~Fog~} & \rotatebox{90}{\textcolor{robo_red}{$\bullet$}~Rain~} & \rotatebox{90}{\textcolor{robo_red}{$\bullet$}~Lens Obstacle~} & \rotatebox{90}{\textcolor{robo_red}{$\bullet$}~Water Splash~} & \rotatebox{90}{\textcolor{robo_red}{$\bullet$}~Camera Crash~} & \rotatebox{90}{\textcolor{robo_red}{$\bullet$}~Frame Lost~} & \rotatebox{90}{\textcolor{robo_red}{$\bullet$}~Saturate~} & \rotatebox{90}{\textcolor{robo_red}{$\bullet$}~Motion Blur~} & \rotatebox{90}{\textcolor{robo_red}{$\bullet$}~Zoom Blur~} & \rotatebox{90}{\textcolor{robo_red}{$\bullet$}~Bit Error~} & \rotatebox{90}{\textcolor{robo_red}{$\bullet$}~Color Quant~} & \rotatebox{90}{\textcolor{robo_red}{$\bullet$}~H.265 Compression~~}
    \\
    \midrule\midrule
    \midrule
    \textcolor{gray}{GPT-4o} & \textcolor{gray}{$19.74$} & \textcolor{gray}{$18.58$} & \textcolor{gray}{$19.77$} & \textcolor{gray}{$19.58$} & \textcolor{gray}{$19.67$} & \textcolor{gray}{$19.71$} & \textcolor{gray}{$19.69$} & \textcolor{gray}{$20.22$} & \textcolor{gray}{$19.94$} & \textcolor{gray}{$19.89$} & \textcolor{gray}{$19.66$} & \textcolor{gray}{$19.70$} & \textcolor{gray}{$19.39$} & \textcolor{gray}{$19.48$} & \textcolor{gray}{$19.47$} & \textcolor{gray}{$19.90$} & \textcolor{gray}{$19.67$}
    \\
    \midrule
    \rowcolor{robo_gray!10} Phi3 & $17.76$ & $14.71$ & $17.21$ & $21.55$ & $25.81$ & $15.28$ & $23.77$ & $18.56$ & $19.27$ & $17.59$ & $17.42$ & $16.90$ & $16.75$ & $14.36$ & $25.23$ & $26.88$ & $22.55$ \\
    Phi-3.5 & $19.36$ & $18.76$ & $18.37$ & $5.78$ & $17.28$ & $15.74$ & $18.24$ & $18.38$ & $17.98$ & $17.10$ & $12.63$ & $10.84$ & $17.85$ & $12.46$ & $17.12$ & $8.50$ & $17.03$ \\
    \rowcolor{robo_gray!10} LLaVA1.5$_\text{7B}$ & $21.18$ & $23.21$ & $21.75$ & $22.64$ & $22.21$ & $22.17$ & $22.19$ & $19.61$ & $22.05$ & $22.53$ & $22.36$ & $22.65$ & $21.71$ & $20.28$ & $22.46$ & $22.57$ & $21.17$ \\
    LLaVA1.5$_\text{13B}$ & $24.12$ & $24.79$ & $24.03$ & $24.04$ & $24.09$ & $24.37$ & $24.14$ & $24.18$ & $24.35$ & $23.91$ & $23.91$ & $23.84$ & $24.04$ & $24.59$ & $23.93$ & $24.31$ & $24.12$ \\
    \rowcolor{robo_gray!10} qwen2-7b & $25.49$ & $24.15$ & $25.64$ & $24.80$ & $25.21$ & $24.82$ & $24.86$ & $25.17$ & $25.64$ & $25.62$ & $25.99$ & $24.74$ & $25.11$ & $23.73$ & $25.09$ & $25.31$ & $25.35$ \\
    Qwen2VL$_\text{72B}$ & $23.42$ & $16.10$ & $20.05$ & $18.46$ & $17.97$ & $19.08$ & $19.59$ & $18.68$ & $18.51$ & $22.05$ & $21.17$ & $17.97$ & $18.98$ & $19.03$ & $18.89$ & $19.35$ & $18.13$ \\
    \rowcolor{robo_gray!10} LLaVA1.6-7b & $16.09$ & $13.93$ & $16.29$ & $16.99$ & $17.02$ & $17.14$ & $16.54$ & $16.27$ & $17.08$ & $16.25$ & $16.59$ & $18.05$ & $16.69$ & $16.93$ & $16.58$ & $17.33$ & $16.46$ \\
    InternVL$_\text{8B}$ & $13.92$ & $13.25$ & $20.57$ & $15.14$ & $14.67$ & $13.64$ & $15.15$ & $16.39$ & $15.45$ & $13.92$ & $16.12$ & $14.47$ & $16.29$ & $14.79$ & $14.02$ & $14.26$ & $14.07$ \\
    \rowcolor{robo_gray!10} Oryx & $21.03$ & $15.43$ & $18.79$ & $17.45$ & $16.99$ & $18.04$ & $18.19$ & $17.81$ & $17.68$ & $20.38$ & $19.56$ & $16.54$ & $18.21$ & $17.84$ & $18.20$ & $17.92$ & $17.03$ \\
    Qwen2VL$_\text{7B}$ & $25.49$ & $23.15$ & $25.64$ & $24.80$ & $25.21$ & $24.82$ & $24.86$ & $25.17$ & $25.64$ & $25.62$ & $25.99$ & $24.74$ & $25.11$ & $23.73$ & $25.09$ & $25.31$ & $25.35$ \\
    \rowcolor{robo_gray!10} Qwen2VL$_\text{72B}$ & $23.42$ & $16.10$ & $20.05$ & $18.46$ & $17.97$ & $19.08$ & $19.59$ & $18.68$ & $18.51$ & $22.05$ & $21.17$ & $17.97$ & $18.98$ & $19.03$ & $18.89$ & $19.35$ & $18.13$ \\
    \midrule
    Dolphin & $25.18$ & $23.64$ & $24.71$ & $24.84$ & $25.95$ & $24.98$ & $26.19$ & $25.09$ & $25.18$ & $24.52$ & $24.75$ & $24.48$ & $24.99$ & $24.63$ & $24.87$ & $24.34$ & $24.95$ \\
    \rowcolor{robo_gray!10} DriveLM & $40.00$ & $23.00$ & $37.00$ & $45.00$ & $35.00$ & $46.50$ & $45.50$ & $40.50$ & $37.50$ & $39.50$ & $29.50$ & $34.00$ & $42.00$ & $36.50$ & $36.00$ & $30.50$ & $47.00$ \\
    \bottomrule
    \end{tabular}}
\label{tab:prediction-open-rougel}
\end{table*}
\begin{table*}[t]
\centering
\caption{Detailed GPT score results of the \textbf{open-ended questions} for the \includegraphics[width=0.026\linewidth]{figures/icons/planning.png}~\textbf{Planning}. ``\textcolor{robo_green}{Clean}'' represents clean image inputs. ``\textcolor{robo_blue}{T.O.}'' represents text-only evaluation. The ``\textcolor{robo_red}{Corrupt}'' settings range from weather conditions, external disturbances, sensor failures, motions, and transmission errors. The benchmarked VLMs include commercial, open-sourced, and driving specialist models, respectively.}
\vspace{-0.1cm}
    \resizebox{\linewidth}{!}{
    \begin{tabular}{r|c|c|ccccc|cc|ccc|cc|ccc}
    \toprule
    \textbf{Method} & \rotatebox{90}{\textcolor{robo_green}{$\bullet$~Clean~}} & \rotatebox{90}{\textcolor{robo_blue}{$\bullet$~T.O.}} & \rotatebox{90}{\textcolor{robo_red}{$\bullet$}~Brightness~} & \rotatebox{90}{\textcolor{robo_red}{$\bullet$}~Dark~} & \rotatebox{90}{\textcolor{robo_red}{$\bullet$}~Snow~} & \rotatebox{90}{\textcolor{robo_red}{$\bullet$}~Fog~} & \rotatebox{90}{\textcolor{robo_red}{$\bullet$}~Rain~} & \rotatebox{90}{\textcolor{robo_red}{$\bullet$}~Lens Obstacle~} & \rotatebox{90}{\textcolor{robo_red}{$\bullet$}~Water Splash~} & \rotatebox{90}{\textcolor{robo_red}{$\bullet$}~Camera Crash~} & \rotatebox{90}{\textcolor{robo_red}{$\bullet$}~Frame Lost~} & \rotatebox{90}{\textcolor{robo_red}{$\bullet$}~Saturate~} & \rotatebox{90}{\textcolor{robo_red}{$\bullet$}~Motion Blur~} & \rotatebox{90}{\textcolor{robo_red}{$\bullet$}~Zoom Blur~} & \rotatebox{90}{\textcolor{robo_red}{$\bullet$}~Bit Error~} & \rotatebox{90}{\textcolor{robo_red}{$\bullet$}~Color Quant~} & \rotatebox{90}{\textcolor{robo_red}{$\bullet$}~H.265 Compression~~}
    \\
    \midrule\midrule
    \textcolor{gray}{GPT-4o} & \textcolor{gray}{$75.75$} & \textcolor{gray}{$73.21$} & \textcolor{gray}{$77.56$} & \textcolor{gray}{$74.33$} & \textcolor{gray}{$73.00$} & \textcolor{gray}{$76.58$} & \textcolor{gray}{$76.53$} & \textcolor{gray}{$75.14$} & \textcolor{gray}{$74.65$} & \textcolor{gray}{$84.08$} & \textcolor{gray}{$74.54$} & \textcolor{gray}{$74.84$} & \textcolor{gray}{$77.34$} & \textcolor{gray}{$73.09$} & \textcolor{gray}{$74.30$} & \textcolor{gray}{$76.36$} & \textcolor{gray}{$76.82$}
    \\
    \midrule
    \rowcolor{robo_gray!10}Phi-3 & $60.03$ & $46.88$ & $61.48$ & $59.14$ & $64.59$ & $64.81$ & $63.47$ & $61.83$ & $62.31$ & $58.32$ & $54.63$ & $63.04$ & $61.61$ & $58.93$ & $58.11$ & $64.04$ & $63.40$ 
    \\
    Phi-3.5 & $31.91$ & $46.30$ & $30.57$ & $28.64$ & $32.88$ & $28.38$ & $34.70$ & $30.88$ & $29.53$ & $23.16$ & $24.53$ & $25.47$ & $32.73$ & $23.00$ & $28.24$ & $23.42$ & $29.31$ 
    \\
    \rowcolor{robo_gray!10}LLaVA-1.5$_\text{7B}$ & $29.15$ & $32.45$ & $31.52$ & $31.49$ & $32.58$ & $32.42$ & $31.00$ & $29.81$ & $31.72$ & $33.70$ & $35.95$ & $29.93$ & $31.20$ & $30.65$ & $30.05$ & $30.00$ & $30.61$ 
    \\
    LLaVA-1.5$_\text{13B}$ & $34.26$ & $38.85$ & $33.63$ & $34.36$ & $35.06$ & $39.43$ & $35.18$ & $33.01$ & $34.80$ & $37.61$ & $39.32$ & $32.77$ & $34.74$ & $32.99$ & $33.75$ & $33.33$ & $33.67$ 
    \\
    \rowcolor{robo_gray!10}LLaVA-NeXT & $45.27$ & $27.58$ & $45.64$ & $44.54$ & $43.55$ & $44.17$ & $45.08$ & $44.62$ & $44.21$ & $45.69$ & $44.51$ & $41.17$ & $45.30$ & $44.57$ & $43.67$ & $43.57$ & $45.13$ 
    \\
    InternVL$_\text{8B}$ & $53.27$ & $34.56$ & $54.70$ & $60.02$ & $63.89$ & $53.69$ & $60.64$ & $54.31$ & $56.68$ & $52.14$ & $46.12$ & $54.44$ & $55.94$ & $54.93$ & $49.08$ & $57.02$ & $55.13$ 
    \\
    \rowcolor{robo_gray!10}Oyrx & $53.57$ & $48.26$ & $55.46$ & $58.60$ & $58.91$ & $57.93$ & $55.04$ & $57.01$ & $58.35$ & $52.28$ & $51.05$ & $55.76$ & $55.88$ & $53.63$ & $52.94$ & $56.00$ & $57.57$ 
    \\
    Qwen2VL$_\text{7B}$ & $57.04$ & $41.66$ & $54.19$ & $58.37$ & $52.70$ & $58.18$ & $55.85$ & $53.98$ & $55.64$ & $54.71$ & $53.18$ & $51.93$ & $55.22$ & $51.73$ & $56.79$ & $53.16$ & $56.04$ 
    \\
    \rowcolor{robo_gray!10}Qwen2VL$_\text{72B}$ & $61.30$ & $53.35$ & $62.01$ & $65.31$ & $66.69$ & $68.15$ & $67.83$ & $65.67$ & $65.26$ & $57.42$ & $56.34$ & $62.06$ & $64.20$ & $61.66$ & $58.23$ & $59.86$ & $65.31$ 
    \\
    \midrule
    Dolphin & $52.91$ & $60.98$ & $51.85$ & $55.39$ & $53.09$ & $54.78$ & $53.92$ & $51.79$ & $53.57$ & $55.73$ & $57.81$ & $55.78$ & $51.42$ & $50.95$ & $53.35$ & $54.89$ & $52.17$ 
    \\
    \rowcolor{robo_gray!10}DriveLM & $68.71$ & $65.24$ & $67.25$ & $67.52$ & $65.72$ & $63.08$ & $69.60$ & $69.04$ & $67.97$ & $67.85$ &$66.47$ & $66.25$ & $67.93$ & $70.17$ & $68.46$ & $68.30$ & $68.59$ 
    \\
    \bottomrule
\end{tabular}}
\label{tab:planning-open}
\end{table*}
\begin{table*}[t]
\centering
\caption{Detailed ROUGE-L score results of \textbf{open-ended questions} for the \includegraphics[width=0.026\linewidth]{figures/icons/planning.png}~\textbf{Planning} task. ``\textcolor{robo_green}{Clean}'' represents clean image inputs. ``\textcolor{robo_blue}{T.O.}'' represents text-only evaluation. The ``\textcolor{robo_red}{Corrupt}'' settings range from weather conditions, external disturbances, sensor failures, motion blur, and transmission errors. The benchmarked VLMs include commercial, open-sourced, and driving specialist models, respectively.}
\vspace{-0.1cm}
\resizebox{\linewidth}{!}{
    \begin{tabular}{r|c|c|ccccc|cc|ccc|cc|ccc}
    \toprule
    \textbf{Method} & \rotatebox{90}{\textcolor{robo_green}{$\bullet$~Clean~}} & \rotatebox{90}{\textcolor{robo_blue}{$\bullet$~T.O.}} & \rotatebox{90}{\textcolor{robo_red}{$\bullet$}~Brightness~} & \rotatebox{90}{\textcolor{robo_red}{$\bullet$}~Dark~} & \rotatebox{90}{\textcolor{robo_red}{$\bullet$}~Snow~} & \rotatebox{90}{\textcolor{robo_red}{$\bullet$}~Fog~} & \rotatebox{90}{\textcolor{robo_red}{$\bullet$}~Rain~} & \rotatebox{90}{\textcolor{robo_red}{$\bullet$}~Lens Obstacle~} & \rotatebox{90}{\textcolor{robo_red}{$\bullet$}~Water Splash~} & \rotatebox{90}{\textcolor{robo_red}{$\bullet$}~Camera Crash~} & \rotatebox{90}{\textcolor{robo_red}{$\bullet$}~Frame Lost~} & \rotatebox{90}{\textcolor{robo_red}{$\bullet$}~Saturate~} & \rotatebox{90}{\textcolor{robo_red}{$\bullet$}~Motion Blur~} & \rotatebox{90}{\textcolor{robo_red}{$\bullet$}~Zoom Blur~} & \rotatebox{90}{\textcolor{robo_red}{$\bullet$}~Bit Error~} & \rotatebox{90}{\textcolor{robo_red}{$\bullet$}~Color Quant~} & \rotatebox{90}{\textcolor{robo_red}{$\bullet$}~H.265 Compression~~}
    \\
    \midrule\midrule
    \textcolor{gray}{GPT-4o} & \textcolor{gray}{$6.54$} & \textcolor{gray}{$6.42$} & \textcolor{gray}{$6.54$} & \textcolor{gray}{$6.34$} & \textcolor{gray}{$6.54$} & \textcolor{gray}{$6.54$} & \textcolor{gray}{$6.52$} & \textcolor{gray}{$6.40$} & \textcolor{gray}{$6.52$} & \textcolor{gray}{$6.47$} & \textcolor{gray}{$6.35$} & \textcolor{gray}{$6.39$} & \textcolor{gray}{$6.39$} & \textcolor{gray}{$6.40$} & \textcolor{gray}{$6.35$} & \textcolor{gray}{$6.37$} & \textcolor{gray}{$6.67$} \\
    \midrule
    \rowcolor{robo_gray!10} Phi3 & $9.39$ & $10.05$ & $9.56$ & $9.26$ & $9.51$ & $9.62$ & $9.47$ & $9.65$ & $9.60$ & $10.01$ & $10.01$ & $9.90$ & $9.40$ & $9.42$ & $9.07$ & $10.02$ & $9.36$ \\
    Phi-3.5 & $6.19$ & $8.27$ & $6.52$ & $7.97$ & $6.30$ & $6.10$ & $6.12$ & $6.57$ & $6.59$ & $7.18$ & $7.59$ & $5.98$ & $6.17$ & $7.87$ & $6.64$ & $6.04$ & $6.37$ \\
    \rowcolor{robo_gray!10} LLaVA1.5$_\text{7B}$ & $8.76$ & $9.98$ & $8.93$ & $8.99$ & $9.18$ & $8.96$ & $8.76$ & $9.00$ & $9.04$ & $9.06$ & $9.24$ & $8.88$ & $8.94$ & $9.27$ & $9.25$ & $9.16$ & $8.85$ \\
    LLaVA1.5$_\text{13B}$ & $8.06$ & $8.68$ & $8.14$ & $8.09$ & $8.10$ & $8.06$ & $8.10$ & $8.03$ & $7.99$ & $8.68$ & $8.64$ & $8.43$ & $8.10$ & $8.04$ & $8.39$ & $8.33$ & $8.24$ \\
    \rowcolor{robo_gray!10} LLaVA-NeXT & $5.69$ & $7.75$ & $5.72$ & $5.66$ & $5.50$ & $5.49$ & $5.60$ & $5.67$ & $5.60$ & $5.62$ & $5.88$ & $5.59$ & $5.64$ & $5.45$ & $5.64$ & $5.64$ & $5.58$ \\
    InternVL$_\text{8B}$ & $4.06$ & $9.45$ & $10.37$ & $11.65$ & $12.07$ & $4.05$ & $10.81$ & $10.97$ & $9.88$ & $4.16$ & $9.29$ & $11.02$ & $10.04$ & $9.23$ & $4.03$ & $10.24$ & $9.65$ \\
    \rowcolor{robo_gray!10} Oryx & $21.03$ & $15.43$ & $18.79$ & $17.45$ & $16.99$ & $18.04$ & $18.19$ & $17.81$ & $17.68$ & $20.38$ & $19.56$ & $16.54$ & $18.21$ & $17.84$ & $18.20$ & $17.92$ & $17.03$ \\
    Qwen2VL$_\text{7B}$ & $9.61$ & $8.33$ & $9.31$ & $8.70$ & $8.14$ & $8.33$ & $8.69$ & $8.32$ & $8.82$ & $9.15$ & $8.81$ & $8.69$ & $8.64$ & $9.05$ & $9.13$ & $8.77$ & $7.92$ \\
    \rowcolor{robo_gray!10} Qwen2VL$_\text{72B}$ & $12.26$ & $13.17$ & $12.13$ & $11.60$ & $11.45$ & $10.97$ & $11.62$ & $11.50$ & $11.67$ & $12.01$ & $11.43$ & $12.13$ & $11.85$ & $11.55$ & $12.06$ & $11.84$ & $11.93$ \\
    \midrule
    Dolphin & $12.90$ & $14.82$ & $12.90$ & $13.01$ & $12.61$ & $12.82$ & $12.45$ & $12.81$ & $12.74$ & $13.56$ & $13.91$ & $12.82$ & $12.86$ & $12.89$ & $13.22$ & $13.29$ & $12.69$ \\
    \rowcolor{robo_gray!10} DriveLM & $53.12$ & $46.83$ & $51.55$ & $51.26$ & $49.26$ & $46.74$ & $52.02$ & $53.38$ & $52.77$ & $52.44$ & $50.91$ & $48.66$ & $53.28$ & $52.20$ & $49.88$ & $52.79$ & $51.43$ \\
    \bottomrule
    \end{tabular}}
\label{tab:planning-open-rougel}
\end{table*}
\clearpage\clearpage
\section{Broader Impact \& Limitations}
In this section, we discuss the broader implications of our study and acknowledge its potential limitations.

\subsection{Broader Impact}
Our research focuses on evaluating the reliability of VLMs in autonomous driving, emphasizing three critical perspectives: the model’s robustness, data quality, and evaluation metrics. The findings reveal a concerning tendency of VLMs to fabricate explanations, particularly under conditions of visual degradation. This issue is not limited to autonomous driving but is likely relevant to other VLM-embodied systems, such as robotics and other safety-critical cyber-physical systems. For example, VLM-based robots could generate misleading task explanations or actions based on hallucinated understanding, potentially compromising safety and operational reliability.

The implications of our work extend beyond autonomous driving, calling for a reassessment of benchmark and metric designs to better evaluate the trustworthiness of VLMs in real-world applications. Current benchmarks often fail to account for the complexity and variability of real-world scenarios, particularly in environments where system malfunctions could result in life-threatening consequences. Our study highlights the urgency of addressing these gaps to develop robust, reliable, and interpretable VLMs that can be safely integrated into such systems.

Finally, the design of benchmarks, testbeds, and evaluation metrics that accurately capture the reliability and safety implications of applying VLMs to real-world physical systems is of paramount importance. These tools must go beyond traditional performance metrics to consider the nuanced requirements of autonomous systems, such as contextual understanding, interpretability, and robustness against adversarial conditions.

\subsection{Potential Limitations}
While this study provides valuable insights, it is essential to recognize its limitations to contextualize the findings:
\begin{itemize}
    \item The experimental results are derived exclusively from the DriveLM~\cite{sima2023drivelm} dataset due to the prohibitive computational cost of large-scale VLM inference and GPT-based evaluations. While DriveLM is a comprehensive dataset, its scope may limit the generalizability of our findings to other driving benchmarks or real-world settings. Future work should expand the analysis to additional datasets and environments to validate the observed trends.

    \item The lack of detailed contextual data in the DriveLM dataset poses a constraint on our evaluations. For instance, the GPT-based assessments rely on limited visual descriptions of key objects, which may not comprehensively capture the broader situational context required for accurate and nuanced evaluations. Expanding datasets to include richer temporal and spatial contexts could improve evaluation fidelity.
    
    \item This study primarily investigates the language-based explanations generated by VLMs. While these insights are crucial for understanding VLM reliability, it remains unclear whether the observations generalize to action models that generate vehicle trajectories or other non-language outputs. Exploring how VLMs’ visual grounding affects action-oriented tasks, such as trajectory prediction or manipulation control, represents an important direction for future research.

    \item The study evaluates a finite set of $12$ VLMs across specific tasks, metrics, and settings. Although the insights are significant, the scalability of these findings to emerging VLM architectures or more diverse driving scenarios warrants further investigation.
\end{itemize}
By addressing these limitations in future studies, we aim to build a more comprehensive understanding of the challenges and opportunities in applying VLMs to autonomous driving and other safety-critical domains.

\section{Public Resource Used}
\label{sec:public-resource}
In this section, we acknowledge the use of the following public resources, during the course of this work:
\begin{itemize}
    \item nuScenes\footnote{\url{https://www.nuscenes.org/nuscenes}.} \dotfill CC BY-NC-SA 4.0
    
    \item nuScenes-devkit\footnote{\url{https://github.com/nutonomy/nuscenes-devkit}.} \dotfill Apache License 2.0

    \item DriveLM\footnote{\url{https://github.com/OpenDriveLab/DriveLM}} \dotfill Apache License 2.0

    \item Phi-3.5-vision-instruct\footnote{\url{https://huggingface.co/microsoft/Phi-3.5-vision-instruct}} \dotfill MIT License

    \item Phi-3-mini-4k-instruct\footnote{\url{https://huggingface.co/microsoft/Phi-3-mini-4k-instruct}} \dotfill MIT License

    \item LLaVA-1.5-7B-hf\footnote{\url{https://huggingface.co/llava-hf/llava-1.5-7b-hf}} \dotfill Apache License 2.0

    \item LLaVA-1.5-13B-hf\footnote{\url{https://huggingface.co/llava-hf/llava-1.5-13b-hf}} \dotfill Apache License 2.0

    \item LLaVA-v1.6-mistral-7B\footnote{\url{https://huggingface.co/liuhaotian/llava-v1.6-mistral-7b}} \dotfill Apache License 2.0

    \item InternVL2-8B\footnote{\url{https://huggingface.co/OpenGVLab/InternVL2-8B}} \dotfill Apache License 2.0

    \item Qwen2-VL-7B\footnote{\url{https://huggingface.co/Qwen/Qwen2-VL-7B-Instruct}} \dotfill Apache License 2.0

    \item Qwen2-VL-72B\footnote{\url{https://huggingface.co/Qwen/Qwen2-VL-72B-Instruct}} \dotfill Apache License 2.0
\end{itemize}

\clearpage
\clearpage
{
    \small
    \bibliographystyle{ieeenat_fullname}
    \bibliography{main}
}

\end{document}